\documentclass[lettersize,journal]{IEEEtran}

\usepackage[bookmarks=true]{hyperref}
\usepackage{microtype}
\usepackage{amsmath,amsfonts}
\usepackage{algorithmic}
\usepackage{algorithm}
\usepackage{array}
\usepackage{booktabs}
\usepackage[caption=false,font=small,labelfont=sf,textfont=sf]{subfig}
\usepackage{dsfont}
\usepackage{multirow}
\usepackage{makecell}
\usepackage{textcomp}
\usepackage[table]{xcolor}
\usepackage{stfloats}
\usepackage{url}
\usepackage{verbatim}
\usepackage{graphicx}
\usepackage{cite}
\begin{document}

\title{Behavioural Quality Is Not Learning Utility: Objective Mismatch in Human Behaviour Models for Robot Learning}

\author{
Haoyang Jiang, Dana Kulic, Michael Burke, and Pamela Carreno-Medrano
\thanks{All authors are with the Department of Electrical and Computer Systems Engineering, Monash University, Clayton, Australia. Emails: Haoyang.Jiang1@monash.edu, Dana.Kulic@monash.edu, Michael.G.Burke@monash.edu, Pamela.Carreno@monash.edu.}
}

\maketitle

\begin{abstract}

Human behaviour models are increasingly used to provide simulated human interactions for training robots via reinforcement learning (RL). Existing approaches primarily evaluate these models according to how closely their generated behaviours resemble real human data, while their usefulness as training partners is less commonly assessed. This creates a potential objective mismatch: a model that performs well according to behavioural evaluation metrics may not necessarily provide useful training experiences for downstream robot policy learning. Across two human-robot interaction (HRI) domains, highway driving and object handover, we evaluate different types of human behaviour models using common generative-model evaluation metrics. We consider multiple human trajectory representations and compare the resulting evaluations with the downstream performance of RL policies trained using the corresponding models. We find that existing behavioural evaluation metrics do not consistently predict downstream policy performance across domains and trajectory representations, while their evaluations are sensitive to the choice of trajectory representation. We also show that even when the true testing behaviour distribution is known, training against a different distribution may yield better downstream performance than training directly against the true distribution.

\end{abstract}


\section{Introduction}
\label{sec:introduction}

As robots transition from controlled industrial settings to dynamic, human-centred environments, learning from and adapting to human behaviour becomes increasingly critical for successful deployment in real-world settings \cite{Kroemer2021, Thomaz2016}. In this context, reinforcement learning (RL) has emerged as a promising framework for robot training, as it enables agents to optimise long-horizon behaviour through interaction and feedback  \cite{backman_ReinforcementLearningShared_2023, sen_DomainRandomizationLearning_2025}.

In order to learn how to interact with human partners via RL, a robot must be exposed to interactions with humans during training. Although training robots using data collected from real-world human interactions remains the gold standard for achieving robust performance, this approach is often costly, time-consuming, and difficult to scale \cite{Kober2013}. To address these challenges, the robotics community has increasingly relied on simulated or modelled human behaviour models to train robot policies \cite{chen_CrowdRobotInteractionCrowdAware_2019a, sen_DomainRandomizationLearning_2025}. These models, ranging from simple rule-based approaches \cite{Gipps1981, Krauss1997, Treiber2000} to complex data-driven generators \cite{salzmann_TrajectronDynamicallyFeasibleTrajectory_2020a, zhang2023trafficbots, zhang2024trafficbotsv15}, offer a scalable and cost-effective means of producing the large volumes of interaction data required by modern machine learning methods \cite{Mavrogiannis2023}. Consequently, human behaviour models serve not only as models of human behaviour, but also as interactive partners for robot policy training. Their value therefore depends not only on how well they reproduce human behaviour, but also on how effectively they support downstream robot learning.

This creates an important distinction between \textit{behavioural quality} and \textit{downstream utility}. Behavioural quality concerns how closely a given model reproduces behaviours observed in real human data, whereas downstream utility concerns how effectively interactions generated by this same model support the learning of robot policies. We argue that these objectives may not be aligned. A model may accurately reproduce the distribution of behaviours observed in human data while failing to expose a robot to interactions that are important for learning an effective policy. Conversely, a model that generates behaviours that are less representative of the empirical human distribution may provide training experiences that lead to better downstream policies.
\looseness=-1

This distinction reflects a broader objective-mismatch problem that has been studied in model-based RL \cite{lambert2020objective}: model quality under a predictive or model-fitting objective does not necessarily imply usefulness for downstream decision making. Related work has demonstrated that models that are accurate according to conventional predictive objectives may nevertheless produce poor downstream control performance. This suggests that evaluating a model without considering the downstream task can provide an incomplete picture of its practical value. However, whether a similar mismatch arises when human behaviour models are used as training environments for HRI policy learning has not been systematically investigated.

This question is increasingly relevant in simulation-based robotics, where human behaviour models are used to construct interactive environments for training and evaluating autonomous agents. Recent traffic simulation frameworks, including WOSAC, TrafficBots, Symphony, BITS, and TrafficBots V1.5, have developed increasingly sophisticated models of human road-user behaviour for autonomous driving and evaluate these models primarily according to measures of behavioural realism, such as trajectory likelihood, prediction error, collision and rule-violation rates, diversity, and distributional similarity \cite{montali2023wosac, zhang2023trafficbots, igl2022symphony, xu2023bits, zhang2024trafficbotsv15}. These measures provide valuable assessments of whether a simulator faithfully reproduces human traffic behaviour, but they do not directly establish whether models achieving higher behavioural quality are more effective for downstream policy learning. Similarly, recent work on planner evaluation in reactive traffic environments has shown that changing the human behaviour simulator can affect the resulting assessment of autonomous planners \cite{hagedorn2025planners}. While this highlights the influence of the simulated humans on downstream evaluation, it does not establish whether realism-oriented evaluation metrics can predict the utility of these models for training robot policies. Thus, although existing work has made substantial progress towards evaluating the fidelity or diversity of human behaviour models, the relationship between behavioural faithfulness and downstream utility remains largely unexplored. This motivates our central research question:

\begin{quote}
\textbf{Can existing model evaluation metrics predict the downstream utility of human behaviour models for RL policy learning HRI scenarios?}
\end{quote}

To investigate this question, we evaluate a diverse set of rule-based and data-driven human behaviour models across two representative HRI domains: highway driving and object handover. These domains represent substantially different forms of HRI, involving large-scale coordination and close-proximity physical collaboration, respectively. For each human behaviour model, we evaluate two complementary aspects: behavioural quality, measured using commonly adopted generative-model evaluation metrics across different trajectory representations, and downstream utility, measured by the performance of RL policies trained using the corresponding human behaviour model. By comparing these two evaluations, we investigate whether models that score favourably according to existing behavioural evaluation metrics also provide more useful training interactions for downstream robot learning.

Our study provides the following contributions:
\begin{itemize}
    \item We provide empirical evidence of a mismatch between behavioural similarity and downstream RL utility of human behaviour models. Specifically, we show that existing evaluation metrics are not a consistent predictor of downstream policy performance across both HRI domains and trajectory representations.
    \item We show that the form of this mismatch can be task-dependent, with the relationship between behavioural evaluation and downstream policy utility differing between highway driving and object handover.
    \item We demonstrate that human behaviour model evaluation is sensitive to trajectory representation, with metric scores and model rankings varying substantially across representations and, in particular, learned feature configurations.
\end{itemize}

In addition to these contributions, we provide detailed formulations and parameter inference procedures for the rule-based human behaviour models used in our experiments. The parameters of these models are inferred from real human datasets \cite{highDdataset, khanna_MultimodalDataSet_2023}, with the corresponding model formulations, inference procedures, and fitted parameters provided in the Appendix, which is also included in our project page \href{https://objective-mismatch-hri.github.io/supplementary/}{https://objective-mismatch-hri.github.io/supplementary/}.
\looseness=-1

\section{Background}
\label{sec:background}

\subsection{Reinforcement Learning (RL) in HRI}
\label{subsec:reinforcementLearningInHRI}

RL is commonly used for training robots to operate in interactive and unstructured environments. By optimising long-horizon objectives through trial-and-error, RL enables robots to learn adaptive policies that account for both environmental dynamics and human responses \cite{Kober2013, Kroemer2021}. In HRI, RL has been applied to problems such as social navigation \cite{ahsen_HumanAwareSubgoalGeneration_2022, Xue2024, sen_DomainRandomizationLearning_2025}, assistive manipulation \cite{Chen2022}, collaborative decision-making \cite{Mehri2025} and shared autonomy \cite{backman_NoviceSkilledRLBased_2025}, where explicit supervision is difficult and optimal behaviour depends on sequential interaction with humans \cite{Thomaz2016, Mavrogiannis2023}.

Despite its conceptual appeal, RL faces well-known challenges in HRI. Real-world interaction with humans is expensive, safety-critical, and often ethically constrained, making large-scale data collection infeasible. Moreover, RL algorithms are data-hungry, particularly when learning from sparse or delayed rewards \cite{Ding2020, Akalin2021}, which is often the case in HRI scenarios. These challenges have motivated the use of simulated environments and human behaviour models to generate interaction experience for training RL agents across various domains, including autonomous driving \cite{zhang_LearningRealisticTraffic_2023}, social navigation \cite{vandermeer_ComparativeStudyHuman_2025} and human–agent interaction \cite{liang_LearningCooperateHumans_2024}.

\subsection{Human Behaviour Modelling for Robot Training}
\label{subsec:humanBehaviourModellingForRobotTraining}

To support scalable robot learning, a wide range of approaches have been proposed to model or simulate human behaviour. Early work relied on hand-crafted, heuristic models to encode simplified assumptions about human motion or decision-making \cite{Helbing1995, Trautman2010}. While computationally efficient, such models often fail to capture the diversity and stochasticity of real human behaviour. More recent approaches leverage data-driven techniques, including probabilistic models \cite{Trautman2013}, inverse planning \cite{Chris2009}, and deep generative models \cite{Mavrogiannis2023}, to learn human behaviour directly from data.

In the context of HRI, human behaviour models, either open-loop, where human actions are generated independently of the robot's actions, or interactive, where human behaviour adapts in response to the robot's actions, are frequently used as environment components for training robot policies, particularly when using RL \cite{Abeyruwan2023, sen_DomainRandomizationLearning_2025, jiang_InfluenceBasedRewardModulation_2026, VanDerMeer2026}. These models are used to substitute real humans, producing interactions that shape the robot’s learning process. Thus, their quality has a direct impact on the learned policies.

\subsection{Objective Mismatch in RL}
\label{subsec:objectiveMismatchInRL}

A related issue arises in RL when models are evaluated using objectives that do not directly reflect their intended use. Lambert et al. \cite{lambert2020objective} termed this \textit{objective mismatch}, noting that performance under one evaluation objective does not necessarily translate to downstream control performance. For example, in model-based RL, predictive accuracy alone may not indicate how useful a dynamics model is for learning a control policy.

This has motivated approaches that make model and environment selection more closely tied to the downstream task. Some methods evaluate models based on their effect on policy learning and control performance rather than predictive fidelity alone \cite{farahmand_ValueAwareLossFunction_2017, nikishin_ControlOrientedModelBasedReinforcement_2022}. Other approaches focus on the training environment itself. Domain randomisation exposes policies to diverse simulated environments to improve robustness to deployment variations \cite{tobin_DomainRandomizationTransferring_2017, peng_SimtoRealTransferRobotic_2018, sen_DomainRandomizationLearning_2025}, while other work has investigated how training distributions can be selected or adapted using downstream performance \cite{mehta_ActiveDomainRandomization_2020, chebotar_ClosingSimtoRealLoop_2019, muratore_DataEfficientDomainRandomization_2021}. These studies show that the properties of a model or training environment that are desirable under conventional evaluation do not necessarily lead to better downstream policies.

The same issue may arise when human behaviour models are used to train robots. In HRI, these models form part of the training environment, yet existing evaluation metrics primarily measure their similarity to observed human behaviour rather than their contribution to downstream robot learning. Whether such metrics can reliably identify human behaviour models that provide useful training experiences for RL therefore remains an open question.

\subsection{Evaluation Metrics for Generative Models}
\label{subsec:evaluationMetricsForGenerativeModels}

The dominant approach to evaluating generative models has been to measure their statistical similarity to real data. Metrics such as the \textit{Inception Score} (\textit{IS}) \cite{Salimans2016}, which jointly measures sample quality and diversity, \textit{Fr\'echet Distance} (\textit{FD}) \cite{frechen_inceptive_distance}, which quantifies distributional similarity between generated and real samples, \textit{precision} and \textit{recall} \cite{precision_recall_distributions}, which measure sample fidelity and distributional coverage respectively, and \textit{density} and \textit{coverage} \cite{density_coverage}, which assess sample concentration and support coverage, are widely used to assess sample quality, diversity, and distributional alignment. These metrics are commonly computed in a latent space, as directly measuring them in complex, multidimensional behavioural spaces can be challenging.
\looseness=-1

These metrics have also been adopted in a wide range of human data generation tasks, including gesture generation, motion generation, crowd simulation, and interaction modelling \cite{Ali2025}. Zabala et al. \cite{zabalaQuantitativeAnalysisRobot2021} compared two GAN-based talking-gesture generators deployed on a social robot and introduced the \textit{Fr\'echet Gesture Distance} by adapting the \textit{FD} to the gesture domain. Maiorca et al. \cite{maiorcaObjectiveEvaluationMetric2023} proposed the \textit{Fr\'echet Motion Distance} (\textit{FMD}) as a metric for evaluating human motion generation. Bamani et al. \cite{bamaniDiffusionBasedDataGenerator2024} introduced a diffusion-based generator, DUR, which synthesises images of humans performing gestures at large distances. They evaluated the visual quality of the generated images by comparing DUR against several GAN-based baselines using the IS and \textit{FD} metrics.
\looseness=-1

Collectively, these works illustrate a common evaluation paradigm in which generators are assessed according to their statistical similarity to reference data using variants of FD or related distributional metrics. To date, there is no widely adopted quantitative metric for assessing a human behaviour model according to its utility for downstream robot learning. However, while these measures assess whether generated samples resemble real human data, they do not account for how those samples could influence the robot's subsequent task performance, particularly under RL. 

\section{Problem Formulation and Overview}
\label{sec:problemFormulationAndOverview}

We formulate the closed-loop interaction between an autonomous robot and multiple human agents as a finite Markov Decision Process (MDP), defined by the tuple $\mathcal{M} = \langle \mathcal{S}, \mathcal{A}^{\text{robot}}, \mathcal{A}^{\text{human}}, \mathcal{P}, \mathcal{R}, \gamma \rangle$. At each time step $t$, the environment state $s_t \in \mathcal{S}$ comprises the state of the robot, the surrounding human agents, and static environmental elements. The state captures the agents' current configurations and implicitly reflects the effects of their previous actions. Consequently, the robot policy and human policies are conditioned on the current state rather than explicitly on each other's actions, enabling closed-loop interaction through state feedback. The robot executes an action $a_t^{\text{robot}} \in \mathcal{A}^{\text{robot}}$ sampled from its policy $\pi(\cdot \mid s_t)$, and $a_t^{\text{human}} \in \mathcal{A}^{\text{human}}$ denotes the action taken by the human. The objective of the robot is to learn an optimal policy $\pi^*$ that maximises the expected cumulative discounted return:
\begin{equation}
    J(\pi) = \mathbb{E}_{\pi, \mathcal{P}} \left[ \sum_{t=0}^{T} \gamma^t \mathcal{R}(s_t, a_t^{\text{robot}}) \right],
\end{equation}
where $\mathcal{R}$ is the reward function defining the task-specific objective, $\gamma \in [0, 1)$ is the discount factor, and $T$ is the horizon.

Crucially, the environmental transition dynamics depend not only on the robot's actions but also on the behaviour of the surrounding human agents. Let $a_t^{\text{human}}$ denote the joint actions of the human agents. The true state evolution is therefore governed by the joint dynamics:
\begin{equation}
    s_{t+1} \sim P_{\text{joint}}\!\left(s_{t+1} \mid s_t, a_t^{\text{robot}}, a_t^{\text{human}}\right).
\end{equation}

Rather than treating human behaviour implicitly, we use explicit, parameterised human behaviour models $\mathcal{H}_\theta$, which are constructed from human demonstrations, or analytical or learnable rule-based models. Given the current state, the human model defines a conditional distribution over human actions:
\begin{equation}
    a_t^{\text{human}}\sim p_{\mathcal{H}_\theta}(\cdot \mid s_t).
\end{equation}

Prior to deployment, each human behaviour model is specified either analytically or using human demonstrations. For human behaviour models parameterised by $\theta$, the parameters may represent interpretable behavioural quantities in rule-based approaches or learnable function parameters in data-driven models. When parameter estimation is required, the parameters are obtained from a dataset of human demonstrations
$\mathcal{D} = \{(s_k, a_k^{\text{human}})\}_{k=1}^{K}$ by solving
\begin{equation}
    \theta^* = \arg\max_{\theta} \mathcal{J}(\theta;\mathcal{D}),
\end{equation}
where $\mathcal{J}$ denotes a model-specific objective defined by the chosen trajectory representation and estimation method.

In this work, we consider a set of $N$ human behaviour models $\{ \mathcal{H}_\theta^{(i)}\}_{i=1}^{N}$, including both heuristic and data-driven approaches, and study how their properties influence downstream robot policy performance. Specifically, for each human behaviour model $\mathcal{H}_\theta^{(i)}$, a set of $M$ evaluation metrics $\{\mathcal{E}_j\}_{j=1}^{M}$ is used to characterise its properties, defined as
\begin{equation}
    e_{ij} = f_{\mathcal{E}_j}(\mathcal{H}_\theta^{(i)}).
\end{equation}
For a robot policy $\pi_i$ trained in an environment parameterised by a specific human behaviour model $\mathcal{H}_\theta^{(i)}$, we investigate the relationship between the resulting policy performance and the corresponding model evaluation metrics $\{e_{ij}\}_{j=1}^{M}$ computed from the behaviours generated by that human model.

The evaluation metrics $\{\mathcal{E}_j\}_{j=1}^{M}$ are selected from commonly adopted model evaluation metrics for assessing generative models. These metrics evaluate the generated behaviours in terms of their fidelity and diversity. Fidelity metrics quantify how closely the generated behaviours resemble the real data distribution. Higher fidelity indicates that the generated samples are statistically more consistent with the reference data. In this study, we consider two fidelity metrics: the \textit{Fr\'echet Distance} (\textit{FD}) \cite{frechen_inceptive_distance} and \textit{density} \cite{density_coverage}.

Complementing fidelity, diversity metrics quantify the variability of the generated behaviours. Higher diversity indicates that the generator captures a broader range of behaviours present in the underlying data distribution. We consider three diversity metrics: the \textit{average pairwise distance} (\textit{APD}), \textit{coverage} \cite{density_coverage}, and the \textit{mean maximum similarity} (\textit{MMS}) \cite{cervantes_ImplicitNeuralRepresentations_2022}.

To assess diversity, we first compute baseline \textit{APD} and \textit{MMS} values for the real human trajectories, denoted by $APD_{\mathrm{real}}$ and $MMS_{\mathrm{rr}} = \mathrm{MMS}(\mathcal{X}_{\mathrm{real}}, \mathcal{X}_{\mathrm{real}})$, respectively. These baseline values are then compared with the corresponding metrics computed for the generated trajectories, denoted by $APD_{\mathrm{gen}}$ and $MMS_{\mathrm{gr}} = \mathrm{MMS}(\mathcal{X}_{\mathrm{gen}}, \mathcal{X}_{\mathrm{real}})$. The APD gap, denoted by $APD_{\mathrm{gap}}$, and the MMS gap, denoted by $MMS_{\mathrm{gap}}$, are defined as
\begin{align}
    APD_{\mathrm{gap}} &= APD_{\mathrm{gen}} - APD_{\mathrm{real}} \\
    MMS_{\mathrm{gap}} &= MMS_{\mathrm{gr}} - MMS_{\mathrm{rr}},
\end{align}
and quantify the difference in diversity between the generated and real human trajectories. Details of the evaluation metrics are provided in the Appendix \ref{app:evaluation_metrics}.

Having established the MDP formulation and the model evaluation metrics used to characterise each human behaviour model, we apply them to two representative HRI domains. The first is highway driving, where the robot is an autonomous vehicle interacting with surrounding human-driven vehicles. This domain features continuous, multi-agent interactions in which human and robot's behaviour directly influences the safety and driving decisions of each other. Moreover, there exist large-scale naturalistic datasets \cite{u.s.departmentoftransportationfederalhighwayadministration_NextGenerationSimulation_2017, highDdataset, ettinger_LargeScaleInteractive_2021} that enable the construction of diverse human behaviour models. We use the highD dataset \cite{highDdataset}, which contains naturalistic vehicle trajectories recorded on German motorways, to construct and evaluate the human behaviour models for this domain.

The second domain is a simplified object handover task, where a robot must coordinate with a human to receive an object. Unlike the driving domain, this setting focuses on discrete and close-proximity physical interactions
, providing a complementary evaluation of human behaviour modelling in collaborative manipulation. We use the handover@RPL dataset \cite{khanna_MultimodalDataSet_2023}, which contains human-human handover demonstrations collected from 13 participant pairs, to construct the human behaviour models for this domain.

Our simulation-based evaluation proceeds in three stages. First, we construct a diverse set of human behaviour models from human datasets for each domain. Second, we evaluate these models using commonly adopted generative model evaluation metrics across different trajectory representations. Third, we use each human behaviour model as part of a simulated training environment for an RL agent and evaluate the resulting policy performance. We then analyse the relationship between human model metrics and downstream RL policy performance, with the goal of understanding whether commonly used metrics are indicative of a model’s usefulness for policy learning.

\section{Experiment Setup}\label{sec:experimentSetup}

\subsection{Two Domains}\label{subsec:twoDomains}

\subsubsection{\textbf{Highway Driving}}
We consider a highway driving scenario that simulates an autonomous ego vehicle interacting with human drivers. We partitioned the highD dataset~\cite{highDdataset}into two disjoint subsets: the first 40 recordings are used to train $N$ human behaviour models, denoted by $\{H_i^{\mathrm{HW,train}}\}_{i=1}^N$, used to train the autonomous policy, while the remaining 20 recordings are used to independently train $N$ corresponding evaluation models, denoted by $\{H_j^{\mathrm{HW,eval}}\}_{j=1}^N$. Thus, RL policies are trained against $\{H_i^{\mathrm{HW,train}}\}_{i=1}^N$ and subsequently evaluated against the corresponding models in $\{H_j^{\mathrm{HW,eval}}\}_{j=1}^N$.

We model driving behaviours with rule-based and data-driven models. The resulting human behaviour models control the surrounding vehicles, enabling them to react to other vehicles during downstream RL training and evaluation. For downstream evaluation, we train autonomous driving policies using Deep Q-Learning in the \textit{highway-env} framework~\cite{highway-env}. The ego vehicle interacts with multiple surrounding vehicles, each independently controlled by a single human behaviour model. The ego policy selects discrete driving actions, including acceleration, deceleration, and lane changes, with the objective of maintaining high-speed and safe driving. We use the computationally efficient \textit{highway-fast-v0} environment. Detailed configuration and RL training procedure are provided in the Appendix \ref{subapp:highway_driving} and \ref{subapp:highway_driving_rl_details}.

\subsubsection{\textbf{Object Handover}\label{subsubsec:handoverDomain}}

We consider a human-robot object handover task in which a human giver passes an object to a robot taker. Using the handover@RPL dataset~\cite{khanna_MultimodalDataSet_2023}, we model human giver behaviour from the establishment of a stable grip on the object until its release. 
We consider non-reactive, rule-based models that generate the giver's hand trajectory independently of the taker's movements, and reactive, data-driven models that generate the giver's actions in response to the taker's hand movements.

After preprocessing, 1,281 human giver trajectories remain for modelling. To construct independent training and evaluation settings, we partition the 12 right-handed participant pairs into two disjoint subsets. The first six pairs are used to train $N$ human behaviour models $\{H_i^{\mathrm{HO,train}}\}_{i=1}^N$ used during RL training, while the remaining six pairs are used to independently train $N$ corresponding models $\{H_j^{\mathrm{HO,eval}}\}_{j=1}^N$ used for downstream policy evaluation.

We develop a customised RL environment in which the human behaviour model controls the three-dimensional motion of the giver's hand and the RL agent controls the motion of the taker's hand. Both hands are represented as point masses, and the taker's objective is to reach the giver's hand while respecting the spatial constraints of the environment. An episode terminates when the taker reaches the giver or when a maximum episode length is reached. The agent is rewarded for reducing the distance to the giver and receives an additional reward upon successful completion of the handover. Further details of the dataset preprocessing and RL environment are provided in the Appendix \ref{subapp:object_handover} and \ref{subapp:handover_rl_details}.
\looseness=-1

\subsection{Human Behaviour Models}\label{subsec:humanBehaviourModels}

For each domain, we consider two categories of human behaviour models: rule-based and data-driven models.

\subsubsection{\textbf{Rule-based Driving Models}}

We construct complete microscopic rule-based driving models by combining the Minimising Overall Braking Induced by Lane Changes (MOBIL) model~\cite{Arne1999} for lateral decisions with a longitudinal car-following model. We consider three representative and widely used car-following models~\cite{Zhang2025}: the Gipps model~\cite{Gipps1981}, the Krau{\ss} model~\cite{Krauss1997}, and the Intelligent Driver Model (IDM)~\cite{Treiber2000}. Each complete rule-based behaviour model therefore combines MOBIL with one of these longitudinal models. We fit the parameters of each longitudinal model to highD data using variational inference. The model formulations, parameter inference procedures, and resulting parameter values are provided in Appendix \ref{subapp:mobil} (MOBIL), \ref{subapp:gipps} (Gipps), \ref{subapp:krauss} (Krau{\ss}), and \ref{subapp:idm} (IDM), respectively.

In addition to the calibrated rule-based models, we consider two stochastic vehicle baselines, denoted \textit{Unif} and \textit{Unif-W}. These vehicles independently sample longitudinal and lateral accelerations from uniform distributions at each time step. \textit{Unif-W} differs from \textit{Unif} by allowing a wider range of lateral accelerations, resulting in more unconstrained lateral behaviour. Details for both baselines are provided in the Appendix \ref{subapp:unif}. These baselines provide controlled examples of increasingly unconstrained driving behaviour and allow us to examine the relationship between behavioural diversity and downstream RL policy performance.

\subsubsection{\textbf{Data-driven Driving Models}}
\label{subsubsec:dataDrivenDrivingModels}

We consider three data-driven models with different modelling capabilities. First, we employ Behaviour Cloning (BC) using a multi-layer perceptron (MLP), which serves as a deterministic baseline that directly maps the observed driving state to the vehicle's control action. Second, we use a Mixture Density Network (MDN)~\cite{bishop1994}, allowing the model to represent multimodal action distributions and capture uncertainty in human driving behaviour. Finally, we develop an attention-based model inspired by Trajectron++~\cite{salzmann_TrajectronDynamicallyFeasibleTrajectory_2020a}, which incorporates temporal and interaction information to provide a more expressive representation of the driving context.

All three models receive the same ego-centric state representation, consisting of the ego vehicle's motion state, lane information, and the relative states of up to seven surrounding vehicles. The models predict continuous longitudinal and lateral accelerations, which are used to control the simulated vehicle. Detailed network architectures and the complete input representation are provided in the Appendix \ref{subapp:mlp-highway}, \ref{subapp:mdn-highway} and \ref{subapp:attention-highway}.

\subsubsection{\textbf{Rule-based Giver Models}}

We consider two rule-based models for the handover domain: a Minimum Jerk model and an Elliptical model. The Minimum Jerk model~\cite{flash_CoordinationArmMovements_1985} generates smooth, straight-line movements between sampled start and end positions under a minimum-jerk assumption. The Elliptical model is adapted from~\cite{sheikholeslami_StudyReachingMotions_2020} and represents giver movements as segments of elliptical trajectories embedded within an approximately planar subspace of the three-dimensional workspace.

Both rule-based models are non-reactive, and their trajectory parameters, including the initial and goal positions and movement duration, are estimated from the human demonstrations and sampled during trajectory generation to capture inter-demonstration variability. Detailed model formulations and fitting procedures are provided in the Appendix \ref{subapp:minijerk} and \ref{subapp:elliptical}.

\subsubsection{\textbf{Data-driven Giver Models}}

We first consider a non-reactive probabilistic Dynamic Movement Primitive (DMP) model~\cite{ijspeert_DynamicalMovementPrimitives_2013}, which learns a movement primitive from each human demonstration and models the resulting DMP weights probabilistically. Reactive MLP, MDN, and attention-based models are also considered for the handover domain. Each model predicts the giver's next three-dimensional hand displacement from a short history of the giver's and taker's hand positions. Specifically, the input consists of five consecutive frames of the standardised three-dimensional positions of both hands. Detailed formulation, network architectures and input representations are provided in the Appendix \ref{subapp:dmp} (DMP), \ref{subapp:mlp-handover}, \ref{subapp:mdn-handover} and \ref{subapp:attention-handover}. Table~\ref{tab:behaviour_models} summarises the human behaviour models considered in this study.

\begin{table}[t]
    \centering
    \caption{Human behaviour models considered in the highway driving and object handover domains.}
    \label{tab:behaviour_models}
    \begin{tabular}{lll}
        \toprule
        \textbf{Domain} & \textbf{Type} & \textbf{Model} \\
        \midrule
        \multirow{8}{*}{Highway driving}
            & \multirow{3}{*}{Rule-based}
                & MOBIL+Gipps \\
            & & MOBIL+Krau{\ss} \\
            & & MOBIL+IDM \\
        \cmidrule(lr){2-3}
            & \multirow{3}{*}{Data-driven}
                & MLP \\
            & & MDN \\
            & & Attention \\
        \cmidrule(lr){2-3}
            & \multirow{2}{*}{Stochastic}
                & Unif \\
            & & Unif-W \\
        \midrule
        \multirow{6}{*}{Object handover}
            & \multirow{2}{*}{Rule-based}
                & Minimum Jerk \\
            & & Elliptical \\
        \cmidrule(lr){2-3}
            & \multirow{3}{*}{Data-driven}
                & DMP \\
            & & MLP \\
            & & MDN \\
            & & Attention \\
        \bottomrule
    \end{tabular}
\end{table}

\subsection{Trajectory Collection}\label{subsec:trajectoryCollection}

We generate trajectories from each calibrated human behaviour model in order to evaluate their behavioural characteristics. To ensure a consistent comparison, all models within each domain are simulated under the same environment and initialisation procedure. Trajectories are generated separately for the highway and handover domains, using the models trained (or fit) on the corresponding training subsets $\{H_i^{\mathrm{d,train}}\}$.

For the highway domain, we simulate vehicles in a three-lane highway environment and collect trajectories from each of the eight driving models. For the handover domain, we simulate the customised handover environment and collect trajectories from each of the six giver models. Detailed simulation settings, trajectory initialisation procedures, and the number and duration of generated trajectories are provided in the Appendix \ref{app:trajectory_generation_and_collection}.

We observed that the calibrated data-driven highway driving models exhibit substantially higher crash and drive-off rates than the rule-based models, as shown in Table~\ref{tab:crash_n_drive_off}. In particular, the MLP, MDN, and Attention models generate considerably more crashes and off-road behaviours than Gipps, Krau{\ss}, and IDM. This suggests that these are poor human behaviour models, as they generate behaviour characteristics rarely seen in real-world driving.

\begin{table}
\centering
\caption{Crash rate and drive-off rate of the human driving models.}
\label{tab:crash_n_drive_off}
\begin{tabular}{lrr}
\toprule
Model & Crash Rate (\%) & Drive Off Rate (\%) \\
\midrule
Gipps & 0.00 & 0.00 \\
Krau{\ss} & 0.00 & 0.00 \\
IDM & 0.06 & 0.00 \\
MLP & 63.13 & 75.36 \\
MDN & 49.35 & 41.17 \\
Attention & 44.60 & 4.99 \\
\midrule
Unif & 90.75 & 52.24 \\
Unif-W & 98.17 & 87.16 \\
\bottomrule
\end{tabular}
\end{table}

\begin{figure*}[t]
    \centering
    \resizebox{0.75\textwidth}{!}{%
        \begin{minipage}{\textwidth}

            \begin{minipage}[c]{0.24\textwidth}
                \centering
                \subfloat[Human]{
                    \includegraphics[width=\linewidth,
                    trim={20cm 10cm 8cm 0cm},clip]{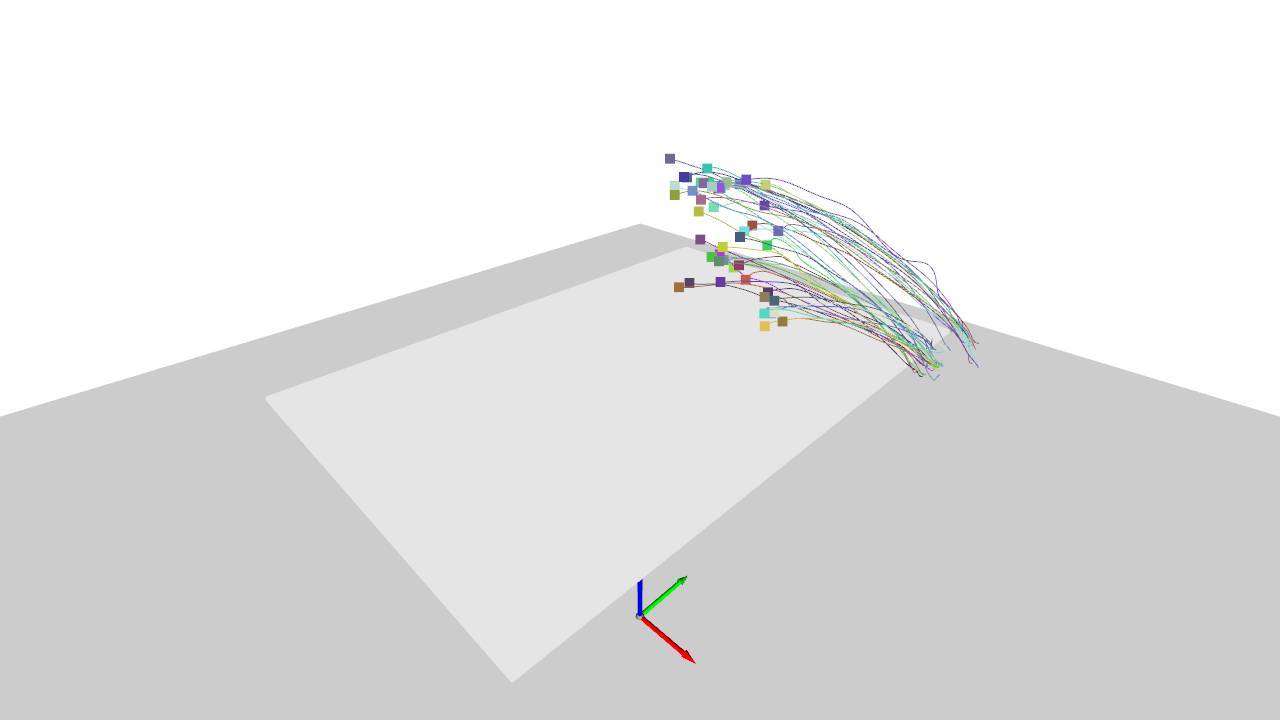}
                }
            \end{minipage}
            \hfill
            %
            \begin{minipage}[c]{0.73\textwidth}

                \centering
                \subfloat[Minimum Jerk]{
                    \includegraphics[width=0.31\linewidth,
                    trim={20cm 10cm 8cm 0cm},clip]{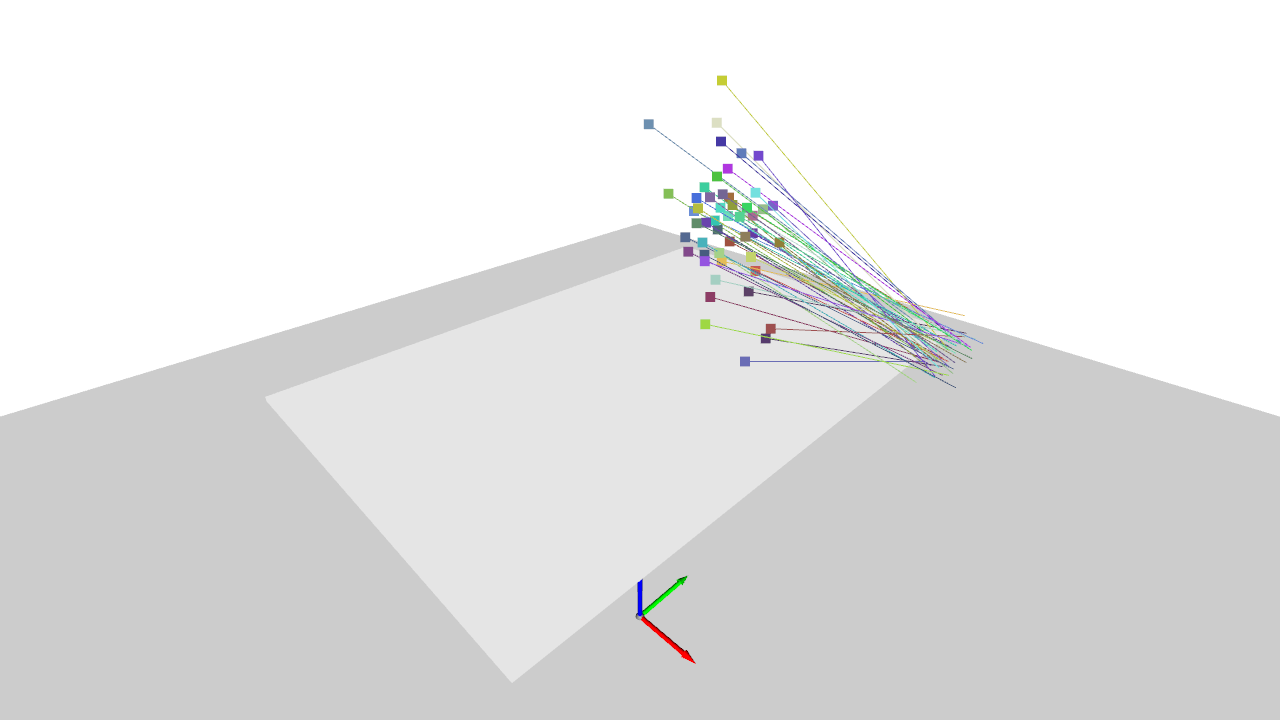}
                }
                \hfill
                \subfloat[Elliptical]{
                    \includegraphics[width=0.31\linewidth,
                    trim={20cm 10cm 8cm 0cm},clip]{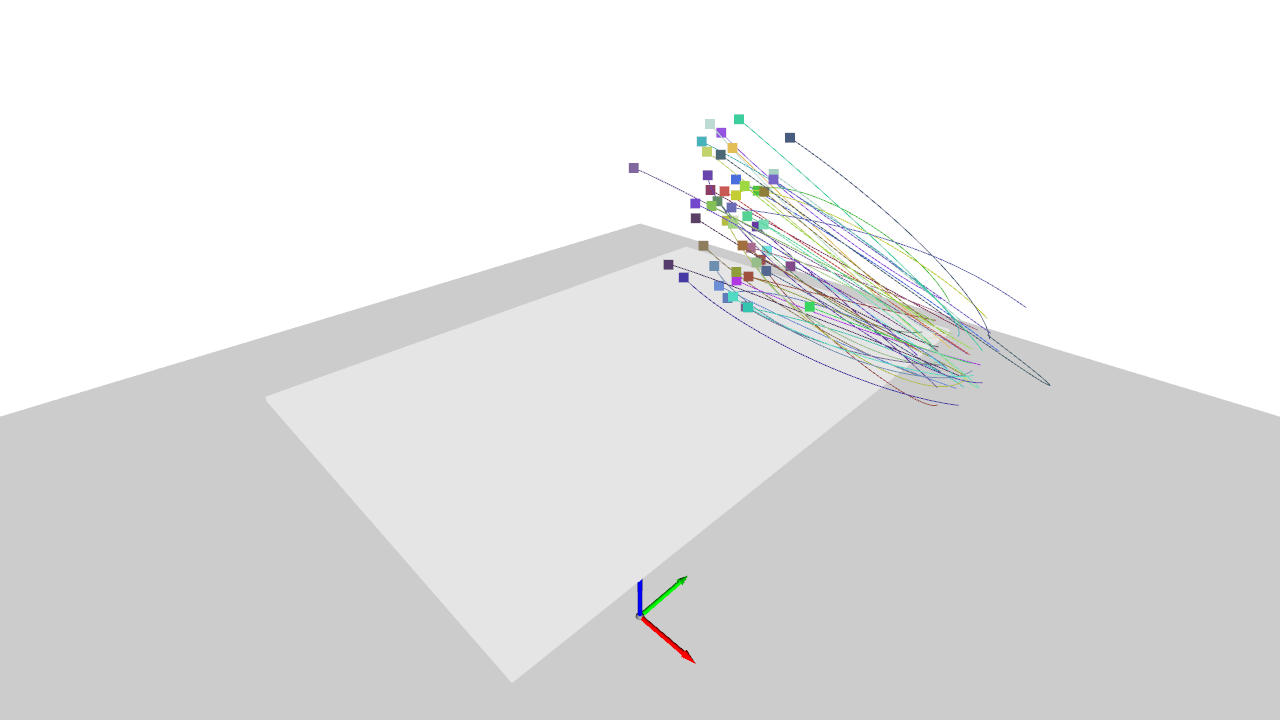}
                }
                \hfill
                \subfloat[DMP]{
                    \includegraphics[width=0.31\linewidth,
                    trim={20cm 10cm 8cm 0cm},clip]{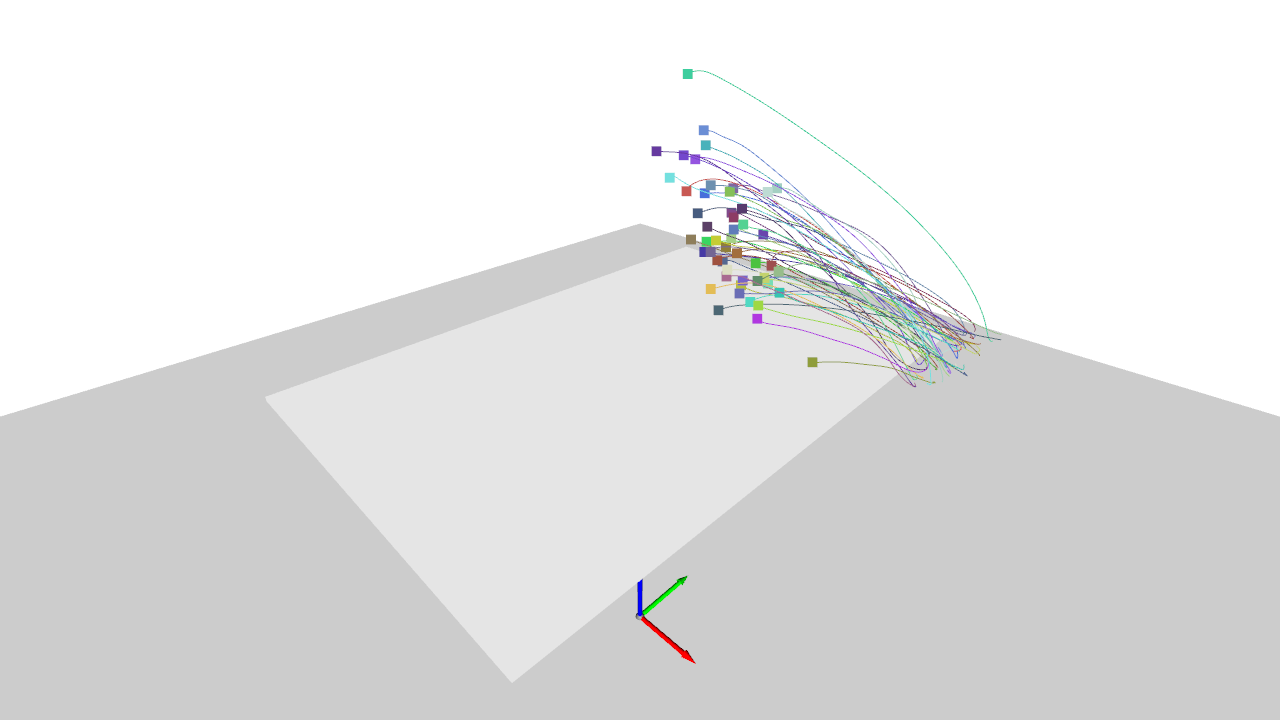}
                }

                \vspace{0.1em}

                \centering
                \subfloat[MLP]{
                    \includegraphics[width=0.31\linewidth,
                    trim={20cm 10cm 8cm 0cm},clip]{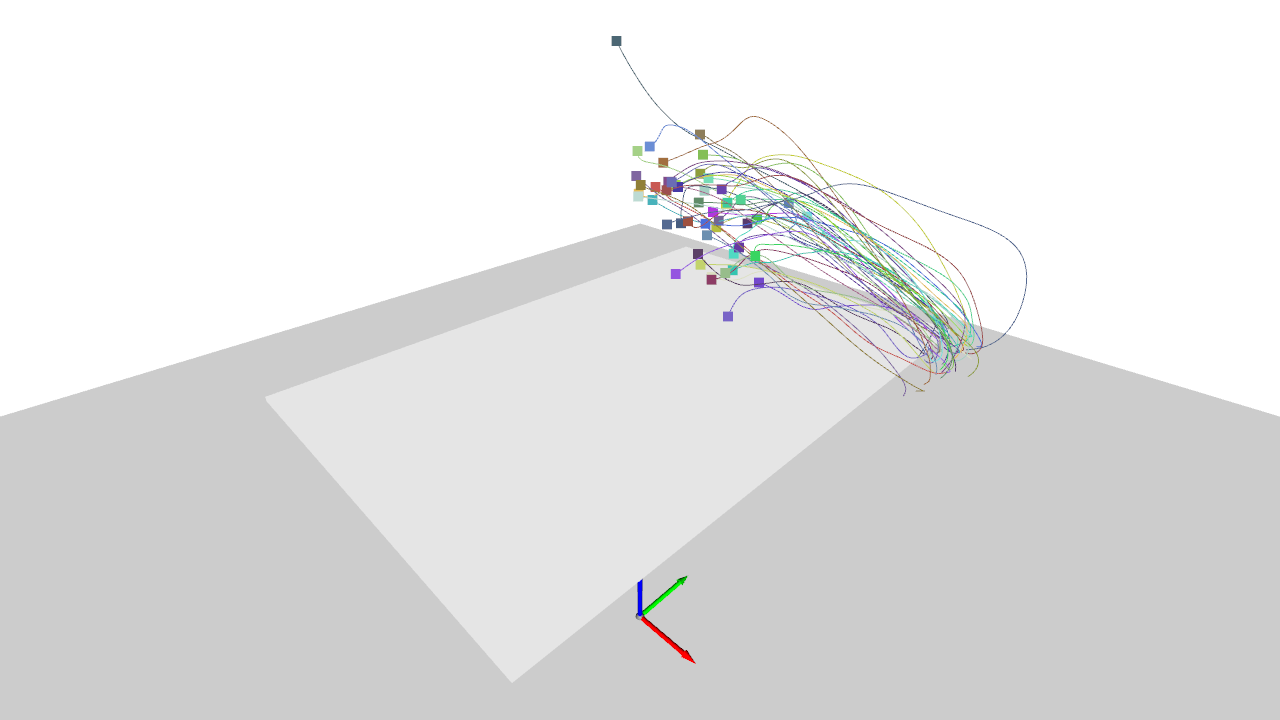}
                }
                \hfill
                \subfloat[MDN]{
                    \includegraphics[width=0.31\linewidth,
                    trim={20cm 10cm 8cm 0cm},clip]{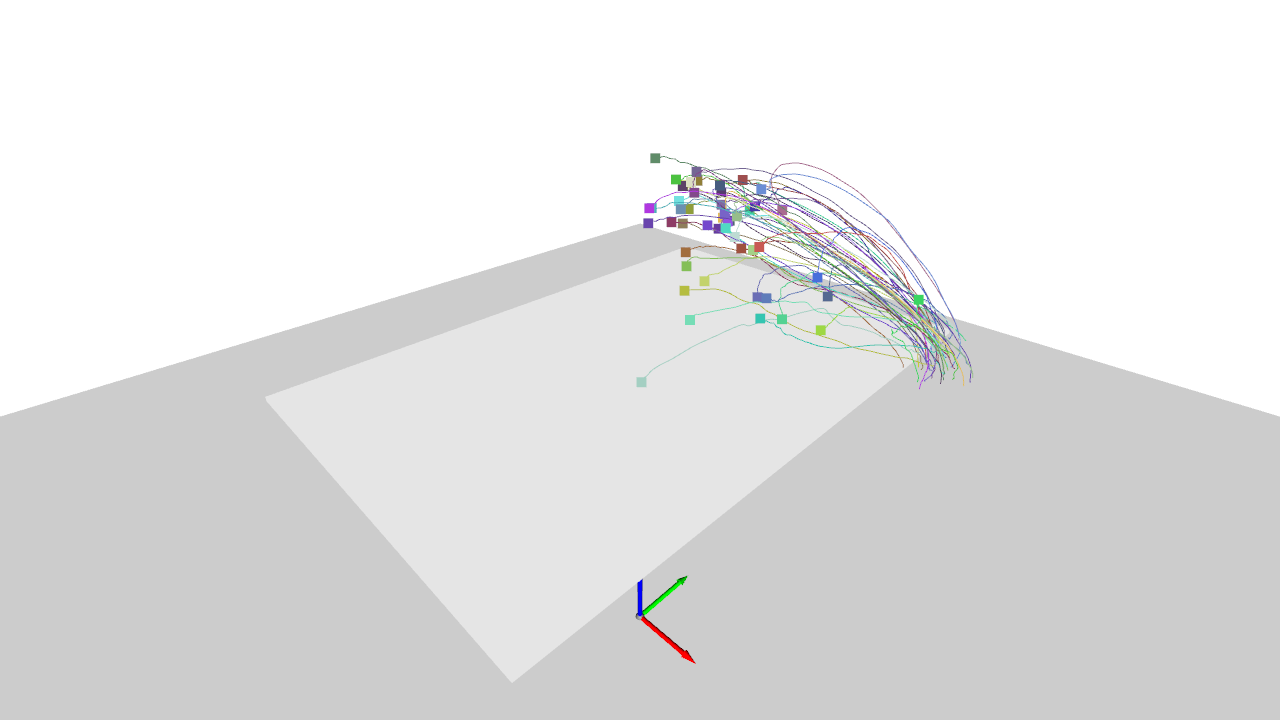}
                }
                \hfill
                \subfloat[Attention]{
                    \includegraphics[width=0.31\linewidth,
                    trim={20cm 10cm 8cm 0cm},clip]{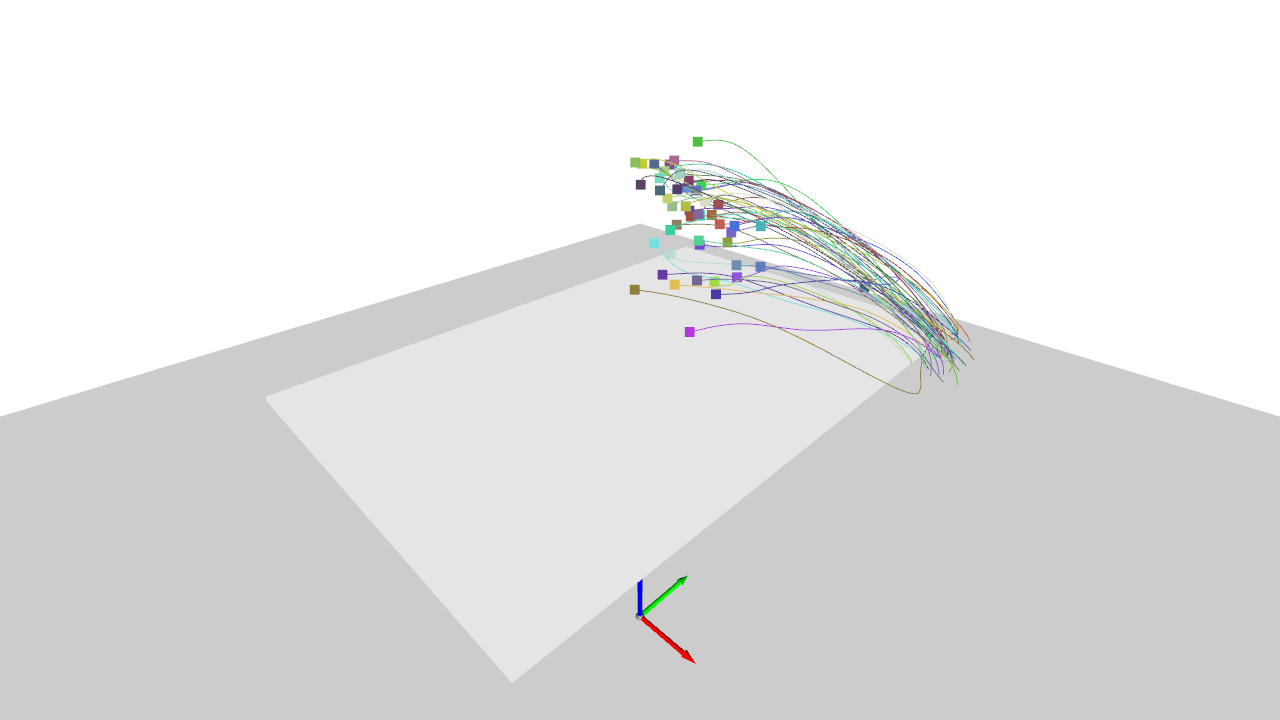}
                }

            \end{minipage}
        \end{minipage}%
    }

    \caption{50 randomly sampled handover trajectories from human demonstrations
    and trajectories generated by the human behaviour models.}
    \label{fig:handover_trajectories}
\end{figure*}

We provide simulation recordings of the calibrated highway driving models on the project page. For the handover domain, we show 50 randomly sampled trajectories generated by each of the calibrated giver models in Figure~\ref{fig:handover_trajectories}. In the highway domain, the Gipps, Krau{\ss}, and IDM models appear qualitatively more similar to the naturalistic driving behaviours observed in the highD dataset than the other models (see project page video). Similarly, in the handover domain, the DMP and Attention models appear qualitatively more similar to human handover behaviour than the remaining models. Further validation of the highway models and handover giver models is provided in Appendix \ref{subapp:model_validation_highway} and \ref{subapp:model_validation_handover}, respectively.

\subsection{Trajectory Representations}\label{subsec:trajectoryRepresentations}

Existing generative model evaluation metrics are typically computed in a feature space rather than directly on raw trajectories~\cite{Ali2025}. The choice of trajectory representation can therefore affect evaluation scores and, consequently, the assessment of human behaviour models. We consider three representations: (1) a PCA-based representation obtained from the principal components of human trajectories, (2) a learned latent representation obtained from an autoencoder trained on human trajectories, and (3) manually extracted behavioural features. The first two representations are used to compute distributional evaluation metrics, while the manually extracted features are compared using KL divergence.

For the PCA representation, we retain the first $N$ principal components for each domain. For the learned representation, we use an autoencoder trained exclusively on human trajectories and extract its latent representation for both human and generated trajectories. The manually extracted features comprise speed, acceleration, distance headway, and relative preceding speed for highway driving, and acceleration, speed, movement smoothness, curvature, trajectory end-to-end length, and path length for object handover.

We additionally vary the representation configurations, including the number of PCA components, autoencoder architecture, and KDE sample size, to assess the sensitivity of evaluation metrics to the choice of representation. These sensitivity analyses are presented in Section~\ref{subsec:metricsDoNotPredictUtility}, with the full representation and preprocessing details provided in the Appendix \ref{app:trajectory_representations}.

\subsection{Evaluation Metrics}\label{subsec:evaluationMetrics}

We evaluate the human behaviour models using two categories of metrics: representation-based metrics applied to the PCA and latent representations, and feature-wise metrics applied to the manually extracted behavioural features.

\subsubsection{\textbf{Representation-Based Metrics}}

Both the PCA-based and latent representations transform the original high-dimensional trajectories into lower-dimensional feature spaces, in which the individual feature dimensions may not have explicit semantic or physical interpretations. We therefore employ distributional metrics that characterise the fidelity and diversity of the generated feature distributions relative to those of the human data, as introduced in Section~\ref{sec:problemFormulationAndOverview}.

\subsubsection{\textbf{Feature-Wise Metrics}}

Unlike PCA-based and latent representations, manually extracted behavioural features have explicit physical interpretations. We therefore compare the distributions of individual behavioural features between the human data and the trajectories generated by each human behaviour model. Specifically, we use the Kullback-Leibler (KL) divergence to quantify the difference between the human and generated feature distributions. A lower KL divergence indicates that the generated distribution more closely matches the corresponding human distribution for that feature. We compute the KL divergence separately for each manually extracted feature and additionally calculate the average KL divergence across all features as an aggregate measure of distributional similarity. Details of the evaluation metrics are provided in the Appendix \ref{subapp:kl}.

Table~\ref{tab:evaluation_metrics} summarises the evaluation representations and metrics used for the human behaviour models in each domain. It has been noted that representation-based evaluation metrics can be sensitive to the parameters of the feature encoder~\cite{Ali2025}. To investigate the extent to which the hyperparameters of representation methods affect the relationship between evaluation metrics and downstream policy performance, we vary, for latent representations, two autoencoder hyperparameters: the latent dimension, with values in $\{16, 32, 64\}$, and the hidden dimension of the MLP layers, with values in $\{64, 128, 256\}$. For the PCA representation, we vary the number of retained principal components over $\{2, 3, 4, 5, 6, 8, 10\}$ for the highway domain and over $\{2, 3, 4, 5, 6\}$ for the handover domain. For the manually extracted features, we vary the KDE sample size over $\{1000, 2500, 5000, 10000, 20000, 40000\}$ for the highway domain. Since the total sample size in the handover domain is comparatively small, we instead use all available samples without performing a sweep.

\begin{table}[t]
    \centering
    \caption{Evaluation representations (Repr.) and metrics used for the human behaviour models in each domain.}
    \label{tab:evaluation_metrics}
    \setlength{\tabcolsep}{3pt}
    \begin{tabular}{@{}lll@{}}
        \toprule
        \textbf{Domain} & \textbf{Repr.} & \textbf{Metrics} \\
        \midrule
        \multirow{3}{*}{\makecell{Highway driving\\(8 models)}}
            & PCA
            & \multirow{2}{*}{FD, density, coverage, $APD_{\mathrm{gap}}$, $MMS_{\mathrm{gap}}$} \\
            & Latent & \\
        \cmidrule(lr){2-3}
            & Manual & 4 feature KL + averaged KL \\
        \midrule
        \multirow{3}{*}{\makecell{Object handover\\(6 models)}}
            & PCA
            & \multirow{2}{*}{FD, density, coverage, $APD_{\mathrm{gap}}$, $MMS_{\mathrm{gap}}$} \\
            & Latent & \\
        \cmidrule(lr){2-3}
            & Manual & 6 feature KL + averaged KL \\
        \bottomrule
    \end{tabular}
\end{table}

\begin{table*}[htbp]
\centering
\caption{Evaluation metrics (mean $\pm$ std over configurations) by domain, model, and representation. Within each domain, the best Latent and Manual values are in bold and the best PCA values are underlined. Empty cells mark metrics that do not apply to a representation.}
\label{tab:metric_summary}
\footnotesize
\setlength{\tabcolsep}{2.5pt}
\resizebox{\textwidth}{!}{%
\begin{tabular}{lllcccccc}
\toprule
\multirow{2}{*}{Domain} & \multirow{2}{*}{Model} & \multirow{2}{*}{Repr.} & \multicolumn{3}{c}{Fidelity} & \multicolumn{3}{c}{Diversity} \\
\cmidrule(lr){4-6} \cmidrule(lr){7-9}
 & & & FD $\downarrow$ & Density $\uparrow$ & KL $\downarrow$ & Coverage $\uparrow$ & $APD_{\mathrm{gap}}$ $\downarrow$ & $MMS_{\mathrm{gap}}$ $\downarrow$ \\
\midrule
\multirow{24}{*}{\rotatebox{90}{Highway}} & \multirow{3}{*}{Gipps} & Latent & $25.24$ {\scriptsize$\pm 20.61$} & $1.73\mathrm{e}{-3}$ {\scriptsize$\pm 3.28\mathrm{e}{-3}$} &  & $2.81\mathrm{e}{-3}$ {\scriptsize$\pm 4.74\mathrm{e}{-3}$} & {\boldmath$2.65$} {\scriptsize$\pm 1.95$} & $2.80$ {\scriptsize$\pm 1.47$} \\
 &  & PCA & \underline{$88.63$} {\scriptsize$\pm 77.15$} & $0.15$ {\scriptsize$\pm 0.18$} &  & $0.08$ {\scriptsize$\pm 0.12$} & $5.67$ {\scriptsize$\pm 1.66$} & \underline{$4.74$} {\scriptsize$\pm 3.42$} \\
 &  & Manual &  &  & $0.23$ {\scriptsize$\pm 0.01$} &  &  &  \\
\cmidrule(lr){2-9}
 & \multirow{3}{*}{Krauss} & Latent & $25.35$ {\scriptsize$\pm 20.70$} & $1.96\mathrm{e}{-3}$ {\scriptsize$\pm 3.87\mathrm{e}{-3}$} &  & $2.93\mathrm{e}{-3}$ {\scriptsize$\pm 4.55\mathrm{e}{-3}$} & $2.66$ {\scriptsize$\pm 1.96$} & {\boldmath$2.80$} {\scriptsize$\pm 1.47$} \\
 &  & PCA & $90.22$ {\scriptsize$\pm 77.74$} & $0.14$ {\scriptsize$\pm 0.18$} &  & $0.07$ {\scriptsize$\pm 0.12$} & \underline{$5.64$} {\scriptsize$\pm 1.63$} & $4.82$ {\scriptsize$\pm 3.41$} \\
 &  & Manual &  &  & $0.29$ {\scriptsize$\pm 0.03$} &  &  &  \\
\cmidrule(lr){2-9}
 & \multirow{3}{*}{IDM} & Latent & $27.73$ {\scriptsize$\pm 22.25$} & $3.00\mathrm{e}{-3}$ {\scriptsize$\pm 5.22\mathrm{e}{-3}$} &  & $3.44\mathrm{e}{-3}$ {\scriptsize$\pm 5.45\mathrm{e}{-3}$} & $3.10$ {\scriptsize$\pm 2.15$} & $3.00$ {\scriptsize$\pm 1.54$} \\
 &  & PCA & $7.73\mathrm{e}{3}$ {\scriptsize$\pm 5.93\mathrm{e}{3}$} & $0.14$ {\scriptsize$\pm 0.17$} &  & $0.06$ {\scriptsize$\pm 0.09$} & $32.84$ {\scriptsize$\pm 16.08$} & $18.54$ {\scriptsize$\pm 10.65$} \\
 &  & Manual &  &  & $0.31$ {\scriptsize$\pm 0.02$} &  &  &  \\
\cmidrule(lr){2-9}
 & \multirow{3}{*}{MLP} & Latent & $27.08$ {\scriptsize$\pm 23.72$} & $7.86\mathrm{e}{-3}$ {\scriptsize$\pm 0.01$} &  & {\boldmath$0.01$} {\scriptsize$\pm 0.02$} & $4.08$ {\scriptsize$\pm 2.50$} & $3.36$ {\scriptsize$\pm 1.77$} \\
 &  & PCA & $4.42\mathrm{e}{3}$ {\scriptsize$\pm 668.84$} & $0.12$ {\scriptsize$\pm 0.14$} &  & $0.14$ {\scriptsize$\pm 0.12$} & $51.74$ {\scriptsize$\pm 5.08$} & $26.57$ {\scriptsize$\pm 5.15$} \\
 &  & Manual &  &  & {\boldmath$0.03$} {\scriptsize$\pm 2.00\mathrm{e}{-3}$} &  &  &  \\
\cmidrule(lr){2-9}
 & \multirow{3}{*}{MDN} & Latent & {\boldmath$24.42$} {\scriptsize$\pm 21.60$} & $6.87\mathrm{e}{-3}$ {\scriptsize$\pm 0.01$} &  & $8.93\mathrm{e}{-3}$ {\scriptsize$\pm 0.01$} & $2.72$ {\scriptsize$\pm 1.98$} & $2.84$ {\scriptsize$\pm 1.55$} \\
 &  & PCA & $120.86$ {\scriptsize$\pm 69.02$} & \underline{$0.18$} {\scriptsize$\pm 0.20$} &  & \underline{$0.17$} {\scriptsize$\pm 0.17$} & $10.44$ {\scriptsize$\pm 1.74$} & $5.36$ {\scriptsize$\pm 2.96$} \\
 &  & Manual &  &  & $0.10$ {\scriptsize$\pm 0.02$} &  &  &  \\
\cmidrule(lr){2-9}
 & \multirow{3}{*}{Attention} & Latent & $25.05$ {\scriptsize$\pm 20.50$} & $2.00\mathrm{e}{-3}$ {\scriptsize$\pm 3.72\mathrm{e}{-3}$} &  & $2.96\mathrm{e}{-3}$ {\scriptsize$\pm 4.27\mathrm{e}{-3}$} & $2.78$ {\scriptsize$\pm 1.97$} & $2.83$ {\scriptsize$\pm 1.48$} \\
 &  & PCA & $90.08$ {\scriptsize$\pm 77.70$} & $0.15$ {\scriptsize$\pm 0.20$} &  & $0.13$ {\scriptsize$\pm 0.19$} & $6.09$ {\scriptsize$\pm 1.73$} & $4.84$ {\scriptsize$\pm 3.45$} \\
 &  & Manual &  &  & $0.15$ {\scriptsize$\pm 0.03$} &  &  &  \\
\cmidrule(lr){2-9}
 & \multirow{3}{*}{Unif} & Latent & $47.92$ {\scriptsize$\pm 53.09$} & $7.73\mathrm{e}{-3}$ {\scriptsize$\pm 0.01$} &  & $9.52\mathrm{e}{-3}$ {\scriptsize$\pm 0.02$} & $6.12$ {\scriptsize$\pm 4.18$} & $4.67$ {\scriptsize$\pm 2.88$} \\
 &  & PCA & $8.22\mathrm{e}{6}$ {\scriptsize$\pm 8.42\mathrm{e}{5}$} & $0.05$ {\scriptsize$\pm 0.07$} &  & $0.07$ {\scriptsize$\pm 0.08$} & $2.13\mathrm{e}{3}$ {\scriptsize$\pm 159.07$} & $1.20\mathrm{e}{3}$ {\scriptsize$\pm 100.75$} \\
 &  & Manual &  &  & $0.05$ {\scriptsize$\pm 3.57\mathrm{e}{-3}$} &  &  &  \\
\cmidrule(lr){2-9}
 & \multirow{3}{*}{Unif-W} & Latent & $53.43$ {\scriptsize$\pm 60.98$} & {\boldmath$8.01\mathrm{e}{-3}$} {\scriptsize$\pm 0.01$} &  & $0.01$ {\scriptsize$\pm 0.02$} & $6.41$ {\scriptsize$\pm 4.48$} & $4.95$ {\scriptsize$\pm 3.17$} \\
 &  & PCA & $1.15\mathrm{e}{8}$ {\scriptsize$\pm 1.06\mathrm{e}{7}$} & $0.04$ {\scriptsize$\pm 0.04$} &  & $0.06$ {\scriptsize$\pm 0.05$} & $8.31\mathrm{e}{3}$ {\scriptsize$\pm 596.98$} & $4.76\mathrm{e}{3}$ {\scriptsize$\pm 350.27$} \\
 &  & Manual &  &  & $0.06$ {\scriptsize$\pm 1.90\mathrm{e}{-3}$} &  &  &  \\
\midrule
\multirow{18}{*}{\rotatebox{90}{Handover}} & \multirow{3}{*}{Minijerk} & Latent & $1.23$ {\scriptsize$\pm 0.32$} & $0.07$ {\scriptsize$\pm 0.02$} &  & $0.05$ {\scriptsize$\pm 0.02$} & $0.35$ {\scriptsize$\pm 0.08$} & $0.33$ {\scriptsize$\pm 0.05$} \\
 &  & PCA & $3.42\mathrm{e}{-3}$ {\scriptsize$\pm 2.28\mathrm{e}{-3}$} & $2.05$ {\scriptsize$\pm 0.17$} &  & $0.59$ {\scriptsize$\pm 0.20$} & $0.01$ {\scriptsize$\pm 1.72\mathrm{e}{-3}$} & \underline{$5.16\mathrm{e}{-3}$} {\scriptsize$\pm 4.50\mathrm{e}{-3}$} \\
 &  & Manual &  &  & $2.37$ &  &  &  \\
\cmidrule(lr){2-9}
 & \multirow{3}{*}{Elliptical} & Latent & $0.90$ {\scriptsize$\pm 0.24$} & $0.11$ {\scriptsize$\pm 0.02$} &  & $0.14$ {\scriptsize$\pm 0.05$} & $0.22$ {\scriptsize$\pm 0.07$} & $0.36$ {\scriptsize$\pm 0.05$} \\
 &  & PCA & $2.26\mathrm{e}{-3}$ {\scriptsize$\pm 1.38\mathrm{e}{-3}$} & \underline{$2.14$} {\scriptsize$\pm 0.17$} &  & $0.73$ {\scriptsize$\pm 0.14$} & \underline{$1.51\mathrm{e}{-3}$} {\scriptsize$\pm 8.38\mathrm{e}{-4}$} & $9.59\mathrm{e}{-3}$ {\scriptsize$\pm 6.39\mathrm{e}{-3}$} \\
 &  & Manual &  &  & $0.67$ &  &  &  \\
\cmidrule(lr){2-9}
 & \multirow{3}{*}{DMP} & Latent & {\boldmath$0.18$} {\scriptsize$\pm 0.07$} & $0.29$ {\scriptsize$\pm 0.02$} &  & {\boldmath$0.46$} {\scriptsize$\pm 0.03$} & $0.14$ {\scriptsize$\pm 0.03$} & {\boldmath$0.28$} {\scriptsize$\pm 0.05$} \\
 &  & PCA & \underline{$1.80\mathrm{e}{-4}$} {\scriptsize$\pm 1.17\mathrm{e}{-4}$} & $1.93$ {\scriptsize$\pm 0.15$} &  & \underline{$0.89$} {\scriptsize$\pm 0.06$} & $4.72\mathrm{e}{-3}$ {\scriptsize$\pm 1.04\mathrm{e}{-3}$} & $6.86\mathrm{e}{-3}$ {\scriptsize$\pm 5.10\mathrm{e}{-3}$} \\
 &  & Manual &  &  & {\boldmath$0.50$} &  &  &  \\
\cmidrule(lr){2-9}
 & \multirow{3}{*}{MLP} & Latent & $0.60$ {\scriptsize$\pm 0.15$} & $0.19$ {\scriptsize$\pm 0.02$} &  & $0.27$ {\scriptsize$\pm 0.04$} & $0.27$ {\scriptsize$\pm 0.08$} & $0.42$ {\scriptsize$\pm 0.05$} \\
 &  & PCA & $9.91\mathrm{e}{-3}$ {\scriptsize$\pm 8.76\mathrm{e}{-4}$} & $1.62$ {\scriptsize$\pm 0.17$} &  & $0.81$ {\scriptsize$\pm 0.11$} & $7.35\mathrm{e}{-3}$ {\scriptsize$\pm 4.85\mathrm{e}{-3}$} & $0.01$ {\scriptsize$\pm 8.73\mathrm{e}{-3}$} \\
 &  & Manual &  &  & $1.49$ &  &  &  \\
\cmidrule(lr){2-9}
 & \multirow{3}{*}{MDN} & Latent & $0.56$ {\scriptsize$\pm 0.20$} & $0.35$ {\scriptsize$\pm 0.04$} &  & $0.39$ {\scriptsize$\pm 0.04$} & $0.11$ {\scriptsize$\pm 0.07$} & $0.29$ {\scriptsize$\pm 0.05$} \\
 &  & PCA & $0.01$ {\scriptsize$\pm 6.05\mathrm{e}{-4}$} & $1.72$ {\scriptsize$\pm 0.11$} &  & $0.85$ {\scriptsize$\pm 0.06$} & $0.02$ {\scriptsize$\pm 3.10\mathrm{e}{-3}$} & $6.74\mathrm{e}{-3}$ {\scriptsize$\pm 4.73\mathrm{e}{-3}$} \\
 &  & Manual &  &  & $2.10$ &  &  &  \\
\cmidrule(lr){2-9}
 & \multirow{3}{*}{Attention} & Latent & $0.65$ {\scriptsize$\pm 0.21$} & {\boldmath$0.38$} {\scriptsize$\pm 0.05$} &  & $0.30$ {\scriptsize$\pm 0.05$} & {\boldmath$0.10$} {\scriptsize$\pm 0.07$} & $0.29$ {\scriptsize$\pm 0.05$} \\
 &  & PCA & $0.01$ {\scriptsize$\pm 8.96\mathrm{e}{-4}$} & $1.74$ {\scriptsize$\pm 0.12$} &  & $0.81$ {\scriptsize$\pm 0.08$} & $0.01$ {\scriptsize$\pm 3.66\mathrm{e}{-3}$} & $6.94\mathrm{e}{-3}$ {\scriptsize$\pm 6.02\mathrm{e}{-3}$} \\
 &  & Manual &  &  & $1.57$ &  &  &  \\
\bottomrule
\end{tabular}}
\end{table*}

\subsection{Experimental Procedure}\label{subsec:experimentalProcedure}

In the highway domain, we train eight driving policies $\{\pi_i^{\mathrm{HW}}\}$, each against a single driving model from $\{H_i^{\mathrm{HW,train}}\}$. Each training configuration is repeated using three random seeds for 40{,}000 environment steps. In the handover domain, we train six taker policies $\{\pi_i^{\mathrm{HO}}\}$, each against a single giver model from $\{H_i^{\mathrm{HO,train}}\}$. Each training configuration is repeated using three random seeds for 100{,}000 training steps. Within each domain, the same set of RL training hyperparameters is used across all training configurations. 

Each trained policy is then cross-evaluated against every corresponding evaluation model in ${H_j^{d,\mathrm{eval}}}$ within its respective domain. Rather than evaluating a policy only against the the same type of behaviour model used during training, cross-evaluation assesses its robustness to variations in human behaviour. This is important for real-world HRI and autonomous driving, where human behaviour is inherently variable and may differ substantially from the training distribution. Evaluating only within the training distribution could therefore overestimate practical utility, whereas cross-evaluation provides a more stringent assessment of whether a policy remains effective under different, independently trained behaviour models. 

Each policy-model pair is evaluated over 100 episodes in the highway domain and 200 episodes in the handover domain. The average reward obtained by policy $\pi_i^{d}$ when evaluated against model $H_j^{d,\mathrm{eval}}$ is denoted by
\begin{equation}
    R_{ij}^{d}
    =
    R\left(
        \pi_i^{d,\mathrm{train}},
        H_j^{d,\mathrm{eval}}
    \right),
    \qquad d\in\{\mathrm{HW},\mathrm{HO}\}.
\end{equation}
We then use the average cross-evaluation reward as the overall downstream performance of policy $\pi_i^{d}$:
\begin{equation}
    R_i^d
    =
    \frac{1}{N_d}
    \sum_{j=1}^{N_d}
    R_{ij}^{d},
    \qquad d\in\{\mathrm{HW},\mathrm{HO}\}, 
\end{equation}
where $N_d$ denotes the number of human behaviour models in domain $d$.

To investigate the relationship between model evaluation metrics listed in Table~\ref{tab:evaluation_metrics} and downstream policy performance, we compute a regret metric. For each metric in $\{\mathcal{E}_j\}_{j=1}^M$, we orient the score so that a higher value indicates a better outcome.
\looseness=-1
\begin{align*}
    q_{\mathrm{FD}}=-\mathrm{FD}&,\quad q_{\mathrm{KL}}=-\mathrm{KL}, \\
    q_{\mathrm{density}}=\mathrm{density}&,\quad q_{\mathrm{coverage}}=\mathrm{coverage}, \\
    q_{\mathrm{APD}}=-|APD_{\mathrm{gap}}|&,\quad q_{\mathrm{MMS}}=-|MMS_{\mathrm{gap}}|,
\end{align*}
Each metric selects the best policy according to its own score,
\begin{equation}
    \hat{i}_j=\arg\max_i q_{ij}^d,
\end{equation}
and we define the regret of each metric as
\begin{equation}\label{eq:regret}
    \mathrm{Reg}_j^d=\frac{\max_i R_i^d - R_{\hat{i}_j}^d}{\max_i R_i^d - \min_i R_i^d}.
\end{equation}
Hence $\mathrm{Reg}_j^d=0$ indicates that metric $\mathcal{E}_j$ selects the best training model, whereas $\mathrm{Reg}_j^d=1$ indicates that it selects the worst training model. The regret of each metric can then be compared with the regret of selecting a random training model, namely the averaged regret across all metrics, $\mathrm{Reg}_{\mathrm{avg}}^d=\frac{1}{M}\sum_{j=1}^{M}\mathrm{Reg}_j^d$. If $\mathrm{Reg}_j^d<\mathrm{Reg}_{\mathrm{avg}}^d$, using metric $\mathcal{E}_j$ as the training-model selector in domain $d$ therefore yields a lower regret than randomly selecting a training model from the model pool. An important assumption underlying this regret metric is that the utility of the models mostly differs, which is satisfied in our experiments (see Figures~\ref{fig:reward_heatmap_highway} and~\ref{fig:reward_heatmap_handover}).

\section{Results}\label{sec:results}
Table~\ref{tab:metric_summary} presents the evaluation results for each domain, model, and representation, averaged across configurations, which are discussed in Section~\ref{subsec:metricsDoNotPredictUtility}. In the highway domain, MLP and MDN achieve good fidelity scores and high diversity according to coverage. Gipps and Krau{\ss} also show good diversity according to $APD_{\mathrm{gap}}$ and $MMS_{\mathrm{gap}}$. In the handover domain, DMP performs well across both fidelity and diversity metrics. Notably, although MLP and MDN achieve good metric scores, they also have relatively high crash and drive-off rates (Table~\ref{tab:crash_n_drive_off}). This suggests that the evaluation metrics capture certain aspects of behavioural fidelity and diversity, but may not fully reflect whether the generated behaviours are realistic or physically plausible.

\subsection{Behavioural Evaluation Metrics Do Not Consistently Predict Downstream Utility}\label{subsec:metricsDoNotPredictUtility}

For each configuration, we compute the regret according to Equation~\ref{eq:regret}. The results across both domains, all representations and configurations are summarised in Table~\ref{tab:full_regret}. The results show that some metrics, under specific hyperparameter settings and representation methods, can serve as effective training-model selectors and provide useful predictions of downstream RL utility. For example, density appears to be a reliable selector in the handover domain with latent representations and in the highway domain with PCA representations. Several KL divergence metrics based on manual features also achieve lower regret than the random-selection baseline across both domains.

\begin{table*}[htbp]
\centering
\caption{Normalised regret of metric-selected models by domain, representation, and configuration. A superscript $\times$ marks a metric whose $\mathrm{Reg}_j^d\ge\mathrm{Reg}_{\mathrm{avg}}^d$.}
\label{tab:full_regret}
\begin{tabular}{lllcccccc}
\toprule
\multirow{2}{*}{Domain} & \multirow{2}{*}{Representation} & \multirow{2}{*}{Configuration} & \multicolumn{3}{c}{Fidelity} & \multicolumn{3}{c}{Diversity} \\
\cmidrule(lr){4-6} \cmidrule(lr){7-9}
 & & & FD & Density & KL & Coverage & APD gap & MMS gap \\
\midrule
\multirow{21}{*}{Highway} & \multirow{9}{*}{Latent} & Latent = 16, Hidden = 64 & \cellcolor[HTML]{006837}{0.00} & \cellcolor[HTML]{FEDA86}{0.61$^{\times}$} & \cellcolor[HTML]{000000}{} & \cellcolor[HTML]{006837}{0.00} & \cellcolor[HTML]{D3EC87}{0.39} & \cellcolor[HTML]{006837}{0.00} \\
 &  & Latent = 16, Hidden = 128 & \cellcolor[HTML]{ECF7A6}{0.45$^{\times}$} & \cellcolor[HTML]{ECF7A6}{0.45$^{\times}$} & \cellcolor[HTML]{000000}{} & \cellcolor[HTML]{006837}{0.00} & \cellcolor[HTML]{026C39}{0.01} & \cellcolor[HTML]{D3EC87}{0.39} \\
 &  & Latent = 16, Hidden = 256 & \cellcolor[HTML]{FEDA86}{0.61$^{\times}$} & \cellcolor[HTML]{FEDA86}{0.61$^{\times}$} & \cellcolor[HTML]{000000}{} & \cellcolor[HTML]{FEDA86}{0.61$^{\times}$} & \cellcolor[HTML]{026C39}{0.01} & \cellcolor[HTML]{FEDA86}{0.61$^{\times}$} \\
 &  & Latent = 32, Hidden = 64 & \cellcolor[HTML]{026C39}{0.01} & \cellcolor[HTML]{026C39}{0.01} & \cellcolor[HTML]{000000}{} & \cellcolor[HTML]{026C39}{0.01} & \cellcolor[HTML]{E44C34}{0.85$^{\times}$} & \cellcolor[HTML]{E44C34}{0.85$^{\times}$} \\
 &  & Latent = 32, Hidden = 128 & \cellcolor[HTML]{026C39}{0.01} & \cellcolor[HTML]{026C39}{0.01} & \cellcolor[HTML]{000000}{} & \cellcolor[HTML]{026C39}{0.01} & \cellcolor[HTML]{E44C34}{0.85$^{\times}$} & \cellcolor[HTML]{E44C34}{0.85$^{\times}$} \\
 &  & Latent = 32, Hidden = 256 & \cellcolor[HTML]{026C39}{0.01} & \cellcolor[HTML]{ECF7A6}{0.45$^{\times}$} & \cellcolor[HTML]{000000}{} & \cellcolor[HTML]{ECF7A6}{0.45$^{\times}$} & \cellcolor[HTML]{026C39}{0.01} & \cellcolor[HTML]{D3EC87}{0.39} \\
 &  & Latent = 64, Hidden = 64 & \cellcolor[HTML]{026C39}{0.01} & \cellcolor[HTML]{E8F59F}{0.44$^{\times}$} & \cellcolor[HTML]{000000}{} & \cellcolor[HTML]{E8F59F}{0.44$^{\times}$} & \cellcolor[HTML]{E44C34}{0.85$^{\times}$} & \cellcolor[HTML]{E44C34}{0.85$^{\times}$} \\
 &  & Latent = 64, Hidden = 128 & \cellcolor[HTML]{026C39}{0.01} & \cellcolor[HTML]{E8F59F}{0.44$^{\times}$} & \cellcolor[HTML]{000000}{} & \cellcolor[HTML]{E8F59F}{0.44$^{\times}$} & \cellcolor[HTML]{D3EC87}{0.39} & \cellcolor[HTML]{D3EC87}{0.39} \\
 &  & Latent = 64, Hidden = 256 & \cellcolor[HTML]{E44C34}{0.85$^{\times}$} & \cellcolor[HTML]{E8F59F}{0.44$^{\times}$} & \cellcolor[HTML]{000000}{} & \cellcolor[HTML]{E8F59F}{0.44$^{\times}$} & \cellcolor[HTML]{E44C34}{0.85$^{\times}$} & \cellcolor[HTML]{E44C34}{0.85$^{\times}$} \\
\cmidrule(lr){2-9}
 & \multirow{7}{*}{PCA} & N Components = 2 & \cellcolor[HTML]{E44C34}{0.85$^{\times}$} & \cellcolor[HTML]{026C39}{0.01} & \cellcolor[HTML]{000000}{} & \cellcolor[HTML]{66BD63}{0.20} & \cellcolor[HTML]{D3EC87}{0.39} & \cellcolor[HTML]{66BD63}{0.20} \\
 &  & N Components = 3 & \cellcolor[HTML]{E44C34}{0.85$^{\times}$} & \cellcolor[HTML]{026C39}{0.01} & \cellcolor[HTML]{000000}{} & \cellcolor[HTML]{026C39}{0.01} & \cellcolor[HTML]{D3EC87}{0.39} & \cellcolor[HTML]{E44C34}{0.85$^{\times}$} \\
 &  & N Components = 4 & \cellcolor[HTML]{E44C34}{0.85$^{\times}$} & \cellcolor[HTML]{E44C34}{0.85$^{\times}$} & \cellcolor[HTML]{000000}{} & \cellcolor[HTML]{026C39}{0.01} & \cellcolor[HTML]{D3EC87}{0.39} & \cellcolor[HTML]{E44C34}{0.85$^{\times}$} \\
 &  & N Components = 5 & \cellcolor[HTML]{E44C34}{0.85$^{\times}$} & \cellcolor[HTML]{026C39}{0.01} & \cellcolor[HTML]{000000}{} & \cellcolor[HTML]{006837}{0.00} & \cellcolor[HTML]{D3EC87}{0.39} & \cellcolor[HTML]{E44C34}{0.85$^{\times}$} \\
 &  & N Components = 6 & \cellcolor[HTML]{E44C34}{0.85$^{\times}$} & \cellcolor[HTML]{026C39}{0.01} & \cellcolor[HTML]{000000}{} & \cellcolor[HTML]{006837}{0.00} & \cellcolor[HTML]{D3EC87}{0.39} & \cellcolor[HTML]{E44C34}{0.85$^{\times}$} \\
 &  & N Components = 8 & \cellcolor[HTML]{E44C34}{0.85$^{\times}$} & \cellcolor[HTML]{026C39}{0.01} & \cellcolor[HTML]{000000}{} & \cellcolor[HTML]{006837}{0.00} & \cellcolor[HTML]{D3EC87}{0.39} & \cellcolor[HTML]{026C39}{0.01} \\
 &  & N Components = 10 & \cellcolor[HTML]{E44C34}{0.85$^{\times}$} & \cellcolor[HTML]{026C39}{0.01} & \cellcolor[HTML]{000000}{} & \cellcolor[HTML]{026C39}{0.01} & \cellcolor[HTML]{D3EC87}{0.39} & \cellcolor[HTML]{E44C34}{0.85$^{\times}$} \\
\cmidrule(lr){2-9}
 & \multirow{5}{*}{Manual} & Speed KL & \cellcolor[HTML]{000000}{} & \cellcolor[HTML]{000000}{} & \cellcolor[HTML]{66BD63}{0.20} & \cellcolor[HTML]{000000}{} & \cellcolor[HTML]{000000}{} & \cellcolor[HTML]{000000}{} \\
 &  & Acceleration KL & \cellcolor[HTML]{000000}{} & \cellcolor[HTML]{000000}{} & \cellcolor[HTML]{026C39}{0.01} & \cellcolor[HTML]{000000}{} & \cellcolor[HTML]{000000}{} & \cellcolor[HTML]{000000}{} \\
 &  & Distance Headway KL & \cellcolor[HTML]{000000}{} & \cellcolor[HTML]{000000}{} & \cellcolor[HTML]{006837}{0.00} & \cellcolor[HTML]{000000}{} & \cellcolor[HTML]{000000}{} & \cellcolor[HTML]{000000}{} \\
 &  & Relative Preceding Speed KL & \cellcolor[HTML]{000000}{} & \cellcolor[HTML]{000000}{} & \cellcolor[HTML]{ECF7A6}{0.45$^{\times}$} & \cellcolor[HTML]{000000}{} & \cellcolor[HTML]{000000}{} & \cellcolor[HTML]{000000}{} \\
 &  & Averaged Feature KL & \cellcolor[HTML]{000000}{} & \cellcolor[HTML]{000000}{} & \cellcolor[HTML]{006837}{0.00} & \cellcolor[HTML]{000000}{} & \cellcolor[HTML]{000000}{} & \cellcolor[HTML]{000000}{} \\
\midrule
\multirow{21}{*}{Handover} & \multirow{9}{*}{Latent} & Latent = 16, Hidden = 64 & \cellcolor[HTML]{A50026}{1.00$^{\times}$} & \cellcolor[HTML]{006837}{0.00} & \cellcolor[HTML]{000000}{} & \cellcolor[HTML]{A50026}{1.00$^{\times}$} & \cellcolor[HTML]{04703B}{0.02} & \cellcolor[HTML]{006837}{0.00} \\
 &  & Latent = 16, Hidden = 128 & \cellcolor[HTML]{A50026}{1.00$^{\times}$} & \cellcolor[HTML]{04703B}{0.02} & \cellcolor[HTML]{000000}{} & \cellcolor[HTML]{A50026}{1.00$^{\times}$} & \cellcolor[HTML]{006837}{0.00} & \cellcolor[HTML]{A50026}{1.00$^{\times}$} \\
 &  & Latent = 16, Hidden = 256 & \cellcolor[HTML]{A50026}{1.00$^{\times}$} & \cellcolor[HTML]{04703B}{0.02} & \cellcolor[HTML]{000000}{} & \cellcolor[HTML]{A50026}{1.00$^{\times}$} & \cellcolor[HTML]{04703B}{0.02} & \cellcolor[HTML]{04703B}{0.02} \\
 &  & Latent = 32, Hidden = 64 & \cellcolor[HTML]{A50026}{1.00$^{\times}$} & \cellcolor[HTML]{04703B}{0.02} & \cellcolor[HTML]{000000}{} & \cellcolor[HTML]{A50026}{1.00$^{\times}$} & \cellcolor[HTML]{006837}{0.00} & \cellcolor[HTML]{006837}{0.00} \\
 &  & Latent = 32, Hidden = 128 & \cellcolor[HTML]{A50026}{1.00$^{\times}$} & \cellcolor[HTML]{04703B}{0.02} & \cellcolor[HTML]{000000}{} & \cellcolor[HTML]{A50026}{1.00$^{\times}$} & \cellcolor[HTML]{04703B}{0.02} & \cellcolor[HTML]{A50026}{1.00$^{\times}$} \\
 &  & Latent = 32, Hidden = 256 & \cellcolor[HTML]{A50026}{1.00$^{\times}$} & \cellcolor[HTML]{04703B}{0.02} & \cellcolor[HTML]{000000}{} & \cellcolor[HTML]{A50026}{1.00$^{\times}$} & \cellcolor[HTML]{04703B}{0.02} & \cellcolor[HTML]{A50026}{1.00$^{\times}$} \\
 &  & Latent = 64, Hidden = 64 & \cellcolor[HTML]{A50026}{1.00$^{\times}$} & \cellcolor[HTML]{006837}{0.00} & \cellcolor[HTML]{000000}{} & \cellcolor[HTML]{A50026}{1.00$^{\times}$} & \cellcolor[HTML]{006837}{0.00} & \cellcolor[HTML]{A50026}{1.00$^{\times}$} \\
 &  & Latent = 64, Hidden = 128 & \cellcolor[HTML]{A50026}{1.00$^{\times}$} & \cellcolor[HTML]{04703B}{0.02} & \cellcolor[HTML]{000000}{} & \cellcolor[HTML]{A50026}{1.00$^{\times}$} & \cellcolor[HTML]{04703B}{0.02} & \cellcolor[HTML]{04703B}{0.02} \\
 &  & Latent = 64, Hidden = 256 & \cellcolor[HTML]{A50026}{1.00$^{\times}$} & \cellcolor[HTML]{04703B}{0.02} & \cellcolor[HTML]{000000}{} & \cellcolor[HTML]{006837}{0.00} & \cellcolor[HTML]{A50026}{1.00$^{\times}$} & \cellcolor[HTML]{04703B}{0.02} \\
\cmidrule(lr){2-9}
 & \multirow{5}{*}{PCA} & N Components = 2 & \cellcolor[HTML]{A50026}{1.00$^{\times}$} & \cellcolor[HTML]{128A49}{0.07} & \cellcolor[HTML]{000000}{} & \cellcolor[HTML]{A50026}{1.00$^{\times}$} & \cellcolor[HTML]{128A49}{0.07} & \cellcolor[HTML]{FDBB6C}{0.67$^{\times}$} \\
 &  & N Components = 3 & \cellcolor[HTML]{A50026}{1.00$^{\times}$} & \cellcolor[HTML]{128A49}{0.07} & \cellcolor[HTML]{000000}{} & \cellcolor[HTML]{A50026}{1.00$^{\times}$} & \cellcolor[HTML]{128A49}{0.07} & \cellcolor[HTML]{FDBB6C}{0.67$^{\times}$} \\
 &  & N Components = 4 & \cellcolor[HTML]{A50026}{1.00$^{\times}$} & \cellcolor[HTML]{128A49}{0.07} & \cellcolor[HTML]{000000}{} & \cellcolor[HTML]{A50026}{1.00$^{\times}$} & \cellcolor[HTML]{128A49}{0.07} & \cellcolor[HTML]{04703B}{0.02} \\
 &  & N Components = 5 & \cellcolor[HTML]{A50026}{1.00$^{\times}$} & \cellcolor[HTML]{128A49}{0.07} & \cellcolor[HTML]{000000}{} & \cellcolor[HTML]{A50026}{1.00$^{\times}$} & \cellcolor[HTML]{128A49}{0.07} & \cellcolor[HTML]{FDBB6C}{0.67$^{\times}$} \\
 &  & N Components = 6 & \cellcolor[HTML]{A50026}{1.00$^{\times}$} & \cellcolor[HTML]{128A49}{0.07} & \cellcolor[HTML]{000000}{} & \cellcolor[HTML]{A50026}{1.00$^{\times}$} & \cellcolor[HTML]{128A49}{0.07} & \cellcolor[HTML]{FDBB6C}{0.67$^{\times}$} \\
\cmidrule(lr){2-9}
 & \multirow{7}{*}{Manual} & Speed KL & \cellcolor[HTML]{000000}{} & \cellcolor[HTML]{000000}{} & \cellcolor[HTML]{A50026}{1.00$^{\times}$} & \cellcolor[HTML]{000000}{} & \cellcolor[HTML]{000000}{} & \cellcolor[HTML]{000000}{} \\
 &  & Acceleration KL & \cellcolor[HTML]{000000}{} & \cellcolor[HTML]{000000}{} & \cellcolor[HTML]{A50026}{1.00$^{\times}$} & \cellcolor[HTML]{000000}{} & \cellcolor[HTML]{000000}{} & \cellcolor[HTML]{000000}{} \\
 &  & End-to-End Length KL & \cellcolor[HTML]{000000}{} & \cellcolor[HTML]{000000}{} & \cellcolor[HTML]{006837}{0.00} & \cellcolor[HTML]{000000}{} & \cellcolor[HTML]{000000}{} & \cellcolor[HTML]{000000}{} \\
 &  & Movement Smoothness KL & \cellcolor[HTML]{000000}{} & \cellcolor[HTML]{000000}{} & \cellcolor[HTML]{04703B}{0.02} & \cellcolor[HTML]{000000}{} & \cellcolor[HTML]{000000}{} & \cellcolor[HTML]{000000}{} \\
 &  & Trajectory Curvature KL & \cellcolor[HTML]{000000}{} & \cellcolor[HTML]{000000}{} & \cellcolor[HTML]{006837}{0.00} & \cellcolor[HTML]{000000}{} & \cellcolor[HTML]{000000}{} & \cellcolor[HTML]{000000}{} \\
 &  & Trajectory Path Length KL & \cellcolor[HTML]{000000}{} & \cellcolor[HTML]{000000}{} & \cellcolor[HTML]{A50026}{1.00$^{\times}$} & \cellcolor[HTML]{000000}{} & \cellcolor[HTML]{000000}{} & \cellcolor[HTML]{000000}{} \\
 &  & Averaged Feature KL & \cellcolor[HTML]{000000}{} & \cellcolor[HTML]{000000}{} & \cellcolor[HTML]{A50026}{1.00$^{\times}$} & \cellcolor[HTML]{000000}{} & \cellcolor[HTML]{000000}{} & \cellcolor[HTML]{000000}{} \\
\bottomrule
\end{tabular}
\end{table*}

However, none of the evaluated metrics demonstrates robust performance across both domains and all hyperparameter configurations. For example, density performs poorly in most highway experiments with latent representations. Similarly, although several manual-feature KL metrics yield promising results, their effectiveness varies substantially across features. The choice of a suitable manual feature is highly domain-dependent, and it is difficult to determine a priori which feature will provide a reliable training-model selector without first evaluating the resulting policies. Overall, none of the metrics evaluated in this work demonstrates sufficient robustness to serve as a general-purpose training-model selector or predictor of downstream RL utility across different domains, representations, and hyperparameter configurations.

\subsection{Training on the Test Distribution Does Not Necessarily Maximise Downstream Performance}\label{subsec:trainModelDoesNotMax}

\begin{figure}
    \centering
    \captionsetup[subfloat]{font=footnotesize}
    \subfloat[Highway\label{fig:reward_heatmap_highway}]{%
        \includegraphics[width=\linewidth]{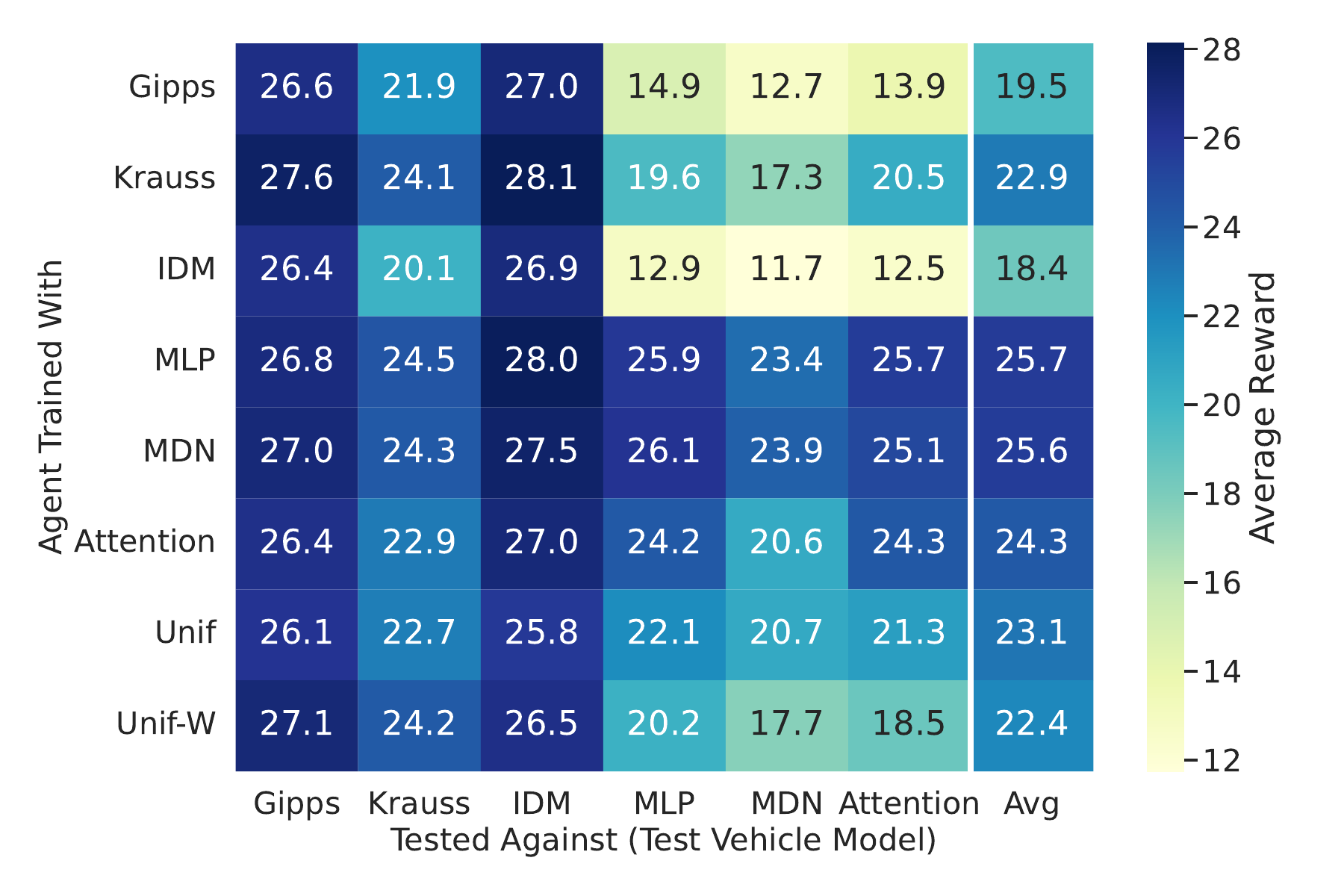}
    }\\
    \subfloat[Handover\label{fig:reward_heatmap_handover}]{%
        \includegraphics[width=\linewidth]{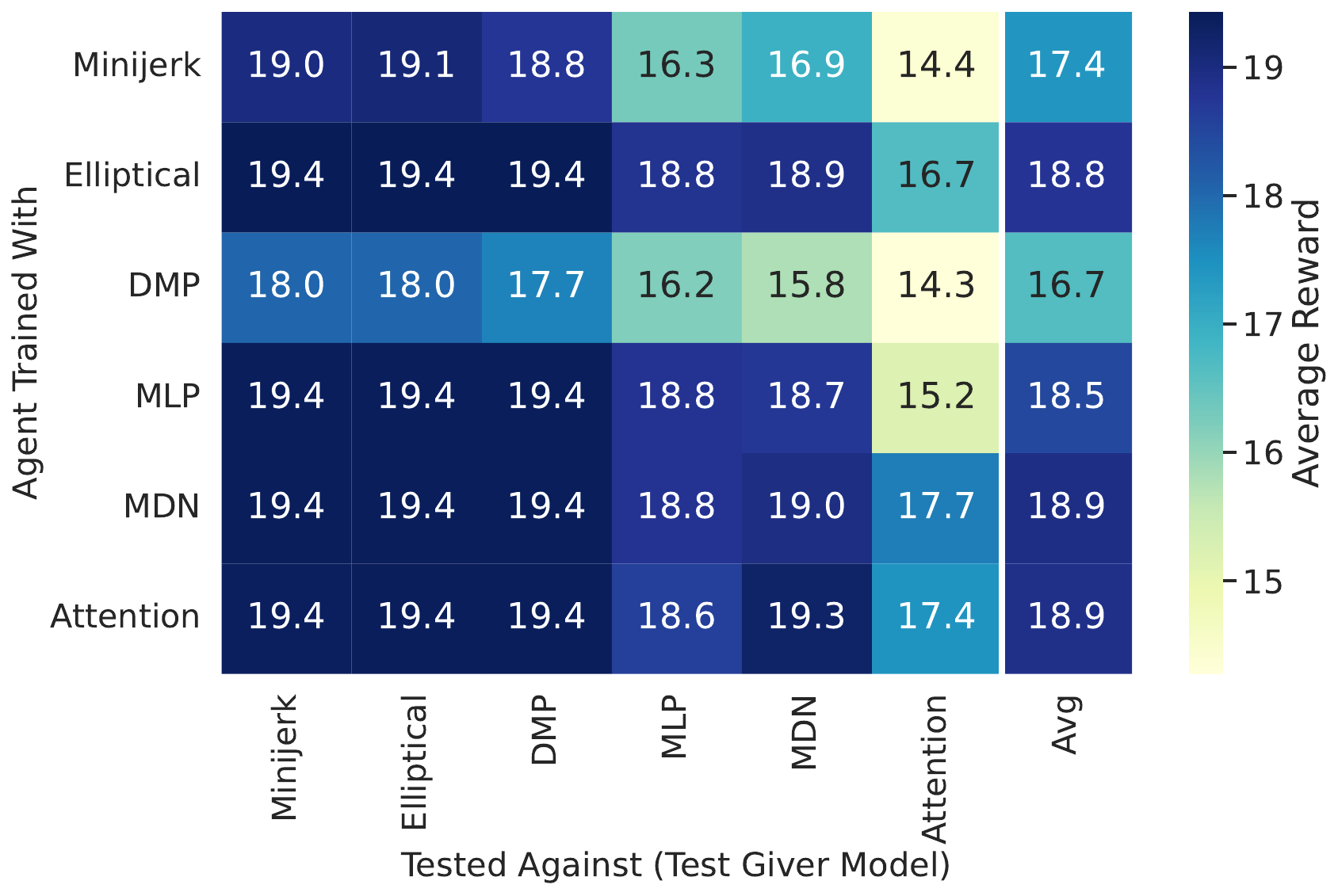}
    }
    \caption{Cross-evaluation results for policies trained and evaluated with different human behaviour models. Each entry represents the mean reward $R_{ij}$ and corresponding reward standard deviation of a policy trained with model $H_i^{\mathrm{train}}$ (row) when evaluated against model $H_j^{\mathrm{eval}}$ (column). (a) Highway domain: $R_{ij}^{\mathrm{HW}}$, averaged over 100 evaluation episodes and three seeds. (b) Handover domain: $R_{ij}^{\mathrm{HO}}$, averaged over 200 evaluation episodes and three seeds. The last column in each subfigure shows the average cross-evaluation reward $R_i$.}
    \label{fig:reward_heatmaps}
\end{figure}

While the preceding analysis examines whether model evaluation metrics are associated with the overall downstream utility of different human behaviour models, the cross-evaluation results provide further insight into how the choice of training model influences policy performance. The heatmaps of $R_{ij}^{d}$ for the highway driving and object handover domains are shown in Figure~\ref{fig:reward_heatmaps}.

A policy trained against human behaviour model $A$ may outperform a policy trained against model $B$ when both policies are evaluated against model $B$. This can be observed from the fact that the diagonal entries in the two cross-evaluation heatmaps are not always the largest values within their corresponding columns. For example, in Figure~\ref{fig:reward_heatmap_highway}, when evaluated against the IDM model, the policy trained against IDM achieves an average reward of $26.9$. However, policies trained against the Gipps, Krau{\ss}, MLP, MDN, and Attention models all achieve higher average rewards under the same evaluation condition. Similar patterns can also be observed for other evaluation models and in the object handover domain. These results indicate that training a policy against a particular human behaviour model does not necessarily maximise its performance when evaluated against the corresponding model. In other words, the human behaviour model that best represents the test distribution is not necessarily the most effective training environment for learning a robust policy.

Qualitative inspection of the generated trajectories provides a further perspective on these results. In the highway domain, the Gipps, Krau{\ss}, and IDM models appear qualitatively more similar to the naturalistic driving behaviours in the highD dataset than the other models (see project page video). Similarly, the DMP and Attention models appear qualitatively more similar to human handover behaviour (see Figure~\ref{fig:handover_trajectories}). However, policies trained against these models do not consistently achieve the highest rewards when evaluated against them. In contrast, some models exhibiting qualitatively undesirable behaviours in the highway domain, including high crash and drive-off rates (see Table~\ref{tab:crash_n_drive_off}), yield policies with better performance.

This suggests that even when the objective is high performance under a particular human-like behaviour distribution, directly training against that distribution may not produce the most effective policy. Instead, training against a different behaviour model may lead to better generalisation to the target interaction conditions.

\section{Discussion}\label{sec:discussion}

Our results provide empirical evidence of an objective mismatch when evaluating human behaviour models for HRI policy learning. Existing evaluation metrics measure different properties of generated behaviour, including similarity to empirical human data, distributional characteristics, diversity, and coverage. However, performing well on these measures does not necessarily mean that a model will be useful for training a downstream RL policy. Across the two domains and three trajectory representations considered in this study, we found no evaluation metric that consistently tracked downstream policy performance. This suggests that behavioural evaluation and downstream utility capture different aspects of a human behaviour model, and that existing generative-model evaluation metrics may not be sufficient for identifying models that are most useful for downstream RL.

The cross-evaluation results further show that the model that provides the best training experience is not necessarily the model against which the resulting policy performs best. A policy trained against one human behaviour model can outperform a policy trained against another when both are evaluated against the latter. This suggests that human behaviour models influence not only the distribution encountered during evaluation, but also the learning process itself. Consequently, selecting a training model based solely on its similarity to the expected evaluation distribution may not lead to the best downstream policy.

Our sweep analysis also highlights that existing evaluation metrics depend on how trajectories are represented. In particular, learned latent representations can produce substantial changes in metric-utility associations when the encoder architecture, training configuration and domain are varied. PCA and manually extracted feature-based representations are comparatively more stable, but are not completely insensitive to their configuration choices or domains. Thus, the choice of trajectory representation is an important part of the evaluation pipeline and can influence both the resulting metric values and the conclusions drawn from them.

Another interesting observation is the asymmetric pattern in the cross-evaluation heatmaps. In the highway domain, the upper-right region of the heatmap, corresponding to policies trained against rule-based models and evaluated against data-driven models, generally contains lower rewards than the other regions. In contrast, policies trained against data-driven models tend to maintain relatively good performance when evaluated against rule-based models. This asymmetry suggests that training against the data-driven models may expose the policy to more challenging behaviours (Table~\ref{tab:crash_n_drive_off}), improving its ability to cope with the more predictable behaviours generated by rule-based models. 

We hypothesise that exposure to such challenging behaviours may have an effect similar to adversarial training, where exposure to difficult examples encourages the learning of more robust policies~\cite{pinto_RobustAdversarialReinforcement_2017, koren_AdaptiveStressTesting_2018, lee_AdaptiveStressTesting_2020}. However, the poor performance of policies trained against \textit{Unif} and \textit{Unif-W}, despite their even higher crash and drive-off rates, indicates that increasing behavioural difficulty alone is insufficient. The challenging behaviours must remain sufficiently structured to provide useful training experience. Excessively unrealistic or unconstrained behaviours may instead hinder policy learning.

This pattern is less apparent in the object handover domain, where successful interaction requires coordination between the human and robot towards a shared goal. The contrast between the two domains suggests that the downstream effect of a human behaviour model may depend on the interaction structure and learning requirements of the task. Importantly, our results do not establish that greater behavioural diversity or higher human-likeness is inherently beneficial. Rather, they suggest that challenging behaviours may sometimes provide useful training experience, while excessively unrealistic behaviours may instead be detrimental. Overall, none of the human model evaluation metrics that are studied in this paper can act as a universal estimator for downstream RL utility, and the usefulness of a model should ultimately be considered in the context of its intended downstream task.

\section{Conclusion}\label{sec:conclusion}

This work investigated whether the objective mismatch observed in model-based RL also arises when human behaviour models are used to train robots in HRI. Across highway driving and object handover, we compared commonly used behavioural evaluation metrics with the performance of RL policies trained using the corresponding human behaviour models.

Our results provide empirical evidence of this mismatch. No evaluation metric consistently identified the models that led to better downstream policy performance across the two domains and different trajectory representations. We also found that evaluation results can depend on how human behaviour is represented, and that training a policy on a particular human behaviour model does not necessarily result in the policy performing best when evaluated against that same model.

These findings show that behavioural evaluation and downstream training utility are distinct aspects of a human behaviour model. A model that appears desirable according to existing evaluation metrics is not necessarily the most useful for robot learning. Future work should therefore investigate evaluation methods that consider how human behaviour models support downstream learning, rather than relying solely on how well they reproduce observed human behaviour.

\bibliographystyle{IEEEtran}
\bibliography{references}

\appendices

\section{Additional Experimental Details}\label{app:additional_experimental_details}

\subsection{Highway Driving}\label{subapp:highway_driving}

The highD dataset contains approximately 110,000 post-processed vehicle trajectories recorded by drones over six German highways near Cologne during 2017 and 2018. In total, the dataset comprises 60 recordings captured at 25 fps, each with an average duration of 17 minutes, covering highway segments of approximately 420 m in length \cite{highDdataset}.

To evaluate downstream policy performance, we train RL agents using Deep Q-Learning in the \textit{highway-env} framework~\cite{highway-env}, a suite of environments designed for decision-making in autonomous driving. In the \textit{Highway} task, the ego vehicle operates on a multi-lane highway populated with surrounding vehicles. At each time step, the agent observes its own position and velocity, together with the relative positions and velocities of nearby vehicles. Based on these observations, it selects one of five discrete actions: \textit{Idle}, \textit{Accelerate}, \textit{Decelerate}, \textit{Lane Left}, and \textit{Lane Right}. The surrounding vehicles are controlled by one of the trained human behaviour models. The objective of the agent is to maintain a high speed while avoiding collisions, with additional rewards for driving in the rightmost lane.

We employ the \textit{highway-fast-v0} environment, a computationally efficient variant of the standard \textit{highway-v0}, characterised by a lower simulation frequency, fewer lanes (three), a reduced number of vehicles (20), and a shorter episode duration (30 seconds). This configuration facilitates faster RL training. All other environment and training hyperparameters are kept at their default settings, except that the behaviour model controlling the surrounding vehicles is replaced with one of our trained human behaviour models.

\subsection{Object Handover}\label{subapp:object_handover}

The handover@RPL dataset comprises 8,672 human-human handover trials collected from 13 participant pairs (26 participants in total). The experimental setup uses a custom 3D-printed baton equipped with three 6D force/torque sensors to measure the giver's grip force, the taker's grip force, and the central interaction wrench. These force measurements are synchronised at $120\,\mathrm{Hz}$ with an OptiTrack motion capture system that tracks a 13-segment upper-body skeletal representation of both participants. To investigate human motor adaptation under different conditions, the dataset contains three experimental settings: normal trials (2,999 handovers), a heavy-object setting in which external weights increase the baton mass from $0.8\,\mathrm{kg}$ to $1.8\,\mathrm{kg}$ (2,764 handovers), and a setting involving giver visual impairment (2,909 handovers). Individual handovers are post-processed and isolated into sequences of $6.675\,\mathrm{s}$, centred around the temporal intersection of the participants' grip forces.

In this study, we focus on the giver's trajectory during the reverse handover task \cite{khanna_MultimodalDataSet_2023}, specifically from the instant at which the giver establishes a stable grip on the baton until its release. This interval corresponds to the interaction phase of interest, during which the human behaviour model acts as the giver and the RL agent acts as the taker. Data from Pair~13 are excluded because the participant is left-handed, whereas the remaining 12 participant pairs consist exclusively of right-handed participants, ensuring consistency in handedness throughout the dataset.

The raw trajectories are preprocessed in two stages. First, each trial is temporally clipped to retain only the interaction segment, with the grip force measured by the baton used to identify the start and end of the handover. Second, the resulting trajectories are filtered and denoised to reduce measurement noise and tracking artefacts. After preprocessing, 1,281 human giver trajectories remain for model training and evaluation.

We develop a customised environment for training the RL agent. The environment contains three entities: the giver's right hand, the taker's right hand, and a table. Both hands are modelled as point masses that move freely in three-dimensional space, without explicitly modelling human arm anatomy or kinematics.

Each hand is constrained to move within a sphere of radius $0.8\,\mathrm{m}$, approximately corresponding to the average human arm length, centred at the respective shoulder position. The shoulder positions are determined by applying fixed offsets to the sampled initial hand positions. The table has dimensions of $1.2 \times 0.8 \times 0.8\,\mathrm{m}$ \cite{khanna_MultimodalDataSet_2023} and is centred at the world origin. A safety margin of $0.1\,\mathrm{m}$ is imposed around the taker to prevent collisions with the table.

At the beginning of each episode, the initial positions of the giver and taker are sampled from the corresponding empirical distributions observed in the human data. An episode terminates when either the taker reaches the giver, defined as a giver-taker distance of less than $0.2\,\mathrm{m}$, or when the maximum episode length of 60 steps is reached. The distance threshold is slightly shorter than the baton length reported in \cite{khanna_MultimodalDataSet_2023}. The reward function is defined as
\begin{equation}
    r_t =
    10(d_{t-1} - d_t)
    + 10\mathbb{I}(d_t < d_{\mathrm{term}}),
\end{equation}
where
$d_t = \lVert \mathbf{p}_{t}^{\mathrm{taker}} - \mathbf{p}_{t}^{\mathrm{giver}} \rVert_2$
is the Euclidean distance between the taker's and giver's hands, $\mathbb{I}(\cdot)$ denotes the indicator function, and $d_{\mathrm{term}} = 0.2\,\mathrm{m}$.

\section{Highway Driving Model Details}\label{app:highway_driving_model_details}

\subsection{Minimising Overall Braking Induced by Lane Changes (MOBIL) Model}\label{subapp:mobil}

MOBIL is a general-purpose lane-changing framework that models both discretionary and mandatory lane changes by evaluating the impact of a potential lane change on the ego vehicle and surrounding vehicles. The highD dataset contains more than 16.5 hours of naturalistic highway driving recordings. However, lane-changing events are relatively rare in highway scenarios. Only around 14,000 lane-changing events are identified in the dataset, and only 5,600 of these are fully completed within the observed area \cite{highDdataset}. This imbalance and limited effective sample size constrain the reliable identification and validation of lane-changing behaviour models. Furthermore, models such as MOBIL require an underlying longitudinal driving model to estimate the consequences of lane changes, such as resulting accelerations of surrounding vehicles. Since this longitudinal component must itself be estimated from data, explicitly calibrating lane-changing behaviour would introduce additional modelling uncertainty. Therefore, to avoid compounding errors and overfitting to sparse lane-changing observations, we adopt standard MOBIL parameter values commonly used in the literature rather than estimating these parameters from the highD dataset.

In MOBIL, a lane change is executed only if two conditions are satisfied. First, the lane change must be safe for the new follower in the target lane:
\begin{equation}\label{eq:MOBIL1}
    \tilde{a}_n \ge -b_{\text{safe}},
\end{equation}
where $\tilde{a}_n$ denotes the post-lane-change acceleration of the new follower in the target lane, and $b_{\text{safe}}$ is the maximum acceptable deceleration.

Second, the lane change must provide sufficient incentive:
\begin{equation}\label{eq:MOBIL2}
    \tilde{a}_c - a_c + p\left( \tilde{a}_n - a_n + \tilde{a}_o - a_o \right) \ge \Delta a_{\text{th}},
\end{equation}
where $a_c$ and $\tilde{a}_c$ denote the accelerations of the centre (ego) vehicle before and after the lane change, respectively. Similarly, $a_o$ and $\tilde{a}_o$ denote the accelerations of the old follower in the current lane before and after the lane change, and $a_n$ and $\tilde{a}_n$ denote the corresponding accelerations of the new follower in the current and target lane. The parameter $p$ is the politeness factor that weights the impact of the lane change on surrounding vehicles, and $\Delta a_{\text{th}}$ is the incentive threshold governing the minimum required advantage for executing a lane change.

As discussed in the paper, we do not explicitly perform parameter estimation for the rule-based lane-changing models. The parameters of the MOBIL model are independently sampled for each vehicle from normal distributions, with means set to the typical values proposed in \cite{Arne1999} for simulations:
\begin{align*}
    p &\sim \lvert X \rvert, \quad X \sim \mathcal{N}(0, 0.25), \\ 
    \Delta a_{\text{th}}&\sim\mathcal{N}(\mu_{\Delta a_{\text{th}}},0.5\mu_{\Delta a_{\text{th}}}), \\ 
    b_{\text{safe}}&\sim\mathcal{N}(\mu_{b_{\text{safe}}},0.5\mu_{b_{\text{safe}}}),
\end{align*}
where
\begin{align*}
    \mu_{\Delta a_{\text{th}}}&=0.1 \text{ m/s}^2, \\ 
    \mu_{b_{\text{safe}}}&=4 \text{ m/s}^2.
\end{align*}

\subsection{Gipps' Model}\label{subapp:gipps}

The Gipps model assumes that drivers maintain smooth and safe car-following behaviour subject to two constraints. First, the vehicle should not exceed its desired speed (i.e., the free-flow speed). Second, if the preceding vehicle performs an emergency stop, the following vehicle must still be able to decelerate safely and avoid a collision.

For the centre vehicle $n$ and the leading vehicle $n-1$:
\begin{itemize}
    \item $x_n(t)$: position of the vehicle
    \item $v_n(t)$: speed of the vehicle
    \item $x_{n-1}(t), v_{n-1}(t)$: position and speed of the leader
    \item $s_n(t) = x_{n-1}(t) - x_n(t) - l_{n-1} - L_{n-1}$: net gap
    \item $l_{n-1}$: physical length of a vehicle
    \item $L_{n-1}$: safe margin
    \item $v_n^{\max}$: desired speed
    \item $a_n$: maximum acceleration
    \item $b_n$: comfortable deceleration
    \item $b_{n-1}$: assumed maximum deceleration of leader
    \item $\tau$: apparent reaction time
\end{itemize}
The speed at the next time step is:
\begin{equation}\label{eq:gippsMin}
v_n(t+\tau) = \min \left\{ v_n^{\text{free}}(t+\tau),\; v_n^{\text{safe}}(t+\tau) \right\},
\end{equation}
where the free-flow speed $v_n^{\text{free}}(t+\tau)$ of the Gipps' model is:
\begin{equation}\label{eq:gippsFF}
v_n^{\text{free}}(t+\tau)=v_n(t)+2.5\, a_n \tau\left(1 - \frac{v_n(t)}{v_n^{\max}}\right)\sqrt{0.025 +\frac{v_n(t)}{v_n^{\max}}},
\end{equation}
and the collision-avoidance speed $v_n^{\text{safe}}(t+\tau)$ is:
\begin{align}\label{eq:gippsCA}
\begin{split}
v_n^{\text{safe}}(t+\tau)&=b_n \tau \\ &\quad + \sqrt{(b_n \tau)^2-b_n\left[2 s_n(t)-v_n(t)\tau-\frac{v_{n-1}(t)^2}{b_{n-1}}\right]}.
\end{split}
\end{align}
Finally, the position is updated as:
\begin{equation}\label{eq:gippsUpdate}
x_n(t+\tau) = x_n(t) + v_n(t+\tau)\,\tau.
\end{equation}

In this work, we infer only the safe margin $L_{n-1}$ and the desired speed $v_n^{\max}$. All other model parameters are held fixed during inference and set to the typical values reported in \cite{Gipps1981}:
\begin{align*}
    a_n&=1.7 \text{ m/s}^2, \\ 
    b_n&=-3.4 \text{ m/s}^2, \\
    b_{n-1}&=-3 \text{ m/s}^2, \\
    \tau&=0.6 \text{ s}.
\end{align*}
Since the $\min\{\cdot\}$ operator is not differentiable, we decompose the likelihood into two separate components. Specifically, we infer the parameters associated with $v_n^{\text{free}}(t+\tau)$ and $v_n^{\text{safe}}(t+\tau)$ independently. To this end, the training data are partitioned based on the distance headway (DHW). Observations with a DHW shorter than 25~m are used to infer the safe margin $L_{n-1}$ through the collision-avoidance component $v_n^{\text{safe}}(t+\tau)$, while observations with a DHW greater than or equal to 25~m are used to infer the desired speed $v_n^{\max}$ through the free-flow component $v_n^{\text{free}}(t+\tau)$. The threshold of 25~m is selected empirically. Both $L_{n-1}$ and $v_n^{\max}$ are inferred using Automatic Differentiation Variational Inference (ADVI) \cite{kucukelbir_automatic_2017}. For notational simplicity, we drop the subscripts $n-1$ and $n$ from $L_{n-1}$ and $v_n^{\max}$ in the following paragraphs, and simply refer to them as $L$ and $v_{max}$, respectively.

To capture inter-driver variability while sharing statistical strength across vehicles, we assume a hierarchical prior structure. Specifically, the safe margin for vehicle $i$, denoted $L_i$, is drawn from a population-level distribution:
\begin{align*}
    \mu_L\sim\mathcal{N}(2.0, 1.0^2),&\quad \sigma_L\sim\text{HalfNormal}(0.5), \\
    L_i^*\sim\mathcal{N}(\mu_L,\sigma_L^2),&\quad L_i^*\in(0, \min(\text{DHW})),
\end{align*}
where the truncation enforces physical feasibility. To ensure strict positivity and numerical stability during inference, we apply a softplus transformation:
\begin{equation}
L_i=\log(1+\exp(L_i^*)).
\end{equation}
For each observation $t$ associated with vehicle $i$, the predicted safe speed at time $t+\tau$ is given by Equation \ref{eq:gippsCA}. Observed speeds $v_n(t+\tau)$ are assumed to be affected by Gaussian noise:
\begin{equation}\label{eq:gippsGaussianNoise}
v(t+\tau)\sim\mathcal{N}(\mu_v(t+\tau), \sigma_{\text{obs}}^2),\quad \sigma_{\text{obs}}\sim\text{HalfNormal}(1.0).
\end{equation}
Posterior inference is performed using ADVI, yielding approximate posterior distributions over both population-level hyperparameters $(\mu_L,\sigma_L)$ and vehicle-specific safe margins $L_i$. We employ a mean-field variational family and optimise the evidence lower bound (ELBO) for 30,000 iterations. After convergence, 2,000 posterior samples are drawn from the fitted variational approximation and used for all subsequent experiments.

Similarly, a hierarchical Bayesian formulation is employed to capture inter-driver variability for the desired speed $v_i^{\max}$:
\begin{align*}
    \mu_v^{\max}\sim\mathcal{N}(20.0, 1.0^2),&\quad \sigma_v^{\max}\sim\text{HalfNormal}(0.5), \\
    v_i^{\max}=\log(1+\exp(v_i^{*\max})),&\quad v_i^{*\max}\sim\mathcal{N}(\mu_v^{\max} ,\sigma_v^{\max 2}).
\end{align*}
The predicted next speed under free-flow conditions is given by Equation \ref{eq:gippsFF} with Gaussian noise applied identical to Equation \ref{eq:gippsGaussianNoise}. Posterior inference is performed using mean-field ADVI for 100,000 iterations, after which 2,000 posterior samples are drawn from the variational approximation for downstream experiments.

We report the ADVI results for population-level hyperparameters $(\mu_L,\sigma_L,\sigma_{\text{obs}})$ and $(\mu_v^{\max},\sigma_v^{\max},\sigma_{\text{obs}})$ in Table \ref{tab:GippsCA} and \ref{tab:GippsFF}, respectively.

\begin{table*}[th]
    \centering
    \caption{Gipps' ADVI report for $L$}
    \label{tab:GippsCA}
    \begin{tabular}{lrrrrrrrr}
        \toprule
        {} & mean & sd & hdi\_3\% & hdi\_97\% & mcse\_mean & mcse\_sd & ess\_bulk & ess\_tail \\
        \midrule
        $\mu_L$      & 2.486 & 0.068 & 2.358 & 2.610 & 0.002 & 0.001 & 1906.0 & 1931.0 \\
        $\sigma_L$   & 1.332 & 0.069 & 1.211 & 1.466 & 0.002 & 0.001 & 2030.0 & 2001.0 \\
        $\sigma_{\text{obs}}$ & 1.466 & 0.021 & 1.430 & 1.508 & 0.001 & 0.000 & 1793.0 & 1865.0 \\
        \bottomrule
    \end{tabular}
\end{table*}

\begin{table*}[th]
    \centering
    \caption{Gipps' ADVI report for $v^{\max}$}
    \label{tab:GippsFF}
    \begin{tabular}{lrrrrrrrr}
        \toprule
        {} & mean  & sd    & hdi\_3\% & hdi\_97\% & mcse\_mean & mcse\_sd & ess\_bulk & ess\_tail \\
        \midrule
        $\mu_v^{\max}$   & 32.336 & 0.216 & 31.926 & 32.734 & 0.005 & 0.003 & 2033.0 & 1882.0 \\
        $\sigma_v^{\max}$ & 5.541 & 0.136 & 5.292 & 5.805 & 0.003 & 0.002 & 1914.0 & 1810.0 \\
        $\sigma_{\text{obs}}$ & 0.243 & 0.002 & 0.240 & 0.246 & 0.000 & 0.000 & 1975.0 & 1886.0 \\
        \bottomrule
    \end{tabular}
\end{table*}

\subsection{Krau{\ss} Model}\label{subapp:krauss}

The Krau{\ss} model determines vehicle motion by selecting a safe speed from dynamically feasible bounds, rather than explicitly solving a worst-case stopping constraint to obtain a single admissible next speed. Due to its computational efficiency and robustness, the model has been widely adopted in traffic simulators such as SUMO~\cite{SUMO2018}.

Using the same notation as the Gipps' model in Section~\ref{subapp:gipps}, the safe speed given by the Krau{\ss} model is:
\begin{equation}\label{eq:KraussSafe}
    v_{\text{safe}}(t)=v_{n-1}(t)+\frac{d_n(t)-v_{n-1}(t)\tau}{\frac{v_n(t)+v_{n-1}(t)}{2b_n}+\tau},
\end{equation}
where $d_n(t) = x_{n-1}(t) - x_n(t) - l_{n-1}$, which does not involve the safe margin. At each discrete time step, the vehicle chooses its speed according to:
\begin{equation}\label{eq:KruassUpdate}
    v_n(t+\Delta t)=\min\left(v_{\text{safe}}(t), v_n(t)+a_n \Delta t, v_n^{\max}\right)
\end{equation}
Note that in the Krau{\ss} model, the numerical update time step $\Delta t$ is distinct from the driver reaction time $\tau$. In this study, we infer only the reaction time $\tau$, while all other model parameters are fixed to the same values as those used for the Gipps' model. An exception is the maximum deceleration parameter $b_n$, which is specified as a positive quantity (e.g., $b_n = 3.4\,\mathrm{m/s^2}$).

As with the Gipps' model, the Krau{\ss} model involves non-differentiable components. To address this issue, we adopt the same strategy employed for the Gipps' model. The likelihood of the safe speed is evaluated independently to infer the reaction time $\tau$, using only training observations with a DHW shorter than 25~m.

To capture inter-driver variability in reaction time, we employ a hierarchical Bayesian formulation for $\tau$:
\begin{align*}
    \mu_\tau\sim\mathcal{N}(0.8, 0.3^2),&\quad \sigma_\tau\sim\text{HalfNormal}(0.5), \\
    \tau_i=\log(1+\exp(\tau_i^{*})),&\quad \tau_i^{*}\sim\mathcal{N}(\mu_\tau ,\sigma_\tau^{2}).
\end{align*}
The predicted next speed is given by Equation \ref{eq:KraussSafe} with Gaussian noise applied identical to Equation \ref{eq:gippsGaussianNoise}. Posterior inference is performed using mean-field ADVI for 20,000 iterations, after which 2,000 posterior samples are drawn from the variational approximation for downstream experiments. We report the ADVI results for population-level hyperparameters $(\mu_\tau,\sigma_\tau,\sigma_{\text{obs}})$ in Table \ref{tab:Krauss}.
\begin{table*}[th]
    \centering
    \caption{Krau{\ss} ADVI report for $\tau$}
    \label{tab:Krauss}
    \begin{tabular}{lrrrrrrrr}
        \toprule
        {} & mean & sd & hdi\_3\% & hdi\_97\% & mcse\_mean & mcse\_sd & ess\_bulk & ess\_tail \\
        \midrule
        $\mu_\tau$   & 0.087 & 0.051 & -0.006 & 0.184 & 0.001 & 0.001 & 1695.0 & 1766.0 \\
        $\sigma_\tau$ & 1.341 & 0.045 & 1.256 & 1.423 & 0.001 & 0.001 & 1930.0 & 1743.0 \\
        $\sigma_{\text{obs}}$ & 0.995 & 0.020 & 0.959 & 1.034 & 0.000 & 0.000 & 1896.0 & 1885.0 \\
        \bottomrule
    \end{tabular}
\end{table*}

\subsection{Intelligent Driver Model (IDM)}\label{subapp:idm}

While both the Gipps and Krau{\ss} models are first-order formulations, the IDM is a second-order model that explicitly incorporates both vehicle velocity and acceleration when computing longitudinal dynamics.

The acceleration of an IDM vehicle is computed at each step according to:
\begin{equation}\label{eq:IDMAcc}
    \dot{v}_n(t) = a_n\left[1-\left(\frac{v_n(t)}{v^{\max}}\right)^{\delta}-\left(\frac{d_n^*(t)}{d_n}\right)^2\right],
\end{equation}
where
\begin{equation}\label{eq:IDMDist}
    d_n^*(t)=d_0 + T\cdot v_n(t) + \frac{v_n(t)(v_n(t)-v_{n-1}(t))}{2\sqrt{a_n\cdot b_n}}.
\end{equation}

The IDM introduces three hyperparameters that do not appear in the Gipps or Krau{\ss} models, namely the acceleration exponent $\delta$, the safe time headway $T$, and the jam distance $d_0$. The acceleration exponent $\delta$ does not have a direct physical interpretation and is typically fixed to $\delta = 4$. In this study, we infer three IDM parameters: the desired velocity $v^{\max}$, the safe time headway $T$, and the jam distance $d_0$. All remaining parameters are fixed to the standard values proposed in \cite{Treiber2000}:
\begin{align*}
    \delta&=4; \\
    a_n&=0.73 \text{ m/s}^2; \\ 
    b_n&=1.67 \text{ m/s}^2.
\end{align*}
Unlike the Gipps' and Krau{\ss} models, the IDM is fully differentiable. Consequently, all available observations are used for training the IDM. A similar hierarchical Bayesian formulation is adopted:
\begin{align*}
    \mu_v^{\max}\sim\mathcal{N}(30.0, 1.0^2),&\quad \sigma_v^{\max}\sim\text{HalfNormal}(0.5), \\
    v_i^{\max}=\log(1+\exp(v_i^{*\max})),&\quad v_i^{*\max}\sim\mathcal{N}(\mu_v^{\max} ,\sigma_v^{\max 2}), \\
    \mu_{d_0}\sim\mathcal{N}(2.0, 1.0^2),&\quad \sigma_{d_0}\sim\text{HalfNormal}(1.0), \\
    {d_0}_i=\log(1+\exp({d_0}_i^{*})),&\quad {d_0}_i^{*}\sim\mathcal{N}(\mu_{d_0} ,\sigma_{d_0}^{2}), \\
    \mu_T\sim\mathcal{N}(1.5, 0.5^2),&\quad \sigma_T\sim\text{HalfNormal}(0.5), \\
    T_i=\log(1+\exp(T_i^{*})),&\quad T_i^{*}\sim\mathcal{N}(\mu_T ,\sigma_T^{2}).
\end{align*}
The likelihood is given by Equation \ref{eq:IDMAcc} and \ref{eq:IDMDist}. Posterior inference is performed using mean-field ADVI for 15,000 iterations, after which 2,000 posterior samples are drawn from the variational approximation for downstream experiments.

We report the ADVI results for population-level hyperparameters $(\mu_v^{\max},\sigma_v^{\max}, \mu_{d_0}, \sigma_{d_0}, \mu_{T}, \sigma_{T}, \sigma_{\text{obs}})$ in Table \ref{tab:IDM}.

\begin{table*}[th]
    \centering
    \caption{IDM ADVI Report for $v_n^{\max}$, $d_0$ and $T$}
    \label{tab:IDM}
    \begin{tabular}{lrrrrrrrr}
        \toprule
        {} & mean & sd & hdi\_3\% & hdi\_97\% & mcse\_mean & mcse\_sd & ess\_bulk & ess\_tail \\
        \midrule
        $\mu_v^{\max}$    & 32.671 & 0.096 & 32.491 & 32.848 & 0.002 & 0.002 & 1993.0 & 1961.0 \\
        $\sigma_{v}^{\max}$ &  1.249 & 0.085 &  1.102 &  1.408 & 0.002 & 0.001 & 2148.0 & 2004.0 \\
        $\mu_{d_0}$    &  0.371 & 0.088 &  0.210 &  0.540 & 0.002 & 0.001 & 2005.0 & 1985.0 \\
        $\sigma_{d_0}$ &  1.338 & 0.085 &  1.183 &  1.504 & 0.002 & 0.001 & 1981.0 & 1984.0 \\
        $\mu_{T}$     & -0.812 & 0.086 & -0.979 & -0.664 & 0.002 & 0.001 & 1996.0 & 1960.0 \\
        $\sigma_{T}$  &  1.258 & 0.084 &  1.110 &  1.427 & 0.002 & 0.001 & 2065.0 & 1698.0 \\
        $\sigma_{\text{obs}}$&  2.094 & 0.137 &  1.863 &  2.371 & 0.003 & 0.002 & 2102.0 & 1876.0 \\
        \bottomrule
    \end{tabular}
\end{table*}

\subsection{Unif and Unif-W}\label{subapp:unif}

\textit{Unif} samples longitudinal and lateral accelerations from $U(-4,\,3)$ and $U(-2,\,2)$, respectively, whereas \textit{Unif-W} uses the same longitudinal range but a wider lateral acceleration range of $U(-6,\,6)$.

\subsection{MLP}\label{subapp:mlp-highway}

Behaviour cloning is an imitation learning approach in which an agent learns to reproduce expert behaviour by formulating the problem as supervised learning over state-action pairs. Given demonstrations from an expert policy, a model is trained to predict the expert action conditioned on the current state, without requiring an explicit reward function. In this study, expert demonstrations are obtained from the highD highway driving dataset.

For all three data-driven models, trajectories extracted from the highD dataset are first processed into structured feature representations comprising both ego-vehicle states and surrounding vehicle states before being provided as model inputs. Specifically, the input state is defined in an ego-centric coordinate frame and consists of the following components:
\begin{itemize}
    \item Longitudinal velocity of the ego vehicle, $v_e^{\text{lon}} \in \mathbb{R}$;
    \item Lateral velocity of the ego vehicle, $v_e^{\text{lat}} \in \mathbb{R}$;
    \item Lateral offset from the lane centre, $\delta y_e = y_e - y_{\text{lane centre}}$;
    \item Physical length of the ego vehicle, $l_e$;
    \item Binary indicators denoting the existence of an adjacent lane to the left and right, $(\mathds{1}_{\text{left}}, \mathds{1}_{\text{right}})$;
    \item Relative states of seven surrounding vehicles: $s_{p}$, $s_{la}$, $s_{lp}$, $s_{lf}$, $s_{ra}$, $s_{rp}$, and $s_{rf}$, corresponding to the preceding vehicle, left-along, left-preceding, left-following, right-along, right-preceding, and right-following vehicles, respectively.
\end{itemize}

A surrounding vehicle is classified as left-along or right-along if its longitudinal projection overlaps with that of the ego vehicle. The relative state of a surrounding vehicle $i$ is defined as
\begin{align}\label{eq:relativeStates}
\begin{split}
    s_i &=
    \bigl[
        \mathds{1}_{i},\,
        l_i,\,
        \Delta x_i,\,
        \Delta y_i,\,
        \Delta v_i^{\text{lon}},\,
        \Delta v_i^{\text{lat}}
    \bigr], \\
    i &\in \{p, la, lp, lf, ra, rp, rf\},
\end{split}
\end{align}
where $\mathds{1}_{i}$ indicates the existence of vehicle $i$, $l_i$ denotes its physical length, $\Delta x_i = x_i - x_e$ and $\Delta y_i = y_i - y_e$ are the relative longitudinal and lateral distances, and $\Delta v_i^{\text{lon}}$ and $\Delta v_i^{\text{lat}}$ denote the relative longitudinal and lateral velocities with respect to the ego vehicle. If vehicle $i$ does not exist, all values in $s_i$ are set to zero.

The complete state vector is given by
\begin{align}\label{eq:dataDrivenStates}
\begin{split}
    \mathbf{s} =
    &\bigl[
        v_e^{\text{lon}},\,
        v_e^{\text{lat}},\,
        l_e,\,
        \delta y_e,\,
        \mathds{1}_{\text{left}},\,
        \mathds{1}_{\text{right}},\, \\
        & \quad s_{p},\,
        s_{la},\,
        s_{lp},\,
        s_{lf},\,
        s_{ra},\,
        s_{rp},\,
        s_{rf}
    \bigr],
\end{split}
\end{align}
resulting in a $48$-dimensional input representation.

The policy network outputs continuous control actions,
\begin{equation}\label{eq:dataDrivenActions}
    \mathbf{a} =
    \bigl[
        \hat{a}^{\text{lon}},\,
        \hat{a}^{\text{lat}}
    \bigr],
\end{equation}
where $\hat{a}^{\text{lon}} \in \mathbb{R}$ denotes the predicted longitudinal acceleration and $\hat{a}^{\text{lat}} \in \mathbb{R}$ denotes the predicted lateral acceleration.

The MLP model is trained using a mean squared error loss (MSE) over the continuous action space:
\begin{equation}
    \mathcal{L}_{\text{MLP}}
    =
    \frac{1}{B}
    \sum_{i=1}^{B}
    \left\|
        \hat{\mathbf{a}}_i - \mathbf{a}_i^{\text{gt}}
    \right\|_2^2,
\end{equation}
where $B$ denotes the batch size and $\mathbf{a}_i^{\text{gt}}$ is the ground-truth action extracted from the dataset.

The raw highD dataset is recorded at 25~fps, resulting in only minor differences between consecutive samples. To reduce temporal redundancy during training, we downsample the dataset to 5~fps. Prior to training, the state-action pairs are standardised.

Specifically, all continuous state and action variables, including vehicle length, are standardised, while binary indicator features are retained as $0$ or $1$. For relative continuous variables such as $\Delta x_i$, a value of zero may correspond either to the non-existence of a vehicle or to zero relative distance. Standardising these variables across all samples would therefore introduce semantic ambiguity. To avoid this issue, we standardise relative state components only for samples where the corresponding presence indicator satisfies $\mathds{1}_i = 1$. Otherwise, all relative-state values are set to zero. We split the dataset into 80\% for training, 10\% for validation, and 10\% for testing, resulting in approximately 1.7 million training samples and around 0.2 million samples each for validation and testing.

We employ a simple MLP with two hidden layers of 64 units each, using ReLU activations throughout. The batch size is set to 1024, and the learning rate is fixed at $10^{-3}$. Training is performed for a maximum of 100 epochs with early stopping applied using a patience of 10 epochs. Early stopping is triggered at the 88th epoch, yielding the best test loss of 0.62.

\subsection{MDN}\label{subapp:mdn-highway}
A mixture density network models a target variable as a weighted mixture of probability distributions (typically Gaussians). Instead of predicting a single point estimate, it outputs the mixture weights and parameters, enabling it to represent multimodal and uncertain relationships between inputs and outputs. We employ a MDN in this study because human driving behaviour is inherently multimodal. For example, drivers may adopt different strategies for car-following and lane-changing.

We reuse the same state representation and action space as those used for BC with an MLP. The MDN models the conditional action distribution as a mixture of Gaussians:
\begin{equation}
    p(\mathbf{a} \mid \mathbf{s}) = \sum_{k=1}^{K} \alpha_k(\mathbf{s}) 
    \mathcal{N}\!\left(\mathbf{a} \mid \boldsymbol{\mu}_k(\mathbf{s}), \boldsymbol{\Sigma}_k(\mathbf{s})\right),
\end{equation}
where $\alpha_k(\mathbf{s})$ denotes the mixing coefficient of the $k$-th component, $\boldsymbol{\mu}_k \in \mathbb{R}^2$ is the mean vector, and $\boldsymbol{\Sigma}_k \in \mathbb{R}^{2 \times 2}$ is the covariance matrix.

In this study, we assume independence between longitudinal and lateral accelerations, and thus restrict the covariance matrices to be diagonal:
\begin{equation}
    \boldsymbol{\Sigma}_k =
    \begin{pmatrix}
        \sigma_{k,\text{lon}}^2 & 0 \\
        0 & \sigma_{k,\text{lat}}^2
    \end{pmatrix}.
\end{equation}

The MDN is trained by minimising a composite loss that extends the standard negative log-likelihood (NLL) with additional regularisation terms to encourage mode utilisation and diversity. For a batch of size $B$, the loss is:

\begin{align}
\begin{split}
&\mathcal{L}_{\text{MDN}} = 
\underbrace{- \frac{1}{B} \sum_{i=1}^{B} \log \left( \sum_{k=1}^{K} 
    \alpha_k^{(i)} \,
    \mathcal{N}\!\left(\mathbf{a}^{\text{gt},(i)} \mid \boldsymbol{\mu}_k^{(i)}, \boldsymbol{\Sigma}_k^{(i)} \right) \right)}_{\text{NLL}} \\
&\quad+ 
\lambda_{\text{usage}} \underbrace{D_{\mathrm{KL}}\!\left( \bar{\mathbf{r}} \,\|\, U \right)}_{\text{Mode usage penalty}} \\
&\quad+
\lambda_{\text{div}} \underbrace{\frac{1}{K(K-1)B} \sum_{i=1}^{B} \sum_{k \neq j} \exp\!\left(- D_{\mathrm{KL}}\big( \mathcal{N}_k^{(i)} \,\|\, \mathcal{N}_j^{(i)} \big) \right)}_{\text{Component diversity penalty}},
\end{split}
\end{align}

where $\alpha_k^{(i)}$ is the mixture weight for the $k$-th component of sample $i$, and $\boldsymbol{\mu}_k^{(i)}$ and $\boldsymbol{\Sigma}_k^{(i)}$ are the corresponding mean and covariance. The vector $\bar{\mathbf{r}} = \frac{1}{B} \sum_{i=1}^{B} \mathbf{r}^{(i)}$ is the average responsibility vector over the batch, where each $\mathbf{r}^{(i)} \in \mathbb{R}^K$ contains the posterior probabilities (responsibilities) of the $K$ mixture components for sample $i$:
\begin{equation}
r_k^{(i)} = \frac{\alpha_k^{(i)} \, \mathcal{N}(\mathbf{a}^{\text{gt},(i)} \mid \boldsymbol{\mu}_k^{(i)}, \boldsymbol{\Sigma}_k^{(i)})}{\sum_{j=1}^{K} \alpha_j^{(i)} \, \mathcal{N}(\mathbf{a}^{\text{gt},(i)} \mid \boldsymbol{\mu}_j^{(i)}, \boldsymbol{\Sigma}_j^{(i)})}.
\end{equation}
And $U$ represents a uniform distribution over the mixture components. $\mathcal{N}_k^{(i)}$ or $\mathcal{N}_j^{(i)}$ denotes the Gaussian for component $k$ or $j$ of sample $i$, respectively.

The first term encourages the model to fit the ground-truth actions, the second term discourages mode collapse by promoting uniform usage of mixture components across the batch, and the third term encourages diversity between components by penalising overlapping or redundant Gaussians. The hyperparameters $\lambda_{\text{usage}}$ and $\lambda_{\text{div}}$ control the strength of these regularisation terms.

The same training, validation, and test splits used for the MLP-based behaviour cloning model are also used for training and evaluating the MDN. We employ a simple MDN with two hidden layers of 128 units each and ReLU activations throughout, producing $K=15$ mixture components. The batch size is set to 512, and the learning rate is fixed at $10^{-3}$. Training is conducted for a maximum of 100 epochs with early stopping applied using a patience of 10 epochs. The regularisation weights $\lambda_{\text{usage}}$ and $\lambda_{\text{div}}$ are initially set to $0.05$ and $0.01$, respectively. The diversity weight $\lambda_{\text{div}}$ is annealed exponentially during training according to
\[
\lambda_{\text{div},e} = \lambda_{\text{div},0} \, \exp(-\gamma \cdot e),
\]
where $\lambda_{\text{div},0}$ is the initial value, $\gamma = 0.05$ is the decay factor, and $e$ denotes the current epoch number. Early stopping is triggered at the 48th epoch, resulting in the test loss of 1.88. 

\subsection{Attention}\label{subapp:attention-highway}
The MLP and MDN models do not utilise the history of states, which can be informative for trajectory forecasting. To address this, we adopt an attention-based design that incorporates past states into the prediction. The multi-head attention (MHA) function, as proposed in the Transformer \cite{vaswani_AttentionAllYou_2017}, has been widely used to model interactions between elements in a sequence. Given queries $Q \in \mathbb{R}^{N_q \times d_k}$, keys $K \in \mathbb{R}^{N_k \times d_k}$, and values $V \in \mathbb{R}^{N_k \times d_v}$, a single attention head computes
\begin{equation}
\text{Attn}(Q,K,V) = \text{softmax}\Big(\frac{Q K^\top}{\sqrt{d_k}}\Big) V,
\end{equation}
where $d_k$ and $d_v$ are the dimension of the keys and values, and $N_q$ and $N_k$ are the number of tokens of queries and keys, respectively. Multi-head attention splits the input into $h$ heads, applies the attention function in parallel, and concatenates the results:
\begin{equation}
\text{MHA}(Q,K,V) = \text{Concat}(\text{head}_1, \dots, \text{head}_h) W_O,
\end{equation}
where each head is $\text{head}_i = \text{Attn}(Q W_Q^i, K W_K^i, V W_V^i)$ and $W_Q^i, W_K^i, W_V^i, W_O$ are learnable projections.

\begin{figure}
    \centering
    \includegraphics[width=1\linewidth, trim=40pt 0pt 70pt 0pt, clip]{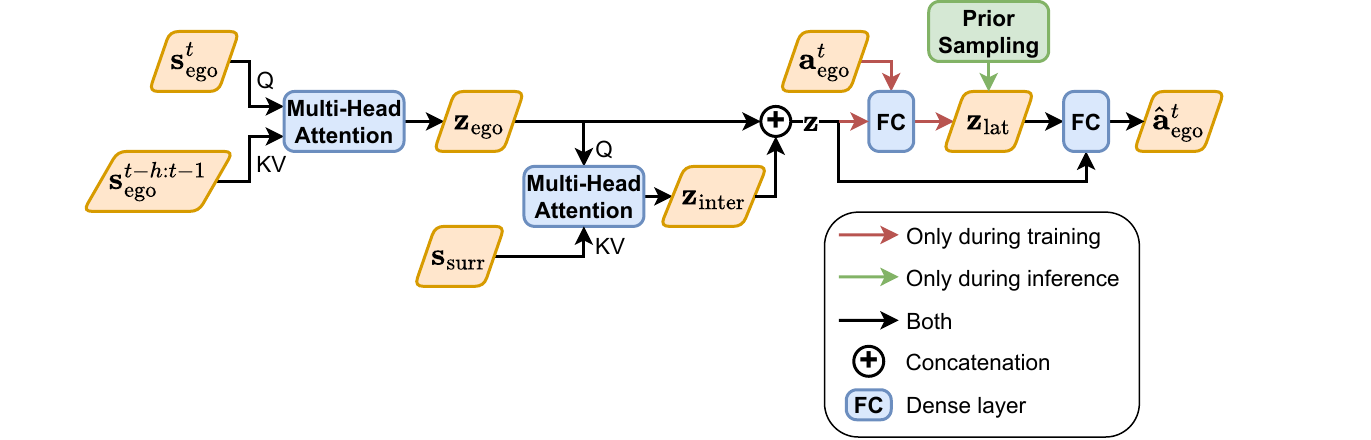}
    \caption{An illustration of the attention-based model architecture. At each time step $t$, the current ego state $\mathbf{s}_{\text{ego}}^{t} \in \mathbb{R}^{6}$ (where bold symbols denote vectors) queries its past $h$ ego states 
    $\mathbf{s}_{\text{ego}}^{t-h:t-1} \in \mathbb{R}^{h \times 6}$ to produce the ego embedding 
    $\mathbf{z}_{\text{ego}} \in \mathbb{R}^{H \cdot d_z}$. This embedding then queries the states of the $N$ surrounding vehicles ($N=7$ in our case)
    $\mathbf{s}_{\text{surr}} \in \mathbb{R}^{N \times (h+1) \times 6}$ to generate the interaction embedding 
    $\mathbf{z}_{\text{inter}} \in \mathbb{R}^{H \cdot d_z}$. The ego and interaction embeddings are concatenated into a single scene representation
    $\mathbf{z} = \text{Concat}(\mathbf{z}_{\text{ego}}, \mathbf{z}_{\text{inter}}) \in \mathbb{R}^{2H \cdot d_z}$,
    which is used as the conditioning input to a Conditional Variational Autoencoder (CVAE) \cite{sohn_LearningStructuredOutput_2015}. During training, the encoder takes both $\mathbf{z}$ and the ground-truth control action $\mathbf{a}_{\text{ego}}^{t} \in \mathbb{R}^{2}$ to infer a latent variable $\mathbf{z}_{\text{lat}}$. The decoder then generates the control action $\hat{\mathbf{a}}_{\text{ego}}^{t} \in \mathbb{R}^{2}$ conditioned on the concatenation of $\mathbf{z}$ and a sampled latent vector. During inference, $\mathbf{z}_{\text{lat}}$ is sampled from a prior distribution.}
    \label{fig:attentionBasedModel}
\end{figure}

In this study, we use the MHA function to learn correlations between the ego vehicle's history states and its current state. Following the same notation as used in MLP and MDN, we denote the state of the ego vehicle at time $t$ as $\mathbf{s}_{\text{ego}}^{t}\in\mathbb{R}^{6}$:
\begin{equation}
    \mathbf{s}_{\text{ego}}^{t} = \bigl[
        v_e^{\text{lon}}(t),\,
        v_e^{\text{lat}}(t),\,
        l_e,\,
        \delta y_e(t),\,
        \mathds{1}_{\text{left}}(t),\,
        \mathds{1}_{\text{right}}(t)
        \bigr],
\end{equation}
and its history over $h$ steps as
\begin{equation}
    \mathbf{s}_{\text{ego}}^{t-h:t-1} = \bigl[
        \mathbf{s}_{\text{ego}}^{t-h},\,
        \mathbf{s}_{\text{ego}}^{t-h+1},\,
        \dots,\,
        \mathbf{s}_{\text{ego}}^{t-1}
        \bigr]\in \mathbb{R}^{h\times6}.
\end{equation}
We then apply a cross-attention function that allows the current ego state $\mathbf{s}_{\text{ego}}^{t}$ to query its history states $\mathbf{s}_{\text{ego}}^{t-h:t-1}$, producing the ego embedding $\mathbf{z}_{\text{ego}}\in \mathbb{R}^{H\cdot d_z}$, where $H$ is the number of attention heads and $d_z$ is the embedding dimension (also the dimension of the values):
\begin{equation}
    \mathbf{z}_{\text{ego}} = \text{Concat}_H\Bigl(\text{MHA}\bigl(\mathbf{s}_{\text{ego}}^{t},\mathbf{s}_{\text{ego}}^{t-h:t-1},\mathbf{s}_{\text{ego}}^{t-h:t-1}\bigr)\Bigr).
\end{equation}
This is followed by another cross-attention, in which the embedded ego state $\mathbf{z}_{\text{ego}}$ queries the states of the seven surrounding vehicles:
\begin{equation}
    \mathbf{s}_{\text{surr}} = \bigl[\mathbf{s}_{i}^{t-h:t}\bigr], \quad i \in \{p, la, lp, lf, ra, rp, rf\},
\end{equation}
where each $\mathbf{s}_{i}^{t}$ is the corresponding relative state of surrounding vehicle $i$ at time $t$ as defined in paper Equation 7. The surrounding states $\mathbf{s}_{\text{surr}} \in \mathbb{R}^{N \times (h+1) \times 6}$ are reshaped into a two-dimensional tensor of size $\mathbb{R}^{N(h+1) \times 6}$ prior to the attention operation. This transformation yields a sequence of $N(h+1)$ tokens, each corresponding to the state of a surrounding vehicle at a specific time step. The cross-attention MHA outputs the embedding of interactions between the ego and surrounding vehicles:
\begin{equation}
    \mathbf{z}_{\text{inter}} = \text{Concat}_H\Bigl(\text{MHA}\bigl(\mathbf{z}_{\text{ego}}, \mathbf{s}_{\text{surr}},\mathbf{s}_{\text{surr}}\bigr)\Bigr)\in \mathbb{R}^{H\cdot d_z}.
\end{equation}
Finally, we concatenate the ego and interaction embeddings into a single vector,
\begin{equation}
    \mathbf{z} = \text{Concat}\bigl(\mathbf{z}_{\text{ego}}, \mathbf{z}_{\text{inter}}\bigr) \in \mathbb{R}^{2 H \cdot d_z}.
\end{equation}
We use a conditional variational autoencoder (CVAE) \cite{sohn_LearningStructuredOutput_2015} to model the distribution over control actions in order to capture the inherent multi-modality and uncertainty in the ego vehicle's future actions. During training, the encoder takes both $\mathbf{z}$ and the ground-truth action $\mathbf{a}_{\text{ego}}^{t} \in \mathbb{R}^{2}$ to compute the posterior distribution over the latent variable:
\begin{equation}
    q_\phi(\mathbf{z}_\text{lat} \mid \mathbf{a}_{\text{ego}}^{t}, \mathbf{z}) = \mathcal{N}(\boldsymbol{\mu}_\phi, \text{diag}(\boldsymbol{\sigma}_\phi^2)), \quad \mathbf{z}_\text{lat} \in \mathbb{R}^{d_\text{lat}}.
\end{equation}
A latent sample $\mathbf{z}_\text{lat}$ is drawn using the reparameterisation trick and concatenated with the embedding $\mathbf{z}$ to produce the predicted action via the decoder:
\begin{equation}
    \hat{\mathbf{a}}_{\text{ego}}^{t} = f_\theta\bigl([\mathbf{z}, \mathbf{z}_\text{lat}]\bigr) \in \mathbb{R}^{2}.
\end{equation}
During inference, since $\mathbf{a}_{\text{ego}}^{t}$ is unavailable, the latent vector is sampled from the prior
\begin{equation}
    \mathbf{z}_\text{lat} \sim \mathcal{N}(0,I),
\end{equation}
and the decoder generates one or multiple plausible control actions conditioned on $\mathbf{z}$. The architecture of the customised attention-based model is illustrated in Figure~\ref{fig:attentionBasedModel}.

The CVAE is trained by minimising the negative variational lower bound (ELBO), which consists of a reconstruction loss and a Kullback--Leibler (KL) divergence regularisation term. Given the conditioning embedding $\mathbf{z}$ and the ground-truth ego action $\mathbf{a}_{\text{ego}}$, the objective is defined as
\begin{align}
\begin{split}
    \mathcal{L}_{\text{CVAE}} =
    \,&\mathbb{E}_{q_\phi(\mathbf{z}_\text{lat} \mid \mathbf{a}_{\text{ego}}, \mathbf{z})}
    \bigl[
    \mathcal{L}_{\text{rec}}\bigl(\mathbf{a}_{\text{ego}}, \hat{\mathbf{a}}_{\text{ego}}\bigr)
    \bigr] \\
    &+
    \beta \,
    D_{\text{KL}}
    \bigl(
    q_\phi(\mathbf{z}_\text{lat} \mid \mathbf{a}_{\text{ego}}, \mathbf{z})
    \;\|\;
    p(\mathbf{z}_\text{lat})
    \bigr),
\end{split}
\end{align}
where $q_\phi(\mathbf{z}_\text{lat} \mid \mathbf{a}_{\text{ego}}, \mathbf{z})$ denotes the encoder posterior, and the prior is chosen as a standard normal distribution $p(\mathbf{z}_\text{lat}) = \mathcal{N}(0,I)$. The scalar $\beta$ controls the trade-off between reconstruction accuracy and latent regularisation.

The reconstruction loss $\mathcal{L}_{\text{rec}}$ is defined as the NLL of the decoder output. Assuming a Gaussian likelihood with fixed diagonal covariance, it reduces to a mean squared error:
\begin{equation}
    \mathcal{L}_{\text{rec}}
    =
    \lVert
    \mathbf{a}_{\text{ego}} - \hat{\mathbf{a}}_{\text{ego}}
    \rVert_2^2.
\end{equation}
The KL divergence term has a closed-form expression:
\begin{equation}
    D_{\text{KL}}
    =
    \frac{1}{2}
    \sum_{i=1}^{d_\text{lat}}
    \Bigl(
    \mu_i^2 + \sigma_i^2 - \log \sigma_i^2 - 1
    \Bigr),
\end{equation}
where $\boldsymbol{\mu}$ and $\boldsymbol{\sigma}$ are the mean and standard deviation predicted by the encoder.

This architecture can be viewed as a simplified attention-based variant of \textit{Trajectron++}, a graph-based recurrent model for trajectory forecasting \cite{salzmann_TrajectronDynamicallyFeasibleTrajectory_2020a}. In contrast to \textit{Trajectron++}, the long short-term memory (LSTM) modules \cite{hochreiter_LongShortTermMemory_1997} are replaced by attention mechanisms to model temporal dependencies, and continuous control actions are predicted directly instead of future trajectories generated by gated recurrent units (GRUs) \cite{cho_LearningPhraseRepresentations_2014}. Moreover, we do not incorporate road context maps or explicit group agents into categories, as the vehicle type is implicitly encoded in the state representation through the vehicle length.

We set the history length to $h=4$, and use the same training, validation, and test splits as those adopted for the MLP and MDN models. The attention-based architecture employs an embedding dimension of 32 and 4 attention heads, with a dropout rate of $0.1$, layer normalisation, and residual connections applied to both cross-attention modules.

For the CVAE, the encoder consists of a two-layer MLP with a hidden dimension of 64, while the decoder is a three-layer MLP with the same hidden dimension. The latent dimension is set to 16. During training, we use a batch size of 64, a learning rate of $10^{-3}$, and the AdamW optimiser with a weight decay of $10^{-4}$. Gradient clipping is applied with a maximum norm of $1.0$, and the weighting coefficient is fixed at $\beta=1.0$. Training is conducted for a maximum of 200 epochs with early stopping patience set to 30 epochs. We do not apply $\beta$-annealing, as preliminary experiments indicate that both linear and cyclical annealing degrade performance. Early stopping is triggered at the 41st epoch, achieving a best test loss of 0.52.

\subsection{Model Validation}\label{subapp:model_validation_highway}

\begin{figure*}
    \centering
    \includegraphics[width=1\linewidth]{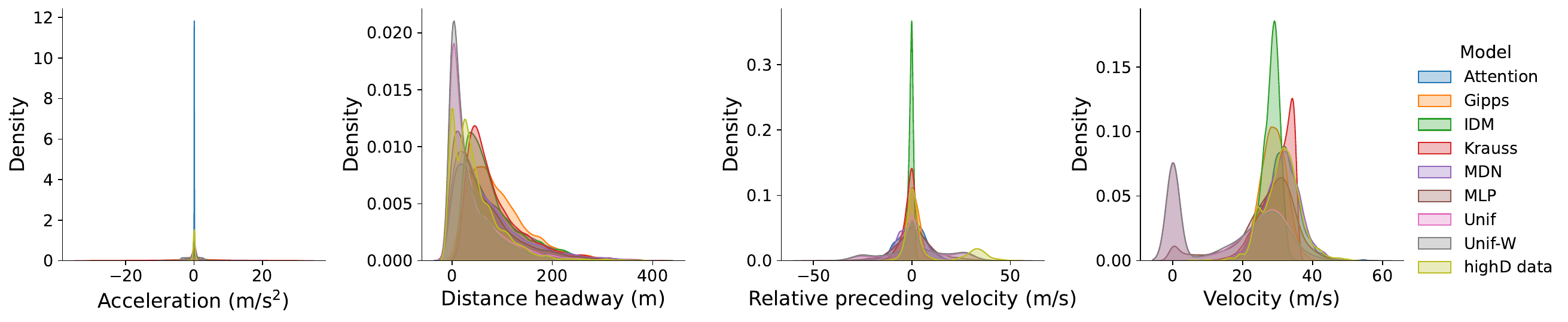}
    \caption{Kernel density estimates of four driving feature distributions generated by each calibrated car-following model, overlaid with the corresponding distributions from highD data.}
    \label{fig:highway_model_feature_distributions}
\end{figure*}

To verify that the calibrated models reproduce the statistical characteristics of naturalistic driving, we compare the distributions of key driving features generated by each model against those observed in the highD data. Figure~\ref{fig:highway_model_feature_distributions} shows the corresponding kernel density estimates, while Table~\ref{tab:highway_models_feature_stats} summarises the feature statistics and Table~\ref{tab:highway_models_feature_kl} reports the corresponding KL divergences. The results show that the calibrated models generally capture the central tendencies of the data, although their variability differs across features. In particular, the data-driven models MLP, MDN, and Attention achieve lower KL divergence for several features, while the rule-based models also reproduce the observed distributions reasonably well. Overall, these results confirm that the calibrated models capture meaningful distributional characteristics of naturalistic driving.

\begin{table*}[th]
\centering
\caption{Summary statistics of driving features across different car-following models and highD data.}
\label{tab:highway_models_feature_stats}
\begin{tabular}{llrrrrrr}
\toprule
Feature & Model & Mean & Variance & Std. Dev. & Median & Q$_{25}$ & Q$_{75}$ \\
\midrule
\multirow{9}{*}{Velocity}
 & highD data & 31.182 & 27.570 & 5.251 & 31.450 & 27.960 & 34.380 \\
 & Gipps & 28.642 & 15.168 & 3.895 & 28.796 & 26.219 & 31.289 \\
 & Krauss & 29.936 & 20.549 & 4.533 & 31.045 & 27.201 & 33.648 \\
 & IDM & 28.125 & 6.866 & 2.620 & 28.508 & 26.713 & 29.885 \\
 & MLP & 26.619 & 83.163 & 9.119 & 28.578 & 23.183 & 32.446 \\
 & MDN & 31.333 & 26.904 & 5.187 & 31.536 & 27.971 & 34.650 \\
 & Attention & 30.831 & 26.577 & 5.155 & 31.068 & 27.632 & 34.128 \\
 & Unif & 16.793 & 191.191 & 13.827 & 21.382 & 0.154 & 28.787 \\
 & Unif-W & 16.535 & 188.677 & 13.736 & 20.586 & 0.153 & 28.420 \\
\midrule
\multirow{9}{*}{Relative preceding velocity}
 & highD data & 7.756 & 198.075 & 14.074 & 0.820 & -0.390 & 6.920 \\
 & Gipps & -0.442 & 17.629 & 4.199 & -0.040 & -2.820 & 2.365 \\
 & Krauss & -1.220 & 14.869 & 3.856 & -0.687 & -3.127 & 1.084 \\
 & IDM & -0.743 & 3.246 & 1.802 & -0.372 & -1.331 & 0.289 \\
 & MLP & 1.184 & 110.508 & 10.512 & 0.878 & -4.536 & 6.406 \\
 & MDN & -0.526 & 49.642 & 7.046 & -0.350 & -5.170 & 3.931 \\
 & Attention & -0.095 & 51.873 & 7.202 & 0.309 & -4.766 & 4.746 \\
 & Unif & 2.408 & 214.666 & 14.651 & 0.128 & -2.787 & 9.677 \\
 & Unif-W & 2.379 & 245.052 & 15.654 & 0.135 & -4.641 & 11.979 \\
\midrule
\multirow{9}{*}{Distance headway}
 & highD data & 48.062 & 2688.000 & 51.846 & 33.720 & 14.370 & 68.340 \\
 & Gipps & 95.033 & 3843.995 & 62.000 & 80.656 & 49.331 & 125.472 \\
 & Krauss & 87.350 & 3615.240 & 60.127 & 68.247 & 45.099 & 111.160 \\
 & IDM & 82.293 & 3503.209 & 59.188 & 63.721 & 39.493 & 107.245 \\
 & MLP & 60.134 & 3855.602 & 62.093 & 41.090 & 14.938 & 85.792 \\
 & MDN & 70.036 & 4541.234 & 67.389 & 49.689 & 20.350 & 98.957 \\
 & Attention & 76.388 & 4438.971 & 66.626 & 58.192 & 25.220 & 110.644 \\
 & Unif & 41.064 & 2870.817 & 53.580 & 20.481 & 3.814 & 59.314 \\
 & Unif-W & 35.615 & 2301.892 & 47.978 & 17.797 & 3.969 & 50.476 \\
\midrule
\multirow{9}{*}{Acceleration}
 & highD data & 0.001 & 0.104 & 0.323 & 0.020 & -0.170 & 0.190 \\
 & Gipps & 0.351 & 38.156 & 6.177 & 0.332 & -3.763 & 4.480 \\
 & Krauss & 0.620 & 145.241 & 12.052 & 0.701 & -7.717 & 9.111 \\
 & IDM & 0.216 & 4.443 & 2.108 & 0.210 & -1.195 & 1.627 \\
 & MLP & -0.205 & 2.734 & 1.654 & -0.049 & -0.567 & 0.326 \\
 & MDN & 0.142 & 3.470 & 1.863 & 0.138 & -0.195 & 0.428 \\
 & Attention & 0.051 & 0.003 & 0.056 & 0.044 & 0.024 & 0.077 \\
 & Unif & -0.497 & 4.075 & 2.019 & -0.489 & -2.244 & 1.247 \\
 & Unif-W & -0.502 & 4.075 & 2.019 & -0.496 & -2.251 & 1.244 \\
\bottomrule
\end{tabular}
\end{table*}

\begin{table*}[th]
\centering
\caption{KL divergence $D_{\mathrm{KL}}(P_{\text{highD}} \,\|\, Q_{\text{model}})$ of driving feature distributions between highD data and each car-following model (lower is better).}
\label{tab:highway_models_feature_kl}
\begin{tabular}{lrrrr}
\toprule
Model & Velocity & Relative preceding velocity & Distance headway & Acceleration \\
\midrule
Gipps & 0.024 & 0.510 & 0.214 & 0.148 \\
Krauss & 0.067 & 0.515 & 0.316 & 0.188 \\
IDM & 0.142 & 0.687 & 0.219 & 0.087 \\
MLP & 0.010 & 0.071 & \textbf{0.013} & 0.029 \\
MDN & 0.002 & 0.338 & 0.031 & \textbf{0.024} \\
Attention & \textbf{0.001} & 0.340 & 0.053 & 0.267 \\
Unif & 0.030 & \textbf{0.048} & 0.024 & 0.099 \\
Unif-W & 0.032 & 0.055 & 0.031 & 0.099 \\
\bottomrule
\end{tabular}
\end{table*}

\section{Handover Giver Model Details}\label{app:handover_giver_model_details}

\subsection{Minimum Jerk Model}\label{subapp:minijerk}

The Minimum Jerk model is a classical model of human reaching movement that assumes humans generate trajectories by minimising the integrated squared jerk, where jerk is the third derivative of position with respect to time~\cite{flash_CoordinationArmMovements_1985}. Given an initial position $\mathbf{p}_0$ and a target position $\mathbf{p}_T$, the model generates a smooth trajectory $\mathbf{p}(t)$ by solving

\begin{equation}
    \min_{\mathbf{p}(t)}
    \int_0^T
    \left\|
    \frac{d^3 \mathbf{p}(t)}{dt^3}
    \right\|_2^2
    dt,
\end{equation}

subject to the boundary conditions

\begin{equation}
\begin{aligned}
    \mathbf{p}(0) &= \mathbf{p}_0, &
    \mathbf{p}(T) &= \mathbf{p}_T, \\
    \dot{\mathbf{p}}(0) &= \dot{\mathbf{p}}(T) = \mathbf{0}, &
    \ddot{\mathbf{p}}(0) &= \ddot{\mathbf{p}}(T) = \mathbf{0}.
\end{aligned}
\end{equation}

The resulting trajectory can be expressed as

\begin{equation}
    \mathbf{p}(t)
    =
    \mathbf{p}_0
    +
    \left(
    10\tau^3
    -15\tau^4
    +6\tau^5
    \right)
    (\mathbf{p}_T-\mathbf{p}_0),
    \label{eq:min_jerk}
\end{equation}

where $\tau=t/T\in[0,1]$ denotes the normalised movement time.

In our handover setting, the Minimum Jerk model is used to generate the trajectory of the human giver's hand from its initial position towards the handover location. The model therefore produces a smooth, deterministic movement between the initial and target positions, with the movement duration $T$ controlling the temporal evolution of the trajectory. The Minimum Jerk model has no learnable parameters. However, generating a trajectory requires specifying its initial position, final position, and duration. To obtain these quantities, we model the distributions of the trajectory start positions, end positions, and lengths using Gaussian distributions fitted to the corresponding quantities in the human training data. During deployment, these quantities are independently sampled from their respective Gaussian distributions and used to generate a Minimum Jerk trajectory.

\subsection{Elliptical Model}\label{subapp:elliptical}

The Elliptical model represents each human trajectory as a planar movement along an ellipse. For each demonstration, we first assess whether the trajectory can be reasonably approximated as planar by performing principal component analysis (PCA) on its three-dimensional coordinates. We perform a planarity test based on the variance explained by the three principal components. For each trajectory, PCA is applied independently to its three-dimensional coordinates, yielding three eigenvalues corresponding to the variance along PC1, PC2, and PC3. A planar trajectory is expected to exhibit substantially smaller variance along PC3 than along the first two principal axes.

We first compare the eigenvalues across the three principal axes using a repeated-measures ANOVA, where the trajectory index is treated as the subject and the principal axis (PC1, PC2, or PC3) is the within-subject factor. The ANOVA indicates a significant difference in variance among the three principal axes ($F(2,1136)=6628.24$, $p_{\mathrm{GG}}<0.001$). The Greenhouse-Geisser correction is applied because the sphericity assumption is violated ($W=0.00163$, $p<0.001$; $\epsilon=0.5004$).

Following the significant omnibus test, we perform pairwise paired $t$-tests between the principal axes with Bonferroni correction. All three pairwise comparisons are statistically significant: PC1 has significantly greater variance than PC2 ($t(568)=81.48$, $p_{\mathrm{corr}}<0.001$), PC1 has significantly greater variance than PC3 ($t(568)=81.37$, $p_{\mathrm{corr}}<0.001$), and PC2 has significantly greater variance than PC3 ($t(568)=23.80$, $p_{\mathrm{corr}}<0.001$). The corresponding effect sizes are also substantial, with Hedges' $g$ values of 4.78, 4.82, and 1.36 for the PC1-PC2, PC1-PC3, and PC2-PC3 comparisons, respectively.

Most importantly, the variance along PC3 is substantially smaller than that along both PC1 and PC2. We therefore consider the trajectories to have a dominant two-dimensional structure and use the first two principal components to define the movement plane. Each trajectory is subsequently projected onto this plane before fitting the elliptical model. An ellipse is fitted to the projected trajectory using the direct least-squares method of Fitzgibbon et al.~\cite{fitzgibbon_DirectLeastSquare_1999}, yielding the conic parameters
\begin{equation}
Ax^2 + Bxy + Cy^2 + Dx + Ey + F = 0.
\end{equation}
The fitted ellipse is parameterised by its centre, orientation, and semi-axis lengths. For each demonstration, we additionally record its start and end points, trajectory length, the two principal axes defining the movement plane, and the three-dimensional offset of the plane. These quantities are then modelled statistically to capture variation across human demonstrations.

Specifically, trajectory length is modelled using a univariate Gaussian distribution, while the start and end points together with the ellipse parameters are modelled using a multivariate Gaussian. A separate multivariate Gaussian is fitted to the two principal axes and the three-dimensional plane offset. During generation, these distributions are sampled to obtain the trajectory geometry, spatial orientation, and duration.

For each sampled trajectory, the start and end points are projected onto the sampled principal plane, and the shortest arc of the sampled ellipse connecting these points is generated. The resulting two-dimensional trajectory is then mapped back into three-dimensional space using the sampled principal axes and offset. To impose a smooth temporal profile, the spatial trajectory is subsequently reparameterised according to a minimum-jerk profile,
\begin{equation}
s(\tau) = 10\tau^3 - 15\tau^4 + 6\tau^5,
\qquad \tau \in [0,1].
\end{equation}
Finally, a linear endpoint correction is applied to ensure that the generated trajectory exactly matches the sampled start and end points. The resulting trajectory therefore captures both the characteristic elliptical spatial geometry and the variability in movement duration, endpoints, orientation, and trajectory shape observed in the human demonstrations.

\subsection{Dynamic Movement Primitive (DMP)}\label{subapp:dmp}

The Dynamic Movement Primitive (DMP) model represents a movement as a combination of a stable spring--damper system and a nonlinear forcing function~\cite{ijspeert_DynamicalMovementPrimitives_2013}. This formulation allows the general shape of a demonstrated trajectory to be captured while retaining the ability to adapt the initial position, goal position, and movement duration. For each spatial dimension, the DMP is defined by the transformation system

\begin{equation}
\tau \dot{v} = K(g-y)-Dv+(g-y_0)f(s),
\end{equation}

\begin{equation}
\tau \dot{y}=v,
\end{equation}

where $y$ denotes the position, $v$ is a scaled velocity variable, $y_0$ and $g$ are the initial and goal positions, respectively, and $\tau$ controls the movement duration. The parameters $K$ and $D$ determine the stiffness and damping of the underlying spring--damper system. The nonlinear forcing term $f(s)$ captures the characteristic shape of the demonstrated movement and is modulated by a canonical system

\begin{equation}
\tau \dot{s}=-\alpha_s s,
\end{equation}
whose solution is

\begin{equation}
s(t)=\exp\left(-\alpha_s\frac{t}{\tau}\right).
\end{equation}

The forcing function is represented using a set of $N$ Gaussian basis functions:

\begin{equation}
f(s)=\frac{\sum_{i=1}^{N}w_i\psi_i(s)}{\sum_{i=1}^{N}\psi_i(s)}s,
\end{equation}

where $w_i$ denotes the weight of the $i$-th basis function and

\begin{equation}
\psi_i(s)=\exp\left[-h_i(s-c_i)^2\right].
\end{equation}

In our implementation, the basis-function centres are exponentially distributed in the phase variable according to

\begin{equation}
c_i=\exp\left(-\alpha_s\frac{i-1}{N-1}\right),
\end{equation}

with the basis-function widths determined by

\begin{equation}
h_i=\frac{N}{c_i^2}.
\end{equation}

We use $N=20$, $K=100$, $D=20$, and $\alpha_s=4.6$. To fit the DMP to a human demonstration, the position, velocity, and acceleration are first obtained from the recorded trajectory using numerical differentiation. The target forcing function is then computed from the demonstrated trajectory by rearranging the DMP transformation system:

\begin{equation}
f_{\mathrm{target}}(t)=\frac{\tau^2\ddot{y}(t)-K(g-y(t))+D\tau\dot{y}(t)}{g-y_0}.
\end{equation}

The basis-function weights are subsequently estimated using locally weighted regression. This procedure is performed independently for each spatial dimension of the trajectory. Rather than using a single set of fitted weights, we model the variability across human demonstrations probabilistically. Specifically, a DMP is fitted independently to every human trajectory, and the resulting weight vectors are collected across demonstrations. A multivariate Gaussian distribution is then fitted to these weight vectors, with the empirical mean and covariance used as the parameters of the model:

\begin{equation}
\mathbf{w}\sim\mathcal{N}\left(\boldsymbol{\mu}_w,\boldsymbol{\Sigma}_w\right).
\end{equation}

During trajectory generation, the initial position, goal position, and movement duration are sampled from the corresponding distributions learned from the human training data. A DMP weight vector is independently sampled from the fitted multivariate Gaussian distribution and used in the transformation system to generate the trajectory. The system is then numerically integrated using Euler integration. This produces a stochastic DMP model that captures both the characteristic movement shape and variation observed in the human demonstrations.

\subsection{MLP}\label{subapp:mlp-handover}

The original giver trajectories are downsampled to $20\,\mathrm{Hz}$, and a sliding window is applied to construct the state representation. Each state consists of five consecutive frames, from $T-4$ to $T$, containing the standardised 3D coordinates of the giver's and taker's right hands, resulting in a state representation $\mathbf{s}_T \in \mathbb{R}^{5\times6}$, which is flattened to $\mathbb{R}^{30}$ for the MLP. The prediction target is the standardised 3D displacement of the giver's hand between $T$ and $T+1$. The MLP follows the same architecture as the model described in Section~\ref{subapp:mlp-highway}, with only the input and output dimensions adjusted to the handover task. No other architectural modifications are introduced, and the model is trained using the same supervised learning procedure as the highway MLP model.

\subsection{MDN}\label{subapp:mdn-handover}

The MDN model used in the object handover domain follows the same architecture and training procedure as the highway MDN model described in Section~\ref{subapp:mdn-highway}. The only architectural modification is the number of Gaussian mixture components in the output distribution, which is reduced from 15 to 5. Specifically, given the input trajectory representation, the network predicts the parameters of a five-component Gaussian mixture distribution over the next-step displacement of the giver's hand. The reduction in the number of mixture components is intended to reflect the lower behavioural complexity of the object handover task. All other architectural and training settings are kept consistent with the corresponding highway MDN model.

\subsection{Attention}\label{subapp:attention-handover}

The attention-based model used in the object handover domain is designed to capture temporal dependencies among the frames in the input trajectory window. Unlike the attention model used in the highway domain, which uses a more complex architecture, the handover model consists of a single multi-head self-attention layer followed by temporal pooling and a small MLP.

The input to the model is a sequence of five consecutive frames, with each frame represented by the 6-dimensional Cartesian coordinates of the giver's and taker's right hands. Thus, the input has the form
\begin{equation}
\mathbf{X} \in \mathbb{R}^{T \times 6},
\qquad T=5.
\end{equation}
Each 6-dimensional frame is first projected into a 128-dimensional embedding space using a linear layer,
\begin{equation}
\mathbf{Z} = \mathbf{X}\mathbf{W}_{e}+\mathbf{b}_{e},
\qquad
\mathbf{Z}\in\mathbb{R}^{T\times128}.
\end{equation}
The resulting sequence is then processed by a multi-head self-attention layer with four attention heads. Query, key, and value are all obtained from the same projected sequence, allowing the model to capture dependencies between different time steps. The attention output is combined with the original embedded features through a residual connection and subsequently normalised using layer normalisation:
\begin{equation}
\mathbf{Z}' =
\operatorname{LayerNorm}
\left(
\mathbf{Z}+
\operatorname{MHA}(\mathbf{Z},\mathbf{Z},\mathbf{Z})
\right).
\end{equation}
The temporal dimension is then removed using global average pooling,
\begin{equation}
\mathbf{z}_{\mathrm{pool}}
=
\frac{1}{T}\sum_{t=1}^{T}\mathbf{Z}'_t,
\qquad
\mathbf{z}_{\mathrm{pool}}\in\mathbb{R}^{128}.
\end{equation}
The pooled representation is passed through a two-layer MLP consisting of a 128-unit linear layer, a ReLU activation, and a final linear layer. The final layer predicts the 3-dimensional displacement of the giver's hand at the next time step. The model is trained using the same supervised learning procedure as the other data-driven handover models. This architecture provides a lightweight mechanism for modelling temporal dependencies within the five-frame observation window while avoiding the additional complexity of deeper attention architectures.

\subsection{Model Validation}\label{subapp:model_validation_handover}

\begin{figure*}
    \centering
    \includegraphics[width=0.8\linewidth]{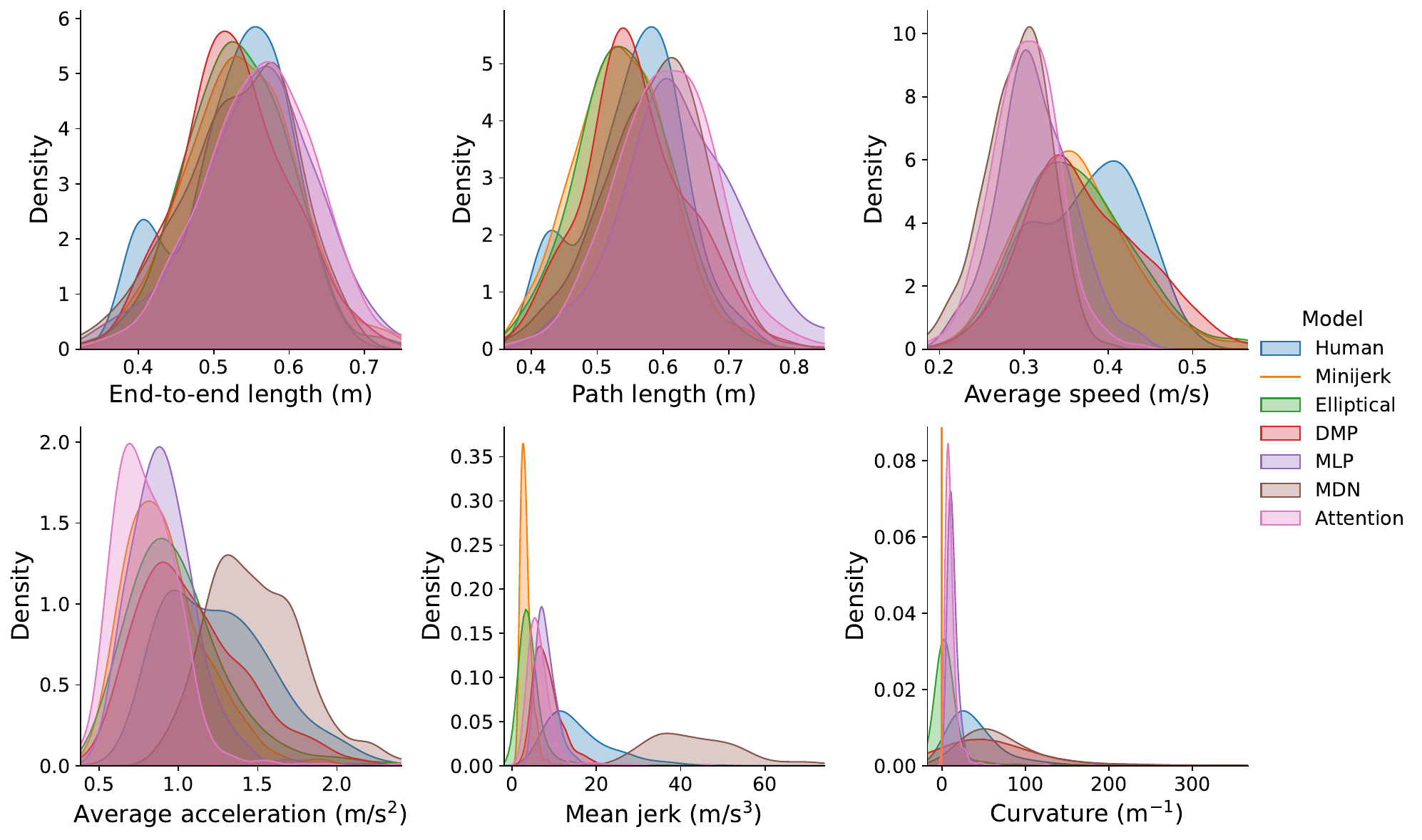}
    \caption{Comparison of handover giver trajectory feature distributions between calibrated giver models and human data. Kernel density estimates are shown for end-to-end length, path length, average speed, average acceleration, movement smoothness, and trajectory curvature.}
    \label{fig:giver_feature_distributions}
\end{figure*}

To verify that the calibrated models reproduce the statistical characteristics of human handover behaviour, we compare the distributions of key trajectory features generated by each giver model against those observed in the human data. Figure~\ref{fig:giver_feature_distributions} shows the corresponding kernel density estimates, while Table~\ref{tab:giver_feature_stats} summarises the feature statistics and Table~\ref{tab:giver_feature_kl} reports the corresponding KL divergences. The results show that the calibrated models capture several characteristics of the human trajectories, although their performance varies substantially across features. In particular, DMP achieves the lowest KL divergence for path length, average speed, and average acceleration, and performs consistently well across the remaining features. The data-driven models achieve the lowest divergence for some individual features, but generally show larger discrepancies in average speed and other trajectory characteristics. Overall, these results indicate that the calibrated models capture meaningful aspects of human handover behaviour while retaining distinct behavioural characteristics.

\begin{table*}[th]
\centering
\caption{Summary statistics of handover giver trajectory features across giver models and human data.}
\label{tab:giver_feature_stats}
\begin{tabular}{llrrrrrr}
\toprule
Feature & Model & Mean & Variance & Std. Dev. & Median & Q$_{25}$ & Q$_{75}$ \\
\midrule
\multirow{7}{*}{End-to-end length}
 & Human & 0.529 & 0.005 & 0.071 & 0.541 & 0.490 & 0.581 \\
 & Minijerk & 0.537 & 0.005 & 0.072 & 0.536 & 0.489 & 0.583 \\
 & Elliptical & 0.531 & 0.005 & 0.069 & 0.529 & 0.485 & 0.579 \\
 & DMP & 0.528 & 0.005 & 0.072 & 0.524 & 0.483 & 0.575 \\
 & MLP & 0.556 & 0.007 & 0.082 & 0.561 & 0.506 & 0.608 \\
 & MDN & 0.531 & 0.006 & 0.076 & 0.539 & 0.483 & 0.587 \\
 & Attention & 0.560 & 0.006 & 0.076 & 0.564 & 0.511 & 0.614 \\
\midrule
\multirow{7}{*}{Path length}
 & Human & 0.556 & 0.006 & 0.075 & 0.565 & 0.510 & 0.607 \\
 & Minijerk & 0.537 & 0.005 & 0.072 & 0.536 & 0.489 & 0.583 \\
 & Elliptical & 0.544 & 0.008 & 0.089 & 0.541 & 0.494 & 0.588 \\
 & DMP & 0.559 & 0.006 & 0.080 & 0.550 & 0.510 & 0.609 \\
 & MLP & 0.628 & 0.008 & 0.089 & 0.622 & 0.571 & 0.686 \\
 & MDN & 0.584 & 0.006 & 0.079 & 0.594 & 0.536 & 0.638 \\
 & Attention & 0.604 & 0.006 & 0.079 & 0.604 & 0.552 & 0.656 \\
\midrule
\multirow{7}{*}{Average speed}
 & Human & 0.373 & 0.004 & 0.062 & 0.378 & 0.322 & 0.420 \\
 & Minijerk & 0.360 & 0.004 & 0.067 & 0.355 & 0.313 & 0.401 \\
 & Elliptical & 0.365 & 0.006 & 0.075 & 0.359 & 0.317 & 0.404 \\
 & DMP & 0.372 & 0.005 & 0.068 & 0.361 & 0.322 & 0.418 \\
 & MLP & 0.314 & 0.002 & 0.045 & 0.311 & 0.285 & 0.343 \\
 & MDN & 0.292 & 0.002 & 0.039 & 0.297 & 0.268 & 0.319 \\
 & Attention & 0.302 & 0.002 & 0.039 & 0.302 & 0.276 & 0.328 \\
\midrule
\multirow{7}{*}{Average acceleration}
 & Human & 1.275 & 0.177 & 0.421 & 1.229 & 0.968 & 1.475 \\
 & Minijerk & 0.926 & 0.075 & 0.275 & 0.882 & 0.726 & 1.054 \\
 & Elliptical & 1.009 & 0.158 & 0.398 & 0.944 & 0.782 & 1.139 \\
 & DMP & 1.085 & 0.117 & 0.342 & 1.017 & 0.835 & 1.296 \\
 & MLP & 0.914 & 0.050 & 0.224 & 0.893 & 0.767 & 1.028 \\
 & MDN & 1.489 & 0.093 & 0.306 & 1.465 & 1.269 & 1.690 \\
 & Attention & 0.799 & 0.046 & 0.214 & 0.776 & 0.652 & 0.922 \\
\midrule
\multirow{7}{*}{Movement smoothness}
 & Human & 16.448 & 108.472 & 10.415 & 13.789 & 10.281 & 19.494 \\
 & Minijerk & 3.303 & 1.846 & 1.359 & 3.030 & 2.353 & 3.867 \\
 & Elliptical & 4.638 & 30.531 & 5.525 & 3.508 & 2.662 & 4.874 \\
 & DMP & 8.493 & 11.857 & 3.443 & 7.702 & 6.013 & 10.073 \\
 & MLP & 8.307 & 8.120 & 2.850 & 7.708 & 6.444 & 9.442 \\
 & MDN & 43.700 & 127.901 & 11.309 & 42.147 & 35.351 & 50.413 \\
 & Attention & 6.763 & 16.035 & 4.004 & 5.904 & 4.659 & 7.671 \\
\midrule
\multirow{7}{*}{Trajectory curvature}
 & Human & 45.331 & 3242.138 & 56.940 & 31.024 & 18.426 & 52.665 \\
 & Minijerk & 0.000 & 0.000 & 0.000 & 0.000 & 0.000 & 0.000 \\
 & Elliptical & 10.088 & 1273.335 & 35.684 & 1.458 & 0.445 & 5.266 \\
 & DMP & 70.640 & 2.53e+04 & 159.159 & 44.277 & 29.865 & 69.188 \\
 & MLP & 13.333 & 103.234 & 10.160 & 11.377 & 8.553 & 15.579 \\
 & MDN & 80.261 & 7008.926 & 83.719 & 59.767 & 39.334 & 88.505 \\
 & Attention & 10.602 & 80.425 & 8.968 & 8.348 & 5.946 & 12.159 \\
\bottomrule
\end{tabular}
\end{table*}

\begin{table*}[th]
\centering
\caption{KL divergence $D_{\mathrm{KL}}(P_{\text{human}} \,\|\, Q_{\text{model}})$ of handover giver trajectory feature distributions between human data and each giver model (lower is better).}
\label{tab:giver_feature_kl}
\begin{tabular}{lrrrrrr}
\toprule
Model & End-to-end length & Path length & Average speed & Average acceleration & Movement smoothness & Trajectory curvature \\
\midrule
Minijerk & 0.054 & 0.064 & 0.084 & 0.696 & 11.743 & 1.602 \\
Elliptical & 0.060 & 0.060 & 0.069 & 0.399 & 2.298 & 1.109 \\
DMP & 0.067 & \textbf{0.057} & \textbf{0.068} & \textbf{0.250} & 2.299 & 0.288 \\
MLP & 0.118 & 0.336 & 1.059 & 0.951 & 2.397 & 4.083 \\
MDN & \textbf{0.037} & 0.093 & 4.088 & 0.644 & 7.449 & \textbf{0.270} \\
Attention & 0.140 & 0.211 & 1.824 & 1.619 & \textbf{1.871} & 3.770 \\
\bottomrule
\end{tabular}
\end{table*}

\section{Trajectory Generation and Collection}
\label{app:trajectory_generation_and_collection}

We generate trajectories from each calibrated human behaviour model using a consistent simulation procedure within each domain. The generated trajectories are subsequently used for offline evaluation.

\subsection{Highway Trajectory Generation}\label{subapp:highway_trajectory_generation}

For the highway domain, the simulation scenario consists of a standard three-lane highway segment with a length of $420\,\mathrm{m}$, matching the road-segment length of the highD dataset. At each simulation step, a new vehicle is spawned in each lane according to a spawn probability estimated from the three-lane recordings in the highD dataset, provided that a sufficiently large gap exists to ensure safe insertion.

For the rule-based driving models, each newly spawned vehicle is assigned model parameters sampled from the corresponding posterior distributions inferred from the human data. To initialise vehicle states with realistic values, we perform kernel density estimation (KDE) on the initial vehicle speeds and vehicle lengths observed in the highD dataset. The initial speed and vehicle length of each newly spawned vehicle are then sampled from the corresponding KDE distributions.

For each driving model, simulations are run for $10{,}000$ frames at $25\,\mathrm{Hz}$, matching the sampling frequency of the highD dataset. A highway trajectory is defined as the time series of the $48$-dimensional state representation described in Eq.~\ref{eq:dataDrivenStates} for an individual vehicle. We collect trajectories generated by each of the eight driving models in $\{H_i^{\mathrm{HW,train}}\}$.

\subsection{Handover Trajectory Generation}\label{subapp:handover_trajectory_generation}

For the handover domain, trajectory generation follows the customised handover environment described in Section~\ref{subapp:object_handover}. For each of the six giver models in $\{H_i^{\mathrm{HO,train}}\}$, we run $500$ episodes and collect one giver trajectory from each episode, resulting in $500$ trajectories per model.

During trajectory generation, taker trajectories are sampled from the empirical human data and replayed as the taker's motion. Each generated handover trajectory therefore consists of the corresponding giver and taker trajectories, represented as a time series of their six-dimensional hand positions.

\section{Trajectory Representations}\label{app:trajectory_representations}

Existing generative model evaluation metrics are typically computed in a feature space rather than directly on raw trajectories~\cite{Ali2025}. The choice of trajectory representation can therefore influence the resulting evaluation scores and, consequently, the assessment of human behaviour models. To investigate this effect, we consider three approaches for representing generated trajectories: PCA-based representations, learned latent representations obtained from an autoencoder, and manually extracted behavioural features. The first two representations are used to compute distributional evaluation metrics, while the manually extracted features are used to compare the corresponding feature distributions between human and generated behaviours.

\subsection{PCA Representation}\label{subapp:pca_representation}

For the PCA-based representation, Principal Component Analysis (PCA) is applied to the collected human trajectories, and the resulting principal axes are used to project both human and generated trajectories into a lower-dimensional feature space. As the reference configuration, we retain the first six principal components for the highway domain and the first four principal components for the handover domain.

\subsection{Latent Representation}\label{subapp:latent_representation}

Generative model evaluation metrics are commonly computed using representations extracted by a deep neural network trained on an auxiliary task using real data~\cite{Ali2025}. Since metrics such as FID were originally developed for image generation, this auxiliary task is often image classification. However, the highD dataset does not naturally provide class labels. Therefore, rather than training a classifier, we adopt an autoencoder framework and learn latent representations through state sequence reconstruction.

Specifically, the trajectory of each identified vehicle in the highD dataset is segmented into multiple sequences of duration \(T\) seconds. The vehicle states are standardised before segmentation. As the highD data are downsampled to \(5~\mathrm{Hz}\), each sequence contains \(5T\) consecutive samples. We set \(T=5\) seconds in this study, resulting in sequences of 25 time steps.

\begin{figure}
    \centering
    \includegraphics[width=\linewidth]{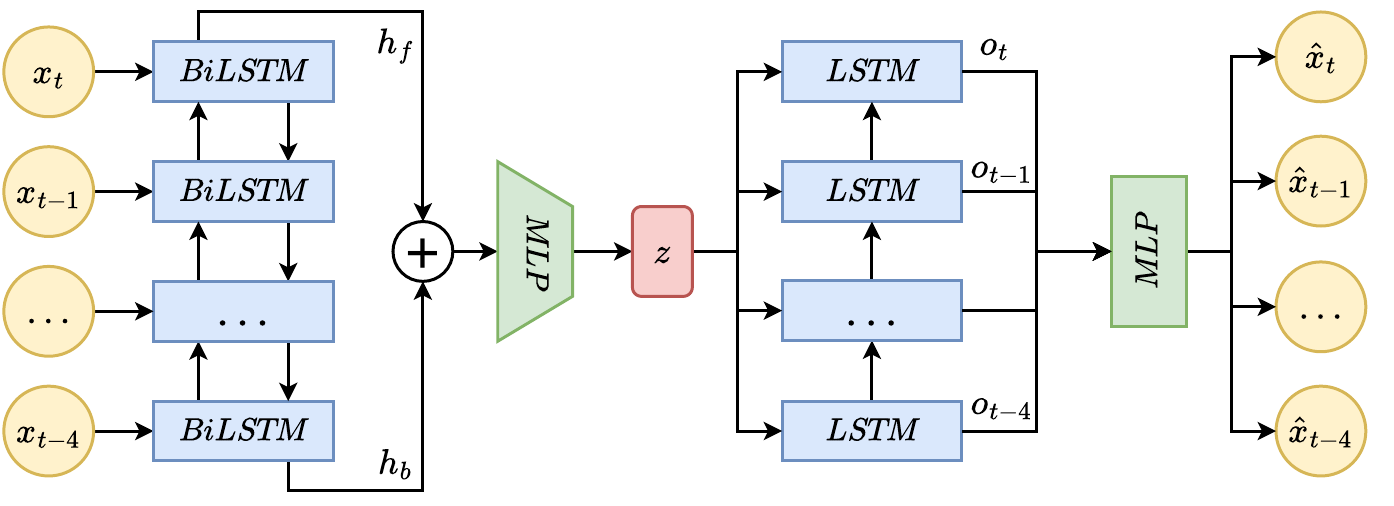}
    \caption{Architecture of the autoencoder used for latent feature extraction. A BiLSTM encoder maps the input state sequence to a latent vector \(z\), which is decoded by a repeated-vector LSTM and an MLP to reconstruct the original sequence.}
    \label{fig:autoencoder}
\end{figure}

The architecture of the autoencoder is illustrated in Figure~\ref{fig:autoencoder}. Given an input state sequence \(X \in \mathbb{R}^{5T \times d}\), where \(T=5\) and \(d=48\) for the highway domain, the sequence is first processed by a bidirectional LSTM (BiLSTM) encoder, which has previously been used for driving behaviour modelling~\cite{sun_IdentificationDrivingBehavior_2025}. The encoder produces the final hidden states from the forward and backward passes, denoted by \(h_f\) and \(h_b\), respectively. These hidden states are concatenated to form a joint representation, which is subsequently processed by an MLP to produce the latent vector \(z\).

During training, the latent vector \(z\) is decoded using a repeated-vector LSTM~\cite{ghimire_StackedLSTMSequencetoSequence_2022}. Specifically, \(z\) is replicated across the temporal dimension and provided as input to the LSTM at each time step. The resulting sequence of hidden states is then processed by a second MLP to reconstruct the original state sequence, denoted by \(\hat{X} \in \mathbb{R}^{5T \times d}\). The autoencoder is trained by minimising the mean squared reconstruction error,
\begin{equation}
    \mathcal{L}_{\mathrm{AE}}
    =
    \frac{1}{5Td}
    \sum_{t=1}^{5T}
    \left\lVert
        \mathbf{x}_t - \hat{\mathbf{x}}_t
    \right\rVert_2^2.
\end{equation}

For the reference configuration, the hidden dimensions of both the BiLSTM encoder and repeated-vector LSTM decoder are set to 128. The encoder and decoder MLPs each contain a single hidden layer with 128 units, and the latent dimension is set to 32. The autoencoder is trained using the human trajectories from the highD dataset. The trained encoder is subsequently used to extract latent features \(z\) from both the human trajectories and trajectories generated by the human behaviour models. The generated trajectories are downsampled from \(25~\mathrm{Hz}\) to \(5~\mathrm{Hz}\) to match the input resolution used to train the autoencoder. These latent representations are then used to compute the model evaluation metrics.

For the handover domain, we use the same general autoencoder architecture, with the input adapted to the handover trajectory representation. Specifically, each LSTM node receives a single frame containing the 6-dimensional state of the giver and taker. The autoencoder is trained using the handover@RPL dataset and is subsequently used to extract latent representations from both the human trajectories and the trajectories generated by each giver model.

\subsection{Manually Extracted Behavioural Features}\label{subapp:manual_representation}

In the highway driving domain, we consider speed, acceleration, distance headway, and relative preceding speed. In the object handover domain, we consider acceleration, speed, movement smoothness, curvature, trajectory end-to-end length, and path length. For each feature, we estimate the distributions of the human and generated data using kernel density estimation (KDE) and compare them using KL divergence. As the reference configuration, we use a KDE sample size of 10,000 for the highway domain. For the handover domain, we use all available samples, as the trajectories are relatively short and the resulting computation is efficient.

\section{Evaluation Metrics}\label{app:evaluation_metrics}

\subsection{Fr\'echet Distance (FD)}\label{subapp:fd}

Fr\'echet Distance is a widely used metric for evaluating generative models, originally proposed for image generation tasks~\cite{frechen_inceptive_distance} as the \textit{Fr\'echet Inception Distance} (\textit{FID}). It has since been adapted to other domains, including gesture and video generation~\cite{maiorcaObjectiveEvaluationMetric2023, wangThisThatLanguageGestureControlled2025}. It measures the similarity between real and generated images by comparing feature representations extracted from a pretrained convolutional neural network. In practice, \textit{FD} models the extracted features of real and generated samples as multivariate Gaussian distributions and computes the Fr\'echet distance \cite{frechet_QuelquesPointsCalcul_1906} between these. A lower \textit{FD} score indicates that the generated data distribution is closer to the real data distribution, suggesting better quality and diversity of the generated samples.

Assuming we have the generated data distribution and the real data distribution as two multivariate Gaussian distributions $P_{\text{gen}} \sim \mathcal{N}(\mu_{\text{gen}}, \Sigma_{\text{gen}})$ and $P_{\text{real}} \sim \mathcal{N}(\mu_{\text{real}}, \Sigma_{\text{real}})$, the \textit{FD} between these is defined as
\begin{equation}\label{eq:FD}
\mathrm{FD} = \|\mu_{\text{gen}} - \mu_{\text{real}}\|_2^2 
+ \mathrm{Tr}\!\left( \Sigma_{\text{gen}} + \Sigma_{\text{real}} 
- 2 \left( \Sigma_{\text{gen}} \Sigma_{\text{real}} \right)^{\frac{1}{2}} \right),
\end{equation}
where $\|\cdot\|_2$ denotes the Euclidean norm and $\mathrm{Tr}(\cdot)$ denotes the trace operator. A lower \textit{FD} value indicates that the generated distribution is closer to the real distribution in terms of both mean and covariance, implying better generation quality. 

Note that if the generated data perfectly matches the real data distribution, the \textit{FD} would be zero. In practice, however, achieving an \textit{FD} of zero is neither realistic nor desired. Therefore, \textit{FD} should be interpreted as a relative metric for comparing generative models. Specifically, given two models with $\mathrm{FD}_{\text{gen1}} > 0$ and $\mathrm{FD}_{\text{gen2}} > 0$, if 
\[
\mathrm{FD}_{\text{gen1}} > \mathrm{FD}_{\text{gen2}},
\]
then generative model~2 is considered to produce samples that are more faithful to the real data distribution than generative model~1 \cite{Ali2025}.

\subsection{Density}\label{subapp:density}

The \textit{density} metric, an improved version of \textit{precision} \cite{precision_recall_distributions} introduced together with coverage \cite{density_coverage}, is used to evaluate generative models by assessing how well generated samples represent the real data manifold. \textit{Density} focuses on the quality of generated samples in regions where real data exist.

Specifically, \textit{density} quantifies how many generated samples fall within a local neighbourhood of real data points in the feature space. A higher \textit{density} score indicates that generated samples are well aligned with the real data distribution and are not scattered in unrealistic regions.

Let $\mathcal{X}_{\text{real}} = \{x_i\}_{i=1}^{N}$ denote the set of real data samples and $\mathcal{X}_{\text{gen}} = \{\hat{x}_j\}_{j=1}^{M}$ denote the set of generated samples. For each real sample $x_i$, let $B(x_i, \mathrm{NND}_k(x_i))$ be the hypersphere centred at $x_i$ with radius $\mathrm{NND}_k(x_i)$, defined as the distance from $x_i$ to its $k$-th nearest neighbour. The \textit{density} metric is then defined as
\begin{equation}\label{eq:density}
    \textit{density} = \frac{1}{kM} \sum_{j=1}^{M} \sum_{i=1}^{N} \mathds{1}_{\hat{x}_j \in B(x_i, \mathrm{NND}_k(x_i))}.
\end{equation}

Higher values of \textit{density} indicate that the generated samples closely cover the regions of the real data manifold. Note that \textit{density} can exceed 1 if most generated samples fall within the hyperspheres of more than $k$ real samples.

\subsection{Coverage}\label{subapp:coverage}

The \textit{coverage} metric \cite{density_coverage} is an improved version of \textit{recall} in \cite{precision_recall_distributions}. It measures the fraction of real data points that are close to at least one generated sample in the feature space, indicating how well the generative model captures the support of the real data distribution. Formally, for a set of real samples $\mathcal{X}_{\text{real}} = \{x_i\}_{i=1}^{N}$ and generated samples $\mathcal{X}_{\text{gen}} = \{\hat{x}_j\}_{j=1}^{M}$, coverage is defined as
\begin{equation}\label{eq:coverage}
\textit{coverage} = \frac{1}{N} \sum_{i=1}^{N} \mathds{1}_{\exists \, j \, \text{s.t. } \hat{x}_j \in B(x_i, \mathrm{NND}_k(x_i))},
\end{equation}
where $B(x_i, \mathrm{NND}_k(x_i))$ is defined the same as in \ref{eq:density}. Higher coverage values indicate that the generated samples adequately cover the real data distribution.

\subsection{Average Pair Distance (APD)}\label{subapp:apd}

The \textit{average pair distance} (\textit{APD}) represents a general metric for measuring diversity. This metric has been applied in many existing studies, for example in image generation~\cite{zhang_UnreasonableEffectivenessDeep_2018} and gesture generation~\cite{guo_Action2MotionConditionedGeneration_2020}.

Given a dataset, a random selection procedure is repeated $R$ times. In each repetition, $S$ pairs of samples are randomly chosen, and the Euclidean distance is computed for each pair. The \textit{APD} is then obtained by averaging these distances across all repetitions. Formally, for a dataset $\mathcal{X}$, \textit{APD} is defined as
\begin{equation}\label{eq:APD}
    \text{APD}(\mathcal{X}) = \frac{1}{R S} \sum_{r=1}^{R} \sum_{i=1}^{S} \| x_i - \tilde{x}_i \|_2,
\end{equation}
where $x_{i}$ and $\tilde{x}_{i}$ denote the elements of the $i$-th pair of samples randomly drawn from $\mathcal{X}$.

\subsection{Mean Maximum Similarity (MMS)}\label{subapp:mms}

MMS was proposed in~\cite{cervantes_ImplicitNeuralRepresentations_2022} as a novelty metric to complement diversity measures such as coverage, which can be misleading when generated samples perfectly replicate real samples, resulting in zero novelty. Formally, the MMS between generated samples $\mathcal{X}_{\text{gen}} = \{\hat{x}_j\}_{j=1}^{M}$ and real samples $\mathcal{X}_{\text{real}} = \{x_i\}_{i=1}^{N}$ is defined as
\begin{equation}
    \text{MMS}(\mathcal{X}_{\text{gen}}, \mathcal{X}_{\text{real}}) 
    = \frac{1}{M} \sum_{j=1}^{M} \min_{i \in \mathcal{X}_{\text{real}}} \| \hat{x}_j - x_i \|_2.
\end{equation}

In~\cite{cervantes_ImplicitNeuralRepresentations_2022}, the novelty of generated samples is quantified as the gap between 
$\text{MMS}(\mathcal{X}_{\text{gen}}, \mathcal{X}_{\text{real}})$ and 
$\text{MMS}(\mathcal{X}_{\text{real}}, \mathcal{X}_{\text{real}})$, where the latter is computed on the real dataset itself. In this case, the Euclidean distance excludes self-comparisons when identifying the nearest neighbour.

\subsection{KL Divergence}\label{subapp:kl}

For a manually extracted feature \(f\), let \(p_{\mathrm{H}}^{(f)}(x)\) and \(p_{\mathrm{G}}^{(f)}(x)\) denote the probability density functions estimated from the human and generated data, respectively. The KL divergence is computed as
\begin{equation}
    D_{\mathrm{KL}}^{(f)}
    \left(
        p_{\mathrm{H}}^{(f)}
        \,\middle\|\,
        p_{\mathrm{G}}^{(f)}
    \right)
    =
    \int
    p_{\mathrm{H}}^{(f)}(x)
    \log
    \frac{
        p_{\mathrm{H}}^{(f)}(x)
    }{
        p_{\mathrm{G}}^{(f)}(x)
    }
    \, dx.
\end{equation}

Given \(F\) manually extracted features, the average KL divergence is defined as
\begin{equation}
    \overline{D}_{\mathrm{KL}}
    =
    \frac{1}{F}
    \sum_{f=1}^{F}
    D_{\mathrm{KL}}^{(f)}.
\end{equation}

\section{Reinforcement Learning Details}
\label{app:rl_details}

\subsection{Highway Driving}\label{subapp:highway_driving_rl_details}

We train the highway driving policies using the Deep Q-Network (DQN) algorithm implemented in Stable-Baselines3. Each policy is trained in the \texttt{highway-fast-v0} environment, with the behaviour of the surrounding vehicles determined by one of the human behaviour models described in the main paper. The same RL algorithm, policy architecture, and hyperparameters are used across all training configurations to ensure that differences in downstream performance arise from the human behaviour model used during training.

The DQN agent uses an MLP policy with two hidden layers of 256 units each. The learning rate is set to $5\times10^{-4}$, with a replay buffer size of 15{,}000 transitions and a batch size of 32. Learning begins after 200 environment steps. The discount factor is $\gamma=0.8$, and one gradient step is performed after every environment step. The target network is updated every 50 training steps. All models are trained on the CPU.

For each human behaviour model, training is repeated using three random seeds. The random seed is applied consistently to the RL training procedure to control stochasticity across repeated training runs. Each training run is performed for 40{,}000 environment steps. The resulting policies are saved and subsequently evaluated against all human behaviour models in the corresponding domain, as described in the main paper.

\subsection{Object Handover}\label{subapp:handover_rl_details}

We train the object handover policies using the Soft Actor-Critic (SAC) algorithm implemented in Stable-Baselines3. Each policy is trained in the custom \texttt{HandoverSim} environment, with the behaviour of the giver determined by one of the human behaviour models described in the main paper. As in the highway domain, the same RL training configuration is used across all giver models.

The SAC agent uses an MLP policy and is trained using 16 parallel environments. The initial learning rate is $10^{-3}$ and is gradually reduced to $5\times10^{-4}$ using a sigmoid learning-rate schedule over the course of training. The replay buffer contains up to 50{,}000 transitions, and learning begins after 1{,}000 environment steps. The batch size is 128, the discount factor is $\gamma=0.99$, and the agent performs eight environment steps between gradient updates and eight gradient steps per update. All models are trained on the CPU.

Each training configuration is repeated using three random seeds, with explicit seeding applied to the RL algorithm and environment. Each policy is trained for 100{,}000 environment steps. During training, episode reward, episode duration, and task success are recorded for monitoring. The resulting policies are subsequently evaluated against each of the six testing giver models for 20 episodes per giver model. During evaluation, the trained policies act deterministically, and the mean episode reward, mean episode duration, and success rate are recorded.

For both domains, the RL training procedure is kept fixed across human behaviour models within the domain. Consequently, the human behaviour model used during training is the primary experimental factor, allowing the downstream effects of different human behaviour models to be compared under a consistent RL training procedure.

\end{document}